\documentclass{article}

\PassOptionsToPackage{numbers, compress}{natbib}
\usepackage[final]{neurips_2022}

\usepackage[utf8]{inputenc} % allow utf-8 input
\usepackage[T1]{fontenc}    % use 8-bit T1 fonts
\usepackage{hyperref}       % hyperlinks
\usepackage{url}            % simple URL typesetting
\usepackage{booktabs}       % professional-quality tables
\usepackage{amsfonts}       % blackboard math symbols
\usepackage{nicefrac}       % compact symbols for 1/2, etc.
\usepackage{microtype}      % microtypography
\usepackage{xcolor}         % colors
\usepackage[utf8]{inputenc} % allow utf-8 input
\usepackage[T1]{fontenc}    % use 8-bit T1 fonts
\usepackage{hyperref}       % hyperlinks
\usepackage{url}            % simple URL typesetting
\usepackage{booktabs}       % professional-quality tables
\usepackage{amsfonts}       % blackboard math symbols
\usepackage{nicefrac}       % compact symbols for 1/2, etc.
\usepackage{graphicx}
\usepackage{algorithm}
\usepackage{wrapfig}
\usepackage{amsmath}
\usepackage{algpseudocode}
\usepackage[normalem]{ulem}
\useunder{\uline}{\ul}{}
\usepackage{bbding}
\usepackage{multirow}
\usepackage{amssymb}
\usepackage{dsfont}
\usepackage{caption}
\usepackage{diagbox}

\title{TOIST: Task Oriented Instance Segmentation Transformer with Noun-Pronoun Distillation}

\author{%
Pengfei Li\textsuperscript{1} \quad Beiwen Tian\textsuperscript{1} \quad Yongliang Shi\textsuperscript{1} \quad Xiaoxue Chen\textsuperscript{1} \NewAnd Hao Zhao\textsuperscript{2,3} \quad Guyue Zhou\textsuperscript{1} \quad Ya-Qin Zhang\textsuperscript{1} \\
$^1$AIR, Tsinghua University \quad $^2$ Peking University \quad $^3$Intel Labs \\
  \texttt{li-pf22@mails.tsinghua.edu.cn} \\ \texttt{zhao-hao@pku.edu} \quad \texttt{hao.zhao@intel.com}\\
}

\begin{document}

\maketitle

\begin{abstract}
  Current referring expression comprehension algorithms can effectively detect or segment objects indicated by nouns, but how to understand verb reference is still under-explored. As such, we study the challenging problem of task oriented detection, which aims to find objects that best afford an action indicated by verbs like \emph{sit comfortably on}. Towards a finer localization that better serves downstream applications like robot interaction, we extend the problem into task oriented instance segmentation. A unique requirement of this task is to select \emph{preferred} candidates among possible alternatives. Thus we resort to the transformer architecture which naturally models pair-wise query relationships with attention, leading to the TOIST method. In order to leverage pre-trained noun referring expression comprehension models and the fact that we can access privileged noun ground truth during training, a novel noun-pronoun distillation framework is proposed. Noun prototypes are generated in an unsupervised manner and contextual pronoun features are trained to select prototypes. As such, the network remains noun-agnostic during inference. We evaluate TOIST on the large-scale task oriented dataset COCO-Tasks and achieve +10.9\% higher $\rm{mAP^{box}}$ than the best-reported results. The proposed noun-pronoun distillation can boost $\rm{mAP^{box}}$ and $\rm{mAP^{mask}}$ by +2.8\% and +3.8\%. Codes and models are publicly available at \url{https://github.com/AIR-DISCOVER/TOIST}.
\end{abstract}

\section{Introduction}

\begin{wrapfigure}{r}{0.5\textwidth}
\begin{center}
\includegraphics[width=0.5\textwidth]{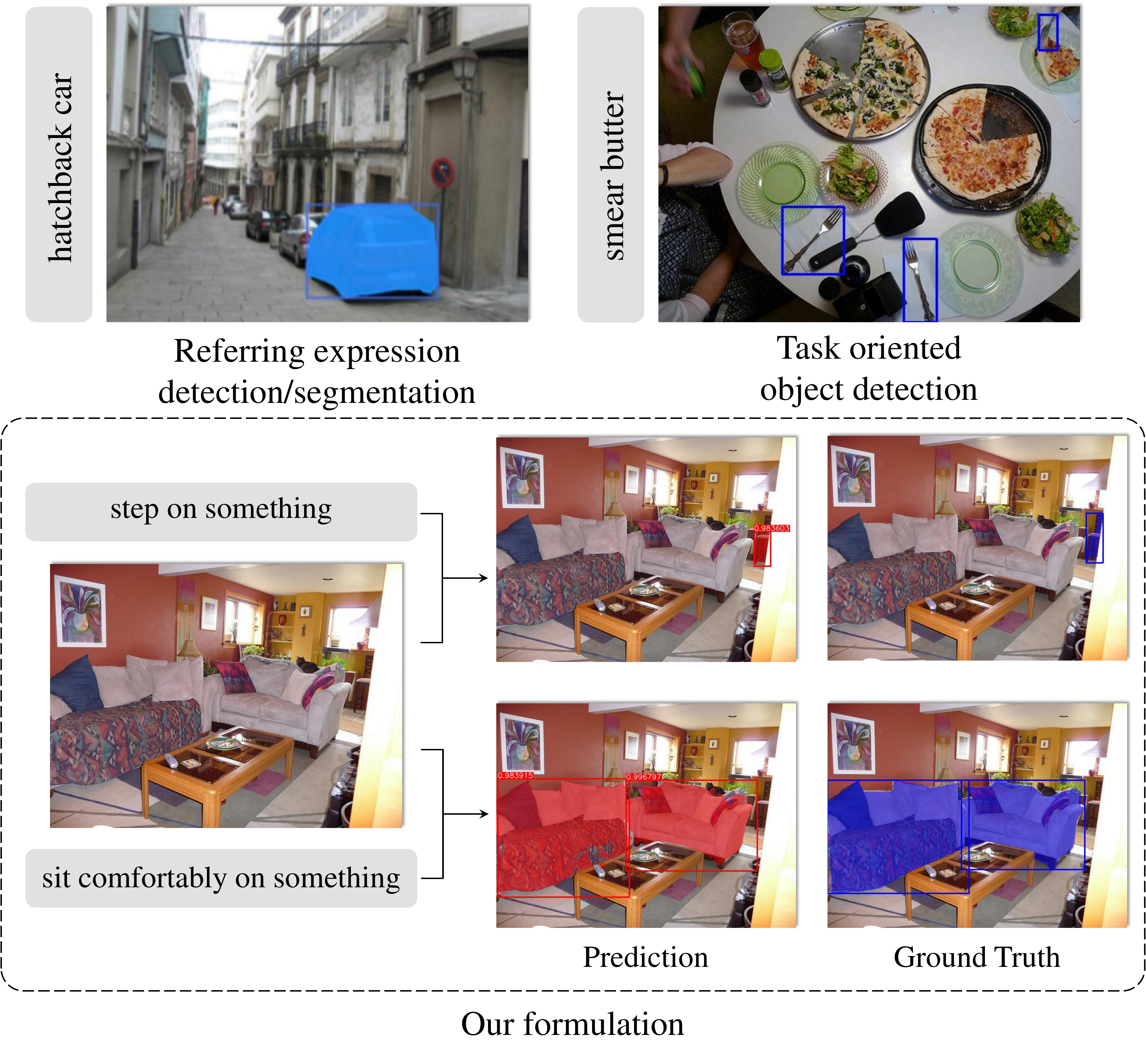}
\end{center}
\vspace{-1.5em}
%\caption{Left top: users specify noun input \emph{hatchback car} and a referring expression understanding model segments the car with an instance mask. Right top: the task driven detector finds objects that afford the task of \emph{smearing butter}. Bottom: our TOIST model segments different objects that afford different verb-pronoun references, using a multimodal transformer architecture.}
\caption{Left top: Noun referring expression comprehension. Right top: Task oriented detection. Bottom: Task oriented instance segmentation.}
\vspace{-1em}
\label{fig:teaser}
\end{wrapfigure}

As benchmarked by the RefCOCO, RefCOCO+ \cite{kazemzadeh2014referitgame}\cite{yu2016modeling} and RefCOCO-g \cite{mao2016generation} datasets, noun referring expression comprehension models have seen tremendous progress, thanks to large-scale vision-language pre-training models like VL-BERT \cite{su2019vl}, VilBERT \cite{lu2019vilbert}, OSCAR \cite{li2020oscar}, UNITER \cite{chen2020uniter}, 12-in-1 \cite{lu202012} and MDETR \cite{kamath2021mdetr}. As shown in the left top part of Fig.~\ref{fig:teaser}, these algorithms take noun prompts like \emph{hatchback car} as inputs and generate a bounding box or an instance mask of that car. However, in real-world applications like intelligent service robots, system inputs usually come in the form of \emph{affordance} (i.e., the capability to support an action or say a verb phrase). Whether modern vision-language model designs can effectively understand verb reference remains under-explored.

To this end, we focus on the challenging problem of task oriented detection, as introduced by the COCO-Tasks benchmark \cite{sawatzky2019object}. As shown in the right top part of Fig.~\ref{fig:teaser}, a task oriented detector outputs three boxes of forks as they can be used to \emph{smear butter}. We also extend the problem to an upgraded instance segmentation version using existing COCO masks \cite{lin2014microsoft}, as the masks can provide finer localization. When RGB-D pairs are available, instance masks can be used to obtain object point clouds. When image sequences are available, instance masks can be used to reconstruct objects using visual hull \cite{lazebnik2001computing}\cite{cheung2003visual}\cite{zuo2015interactive}. As such, the newly proposed task oriented instance segmentation formulation (Fig.~\ref{fig:teaser} bottom) is useful for down-stream robot interaction applications. 

While noun referring expression comprehension datasets aim to minimize ambiguity \cite{mao2016generation}, an interesting and challenging feature of task oriented detection/segmentation is the intrinsic ambiguity. For example, in the right top panel of Fig.~\ref{fig:teaser}, the pizza peel can also be used to \emph{smear butter}. If we have neither forks nor pizza peels at hand, it is still possible to use the plate to \emph{smear butter}. Another example is shown in Fig.~\ref{fig:teaser} bottom. When we consider an object to \emph{step on}, the chair is a better choice because the sofa is soft and the table is heavy to move. When the need switches to \emph{sit comfortably on}, sofas are obviously the best candidates. In one word, objects that afford a verb are ambiguous and the algorithm needs to model \emph{preference}.

To this end, current models \cite{sawatzky2019object} use a two-stage pipeline, in which objects are firstly detected then relatively ranked. 
Inspired by the success of DETR-like methods \cite{carion2020end}\cite{zhu2020deformable}\cite{liu2021dab} \textcolor[RGB]{0,0,0} {and the advantage of the attention mechanism in revealing the relationship between visual elements \cite{li2022distance}\cite{zhao2020learning}}, we resort to the transformer architecture as it imposes self-attention on object queries thus naturally models the pair-wise relative preference between object candidates. 
Our one-stage method is named as \textbf{T}ask \textbf{O}riented \textbf{I}nstance \textbf{S}egmentation \textbf{T}ransformer and abbreviated as TOIST. Transformers are considered to be data hungry \cite{brown2020language}\cite{dosovitskiy2020image}, but obtaining large-scale visually grounded verb reference data with relative preference (e.g., COCO-Tasks \cite{sawatzky2019object}) is difficult. This inspires us to explore the possibility of reusing knowledge in noun referring expression comprehension models. We propose to use pronouns like \emph{something} as a proxy and distill knowledge from noun embedding prototypes generated by clustering. 

Specifically, we first train a TOIST model with verb-noun input (e.g., \emph{step on chair} for the bottom panel of Fig.~\ref{fig:teaser} bottom), using the privileged noun ground truth. But during inference, we cannot access the noun \emph{chair}, thus we train the second TOIST model with verb-pronoun input (e.g., \emph{step on something}) and distill knowledge from the first TOIST model. As such, the second TOIST model remains noun-agnostic during inference and achieves better performance than directly training a model with verb-pronoun input. This framework is named as \emph{noun-pronoun distillation}. Although leveraging knowledge from models with privileged information has been used in robotics research like autonomous driving \cite{chen2020learning} and quadrupedal locomotion \cite{lee2020learning}, the proposed paradigm of distilling privileged noun information into pronoun features is novel, to the best of our knowledge.

To summarize, this paper has the following four contributions: 

%(1) We upgrade the task oriented detection task into task oriented instance segmentation and provide the first solution to it. Although this is a natural extension, this new formulation is of clear practical value to robotics applications. (2) Unlike existing two-stage models that firstly detect objects then rank them, we propose the first transformer-based method TOIST, for task oriented detection/segmentation. It has only one stage and naturally models relative preference with self-attention on object queries. (3) In order to leverage the privileged information in noun refering expression understanding models, we propose a novel noun-pronoun distillation framework that bridges nouns and pronouns using discovered prototypes. It improves TOIST with up to 2.0 mAP. (4) We achieve new state-of-the-art results on the large-scale COCO-Tasks dataset, out-performing the best reported results by 10.0 mAP. Codes and models are publicly available. 

\begin{itemize}
\item We upgrade the task oriented detection task into task oriented instance segmentation and provide the first solution to it. Although this is a natural extension, this new formulation is of practical value to robotics applications.
\item Unlike existing two-stage models that firstly detect objects then rank them, we propose the first transformer-based method TOIST, for task oriented detection/segmentation. It has only one stage and naturally models relative preference with self-attention on object queries. 
\item In order to leverage the privileged information in noun referring expression understanding models, we propose a novel noun-pronoun distillation framework. It improves TOIST by +2.8\% and +3.8\% for $\rm{mAP^{box}}$ and $\rm{mAP^{mask}}$, respectively.
\item We achieve new state-of-the-art (SOTA) results on the COCO-Tasks dataset, out-performing the best reported results by +10.9\% $\rm{mAP^{box}}$. Codes and models are publicly available. 
\end{itemize}

\section{Related Works}

\textbf{Vision and Language.} Connecting vision and language is a long-existing topic for visual scene understanding. Barnard et al. \cite{barnard2003matching} propose a system that translates image regions into nouns. Babytalk \cite{kulkarni2013babytalk} is an early method that turns images into sentences, based upon conditional random fields. Visual question answering \cite{antol2015vqa} aims to answer questions about an image, with potential applications in helping visually impaired people \cite{gurari2018vizwiz}. The CLEVR dataset \cite{johnson2017clevr} focuses on the reasoning ability of question answering models, thanks to a full control over the synthetic data. The Flickr30k benchmark \cite{plummer2015flickr30k} addresses the phrase grounding task that links image regions and descriptions. Vision-language navigation \cite{anderson2018vision}\cite{fried2018speaker} aims to learn navigation policies that fulfill language commands. DALL-E \cite{ramesh2021zero} shows impressive text-to-image generation capability. Video dense captioning \cite{krishna2017dense}\cite{duan2018weakly} generates language descriptions for detected salient regions. Visual madlibs \cite{yu2015visual} focuses on fill-the-blank question answering. Visual commonsense reasoning \cite{zellers2019recognition} proposes the more challenging task of justifying an answer. Referring expression detection \cite{yu2016modeling}\cite{mao2016generation} and segmentation \cite{hu2016segmentation} localize objects specified by nouns. Although this literature is very large with many problem formulations proposed, the task of detecting objects that afford \emph{verbs} (e.g., COCO-Tasks\cite{sawatzky2019object}) is still under-explored.

\textbf{Action and Affordance.} Verbs, as the link between subjects and objects, have been extensively studied in the both vision \cite{wang2013dense}\cite{nagarajan2020learning} and language \cite{hosseini2014learning}\cite{mccarthy2003detecting} communities before, while we focus on the vision side. It is difficult to define standalone \emph{verb recognition} tasks, so existing problem formulations depend on the focus on subjects or objects. A simple taxonomy can be considered as such: recognizing \emph{subjects and verbs} is named as action recognition \cite{simonyan2014two}\cite{girdhar2017attentional}; recognizing \emph{verbs and objects} is named as affordance recognition \cite{zhu2015understanding}\cite{wang2017binge}\cite{do2018affordancenet}\textcolor[RGB]{0,0,0}{\cite{chen2022cerberus}}; recognizing \emph{triplets} is named as human-object-interaction recognition \cite{li2020hoi}\cite{zhang2021mining}. Task oriented object detection/segmentation, as an affordance understanding task, is very challenging and this study explores the noun-pronoun distillation framework to borrow rich knowledge from more visually grounded noun targets.

\textbf{Knowledge Distillation.} This technique is proposed in the deep learning literature by Hinton et al. \cite{hinton2015distilling} to distill knowledge from large models to small models. The insight is that soft logit targets generated by large models contain richer information that better serves as a supervision signal than hard one-hot labels. Knowledge distillation has been extended to show effectiveness in other domains like continual learning \cite{li2017learning} and object detection \cite{chen2017learning}. The survey of Guo et al. \cite{gou2021knowledge} provides a comprehensive summary of knowledge distillation variants and applications. Most related to our method is the privileged knowledge distillation methods in robotics research \cite{chen2020learning}\cite{lee2020learning}, in which the teacher model has access to privileged information that the student cannot access during inference. Our noun-pronoun distillation method is tailored for the task oriented detection/segmentation problem which borrows rich knowledge from noun referring expression compression teacher models while still allowing the student model to be noun-agnostic.

\section{Formulation} \label{section:formulation}
The problem is to detect and segment objects that are preferred to afford a specific task indicated by verb phrases, from an input image. Yet clearly defining \textbf{affordance} and \textbf{preference} is actually challenging so we follow the existing annotation protocol of the COCO-Tasks dataset \cite{sawatzky2019object}:

\textbf{Affordance.} Firstly, the target objects \emph{afford} a specific task. In an input image, it is possible that no objects or multiple objects afford the task. And in the latter case, the objects may belong to multiple classes. For example, in the right top panel of Fig.\ref{fig:teaser}, nothing affords the task \emph{sit comfortably on}. Instead, in the bottom panel, there are at least two sofas, a table and a chair that afford the task.

\textbf{Preference.} Secondly, we need to find the \emph{best} ones from the objects which afford the task. In other words, the \emph{preference} among multiple objects needs to be understood. In Fig.\ref{fig:teaser} bottom, the two sofas are obviously more suitable for the task \emph{sit comfortably on} than other objects enumerated above. Thus the ground truth objects for this task are the two sofas (covered in blue).

%\textbf{Instance Mask.} The problem of task oriented detection stops at detecting the bounding boxes of objects, which lacks fine-grained information and brings ambiguity to down-stream robotics applications.
%For example, in Fig.\ref{fig:teaser} bottom, two ground truth bounding boxes aim to indicate the two selected sofas.
%However, part of the table is also covered in the area of boxes and can be misunderstood by robots. To eliminate this ambiguity, \emph{instance segmentation mask} becomes necessary.

Now we formally define the task.
The input is an RGB image $\mathbf{X}_v \in \mathbb{R}^{3 \times \rm{H_{0}} \times \rm{W_{0}}}$ ($v$ represents \emph{visual}) and a piece of text $\mathbf{X}_l$ ($l$ represents \emph{language}).
$\mathbf{X}_l$ describes a specific task like \emph{sit comfortably on}.
The targets are bounding boxes $\mathbf{B}_{\rm{gt}} = [{b}_{1}, \ldots, {b}_{n_{\rm{gt}}}] \in [0,1]^{n_{\rm{gt}} \times 4}$ and instance segmentation masks $\mathbf{M}_{\rm{gt}} = [{m}_{1}, \ldots, {m}_{n_{\rm{gt}}}] \in \mathbb{R}^{n_{\rm{gt}} \times \rm{H_{0}} \times \rm{W_{0}}}$ of target objects $\mathbf{O_{\rm{gt}}} = \langle\mathbf{B}_{\rm{gt}},\mathbf{M}_{\rm{gt}}\rangle$, where ${n_{\rm{gt}}} \geq 0$ is the count of targets.
The four components of $b_i \in [0,1]^{4}$ are normalized center coordinates, height and width of the $i$-th box.
An algorithm $f$ that addresses this problem works as:
\begin{equation}
    f(\langle\mathbf{X}_v, \mathbf{X}_l\rangle) = \langle\mathbf{B}_{\rm{pred}}, \mathbf{M}_{\rm{pred}}, \mathbf{S}_{\rm{pred}}\rangle,
\end{equation}
where $\mathbf{B}_{\rm{pred}}$ and $\mathbf{M}_{\rm{pred}}$ are predicted boxes and masks, respectively.
$\mathbf{S}_{\rm{pred}} = [{\hat s}_{1}, \ldots, {\hat s}_{n_{\rm{pred}}}] \in [0,1]^{n_{\rm{pred}}}$ is the probability scores of the predicted objects being selected, which reflects preference. We denote the predicted objects as $\mathbf{O_{\rm{pred}}} = \langle\mathbf{B}_{\rm{pred}},\mathbf{M}_{\rm{pred}}, \mathbf{S}_{\rm{pred}}\rangle$.

\section{Method}
\label{Method}
We propose an end-to-end \textbf{T}ask \textbf{O}riented \textbf{I}nstance \textbf{S}egmentation \textbf{T}ransformer, abbreviated as TOIST (Section \ref{toist}). Leveraging pre-trained noun referring expression comprehension models, we further adopt a teacher-student framework for noun-pronoun distillation (Section \ref{n-pn-dis}).

\subsection{Task Oriented Instance Segmentation Transformer}
\label{toist}

The SOTA method \cite{sawatzky2019object} uses a two-stage pipeline to solve the problem. Taking a single image $\mathbf{X}_v$ as input, it first detects the bounding boxes $\mathbf{B}_{\rm{pred}}$ of all objects with a Faster-RCNN \cite{ren2015faster}. Then it ranks $\mathbf{B}_{\rm{pred}}$ with a GNN \cite{li2016gated}, predicting the probabilities $\mathbf{S}_{\rm{pred}}$ of the objects being selected for a task. Our method differs from it in three ways: 
(1) We address the task with a one-stage architecture, allowing joint representation learning for detection and preference modeling. (2) We specify tasks using the text $\mathbf{X}_l$. (3) We predict instance masks $\mathbf{M}_{\rm{pred}}$ along with bounding boxes. 

\begin{figure*}
\begin{center}
\includegraphics[width=0.95\textwidth]{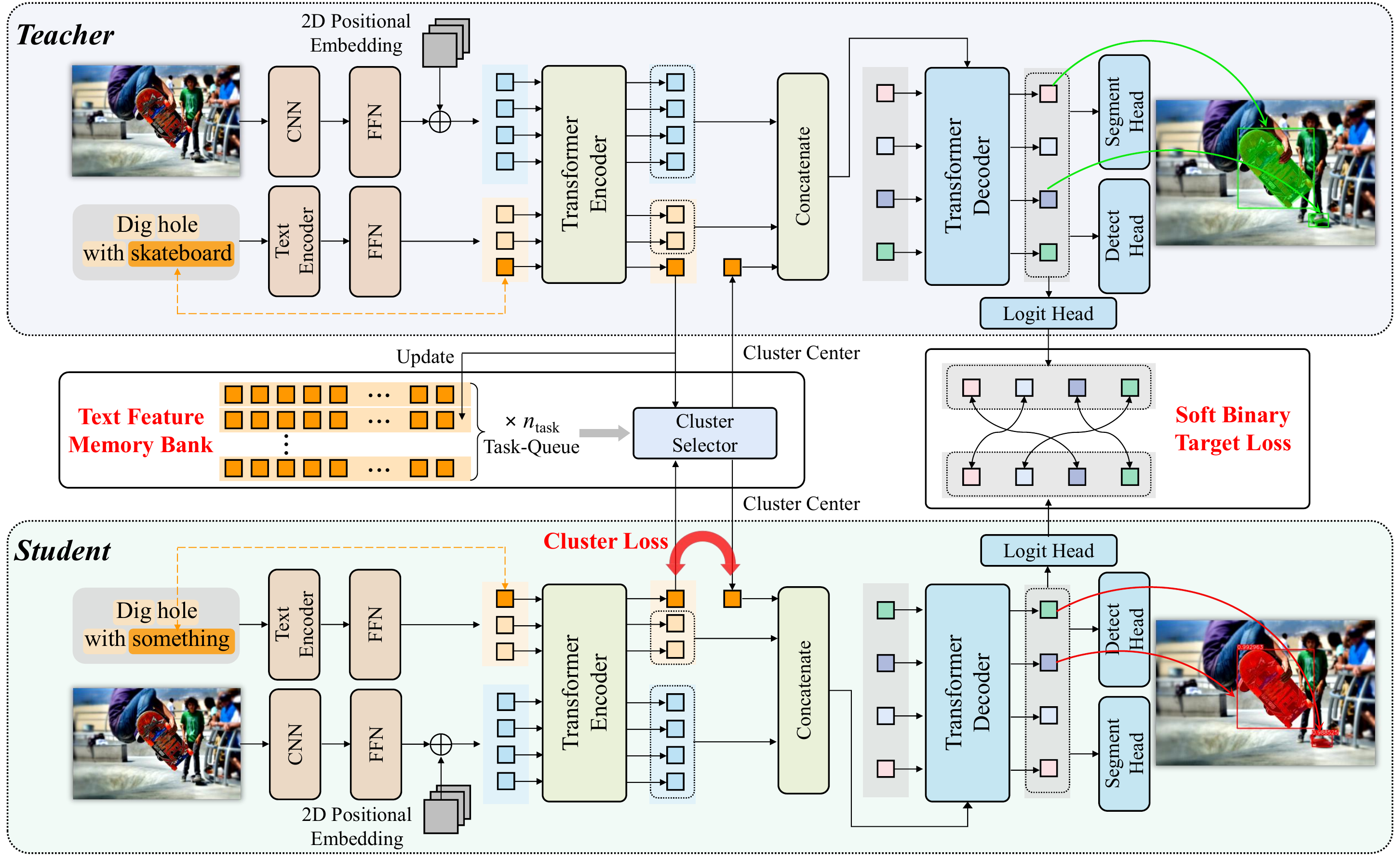}
\end{center}
\vspace{-0.5em}
\caption{TOIST network architecture and the noun-pronoun distillation framework. The cluster loss and soft binary target loss distill privileged noun knowledge and preference knowledge respectively.}
\label{fig:main}
\end{figure*}

We choose to build our method upon the transformer architecture \cite{carion2020end}, because the self-attention operators in the decoder can naturally model pair-wise relative preference between object candidates.
As shown in Fig.\ref{fig:main} bottom, TOIST contains three main components: a multi-modal encoder (\textcolor[RGB]{167,127,92} {color brown}) to extract tokenized features, a transformer encoder (\textcolor[RGB]{135,166,72} {color green}) to aggregate features of two modalities and a transformer decoder (\textcolor[RGB]{119,200,230} {color blue}) to predict the most suitable objects with attention. 

\textbf{Two Input Forms.} To find an object that affords the task of \emph{dig hole}, the first step is to construct a task description input $\mathbf{X}_l$. To achieve this goal, we can extend the task name with the ground truth object category to the \textbf{verb-noun form} like \emph{dig hole with skateboard} or with a pronoun to the \textbf{verb-pronoun form} like \emph{dig hole with something}. While the former violates the noun-agnostic constraint during inference, it can be leveraged to improve the latter within the proposed noun-pronoun distillation framework, which will be detailed later. For a plain TOIST, the verb-pronoun form is selected as task description $\mathbf{X}_l$, which is fed into the multi-modal encoder along with visual input $\mathbf{X}_v$.

% For $\mathbf{X}_v$, a pre-trained CNN-based backbone is leveraged to extract a low-resolution feature map $\mathbf{F}_v \in \mathbb{R}^{\rm{C} \times \rm{H} \times \rm{W}}$. 
% Then a one-layer feed forward network (FFN) projects it to a new feature map $\mathbf{F}_v^{\prime} \in \mathbb{R}^{d \times \rm{H} \times \rm{W}}$ with a smaller dimension $d$ for computational efficiency.
\textbf{Multi-Modal Encoder.} For $\mathbf{X}_v$, a pre-trained CNN-based backbone and a one-layer feed forward network (FFN) are leveraged to extract a low-resolution feature map $\mathbf{F}_v \in \mathbb{R}^{d \times \rm{H} \times \rm{W}}$.
Flattening the spatial dimensions of $\mathbf{F}_v$ into one dimension, we obtain a sequence of tokenized feature vectors ${\mathbf{V}} = [{v}_{1}, \ldots, {v}_{n_{{v}}}] \in \mathbb{R}^{n_{{v}} \times d}$ (\textcolor[RGB]{119,200,230}{light blue squares $\blacksquare$} on the left of the transformer encoder in Fig.\ref{fig:main}), where $n_{{v}}=\rm{H} \times \rm{W}$. 
To preserve the spatial information, 2D positional embeddings are added to ${\mathbf{V}}$. 
For $\mathbf{X}_l$, we use a pre-trained text encoder and another FFN to produce corresponding feature vectors ${\mathbf{L}} = [{l}_{1}, \ldots, {l}_{n_{{l}}}] \in \mathbb{R}^{n_{{l}} \times d}$ (\textcolor[RGB]{255,212,150}{light orange squares $\blacksquare$}), where $n_{{l}}$ is the total count of language tokens.
Among these $n_{{l}}$ features, we denotes the one corresponding to the pronoun (or noun) token as ${l_{\rm{pron}}}$ (or ${l_{\rm{noun}}}$) (\textcolor[RGB]{255,152,0}{dark orange squares $\blacksquare$}).
We concatenate these vectors and obtain the final feature sequence ${[\mathbf{V}, \mathbf{L}]} = [{v}_{1}, \ldots, {v}_{n_{{v}}}, {l}_{1}, \ldots, {l}_{n_{{l}}}] \in \mathbb{R}^{(n_{{v}}+n_{{l}}) \times d}$.

\textbf{Transformer Encoder.} The transformer encoder consists of $n_{\rm{tr}}$ sequential blocks of multi-head self-attention layers. Given the sequence of features $[\mathbf{V}, \mathbf{L}]$, it outputs processed feature sequence ${[\mathbf{V}^{\rm{tr}}, \mathbf{L}^{\rm{tr}}]} = [{v}_{1}^{\rm{tr}}, \ldots, {v}_{n_v}^{\rm{tr}}, {l}_{1}^{\rm{tr}}, \ldots, {l}_{n_l}^{\rm{tr}}] \in \mathbb{R}^{(n_{{v}}+n_{{l}}) \times d}$.
The pronoun (or noun) feature ${l_{\rm{pron}}}$ (or ${l_{\rm{noun}}}$) is encoded into ${l_{\rm{pron}}^{\rm{tr}}}$ (or ${l_{\rm{noun}}^{\rm{tr}}}$), which will be used for noun-pronoun distillation (Section \ref{n-pn-dis}).
Here for the plain TOIST, we directly use the features ${[\mathbf{V}^{\rm{tr}}, \mathbf{L}^{\rm{tr}}]}$ for later processing.

\textbf{Transformer Decoder.} The transformer decoder consists of $n_{\rm{tr}}$ blocks of self-attention and cross-attention layers.
It takes as input a set of learnable parameters serving as an object query sequence ${\mathbf{Q}} = [{q}_{1}, \ldots, {q}_{n_{\rm{pred}}}] \in \mathbb{R}^{n_{\rm{pred}} \times d}$.
${[\mathbf{V}^{\rm{tr}}, \mathbf{L}^{\rm{tr}}]}$ are used as keys and values for cross-attention layers.
The outputs of the transformer decoder are feature vectors ${\mathbf{Q}^{\rm{tr}}} = [{q}_{1}^{\rm{tr}}, \ldots, {q}_{n_{\rm{pred}}}^{\rm{tr}}] \in \mathbb{R}^{n_{\rm{pred}} \times d}$, which are projected to final results by three prediction heads.
Specifically, the detect head predicts bounding boxes $\mathbf{B}_{\rm{pred}}$ and the segment head outputs binary segmentation masks $\mathbf{M}_{\rm{pred}}$.
The logit head outputs logits $\mathbf{G}_{\rm{pred}} = [\hat{\mathbf{g}}_{1}, \ldots, \hat{\mathbf{g}}_{n_{\rm{pred}}}] \in \mathbb{R}^{{n_{\rm{pred}}} \times n_{\rm{max}}}$, where ${\hat{\mathbf{g}}_{i}} = [\hat g_1^i, \ldots, \hat g_{n_{\rm{max}}}^i]$.
$[\hat g_1^i, \ldots, \hat g_{n_l}^i]$ corresponds to text tokens ${\mathbf{L}} = [{l}_{1}, \ldots, {l}_{n_{{l}}}] \in \mathbb{R}^{n_{{l}} \times d}$. 
$[\hat g_{n_l+1}^i, \ldots, \hat g_{n_{\rm{max}}-1}^i]$ is used to pad ${\hat{\mathbf{g}}_{i}}$ to length $n_{\rm{max}}$ (by default $n_{\rm{max}}$=256) and the last one $\hat g_{n_{\rm{max}}}^i$ stands for the logit of "no-object".
With the output logits, we define the preference score ${\hat s}_{i} \in \mathbf{S}_{\rm{pred}}$ of each predicted object as:
\begin{equation}
\label{preference_score}
{\hat s}_{i} = 1 - \frac{\exp \left(\hat g_{n_{\rm{max}}}^i\right)}{\sum_{j=1}^{n_{\rm{max}}} \exp \left(\hat g_{j}^i\right)}.
\end{equation}

During training, as in DETR \cite{carion2020end}, a bipartite matching is computed between $n_{\rm{pred}}$ predicted objects $\mathbf{O_{\rm{pred}}}$ and ground truth objects $\mathbf{O_{\rm{gt}}}$ with the Hungarian algorithm \cite{kuhn1955hungarian}.
The matched object predictions are supervised with L1 loss and Generalized Intersection over Union (GIoU) loss \cite{rezatofighi2019generalized} for localization while Dice/F-1 loss \cite{milletari2016v} and Focal cross-entropy loss \cite{lin2017focal} for segmentation.
We also adopt the soft-token prediction loss and the contrastive alignment loss used in MDETR \cite{kamath2021mdetr}.
But different from them, we do not use a single noun or pronoun as the ground truth token span for a matched object prediction.
Instead, we use the whole verb-pronoun description as token span such that the network can understand the verbs rather than noun/pronoun only.
The total loss for TOIST is:
\begin{equation}
\label{loss_TOIST}
\mathcal{L}_{\rm{TOIST}} = \lambda_1 \mathcal{L}_{\rm{l1}} + \lambda_2 \mathcal{L}_{\rm{giou}} + \lambda_3 \mathcal{L}_{\rm{dice}} + \lambda_4 \mathcal{L}_{\rm{cross}} + \lambda_5 \mathcal{L}_{\rm{token}} + \lambda_6 \mathcal{L}_{\rm{align}},
\end{equation}
where $\lambda_1$\textasciitilde$\lambda_6$ are the weights of losses.

\subsection{Noun-Pronoun Distillation}
\label{n-pn-dis}
% To utilize pre-trained noun referring expression comprehension models, we propose a novel noun-pronoun distillation framework.
% Two TOIST models are trained simultaneously.
% The teacher model (Fig.\ref{fig:main} top) and the student model (Fig.\ref{fig:main} bottom) take as input verb-noun description and verb-pronoun description, respectively.
% A clustering method is used to distill knowledge from noun to pronoun, with the help of a memory bank and a tailored cluster loss (Fig.\ref{fig:main} middle left).
% As the preference scores $\mathbf{S}_{\rm{pred}}$ is obtained from logits $\mathbf{G}_{\rm{pred}}$, a soft binary target loss calculated on $\mathbf{G}_{\rm{pred}}$ is proposed to distill knowledge of preference (Fig.\ref{fig:main} middle right).

\begin{wraptable}{r}{0.5\textwidth}
\vspace{-1.5em}
\centering
\caption{TOIST quantitative results under several different settings related to text.}
\resizebox{0.5\textwidth}{!}{
\begin{tabular}{@{}l|ll@{}}
\toprule
{Text related settings}                            & \multicolumn{1}{c}{$\rm{mAP}^{\rm{box}}$} & \multicolumn{1}{c}{$\rm{mAP}^{\rm{mask}}$} \\ \midrule
verb-pronoun input                     & 41.3                       & 35.2                          \\
verb-noun input                       & 53.1 (+11.8)               & 47.2 (+12.0)                  \\ \midrule
replace ${l_{\rm{pron}}}$ with ${l_{\rm{noun}}}$ & 43.7 (\textcolor[RGB]{52,168,83}{+2.4})                       & 37.3 (\textcolor[RGB]{52,168,83}{+2.1})                         \\
replace ${l_{\rm{pron}}^{\rm{tr}}}$ with ${l_{\rm{noun}}^{\rm{tr}}}$ & 41.9 (\textcolor[RGB]{52,168,83}{+0.6})                      & 35.6 (\textcolor[RGB]{52,168,83}{+0.4})                         \\
noun-pronoun distillation          & \textbf{44.1 (\textcolor[RGB]{52,168,83}{+2.8})}                       & \textbf{39.0 (\textcolor[RGB]{52,168,83}{+3.8})}                          \\ \bottomrule
\end{tabular}}
\vspace{-1em}
\label{table:toist_compare}
\end{wraptable}

\textbf{Motivation.} As mentioned above, there are two possible input forms. Due to the privileged information of target names (nouns), TOIST using verb-noun input out-performs its counterpart by +11.8\% and +12.0\% on $\rm{mAP}^{\rm{box}}$ and $\rm{mAP}^{\rm{mask}}$, as demonstrated in Table.~\ref{table:toist_compare}. We use two preliminary experiments as a motivation: directly replacing the pronoun feature ${l_{\rm{pron}}}$ or ${l_{\rm{pron}}^{\rm{tr}}}$ in the verb-pronoun model with the corresponding noun feature ${l_{\rm{noun}}}$ or ${l_{\rm{noun}}^{\rm{tr}}}$ in the verb-noun model can boost performance (see Table \ref{table:toist_compare}).
However, during inference, the noun of the ground truth object is unavailable. Thus we believe a properly designed noun-pronoun distillation framework can leverage rich knowledge from the verb-noun model without violating the noun-agnostic constraint.

\textbf{Distillation Framework Overview.} Two TOIST models are trained simultaneously.
The teacher (Fig.\ref{fig:main} top) and the student (Fig.\ref{fig:main} bottom) take as input verb-noun and verb-pronoun descriptions, respectively.
A clustering distillation method with a memory bank and a tailored cluster loss is used to distill privileged object-centric knowledge from noun to pronoun (Fig.\ref{fig:main} middle left). Besides, we also use a soft binary target loss imposed on $\mathbf{G}_{\rm{pred}}$ to distill preference knowledge (Fig.\ref{fig:main} middle right), in which $\mathbf{G}_{\rm{pred}}$ are logits used to calculate
preference scores $\mathbf{S}_{\rm{pred}}$.

% \usepackage{booktabs}

%\textbf{Teacher and Student TOIST Models.}
%The teacher model takes a verb-noun description of a specific task like \emph{dig hole with skateboard} as input $\mathbf{X}_l$, where the noun is the ground truth object category.
%The student model uses a verb-pronoun phrase as text input $\mathbf{X}_l$, where the pronoun like \emph{something} works as a proxy.
%Both the two models are initialized with pre-trained noun referring expression comprehension models.
%Because the teacher model specifies the class name of desired object(s) in the text input, it can efficiently find the object(s) in the image with privileged knowledge.

\textbf{Clustering Distillation.}
Since one task can be afforded by objects of many different categories, we build a text feature memory bank to store noun features, with which a prototype can be selected and used to replace pronoun feature and distill knowledge.
We term this process as clustering distillation.
Specifically, we use ${l_{\rm{pron}}^{\rm{tr}}}$ and ${l_{\rm{noun}}^{\rm{tr}}}$ instead of ${l_{\rm{pron}}}$ and ${l_{\rm{noun}}}$ for this process.
The reason is that the former ones are conditioned on the image input and verb tokens of the task by self-attention layers, and thus it is meaningful to select a cluster center that suits the image and the task input.

\textbf{Memory and Selector.}
The size of the memory bank is $n_{\rm{task}} \times n_{\rm{mem}} \times d$.
% It consists of $n_{\rm{task}}$ first-in-first-out queues of length $n_{\rm{mem}}$ for $n_{\rm{task}}$ tasks.
% During training, for each sample from task $j$, we update the $j$-th queue $\mathbf{L}_{\rm{mem}}^j = {[{l_{\rm{1}}^{j}}, \ldots, {l_{{n_{\rm{mem}}}}^{j}} ]}$ with noun feature ${l_{\rm{noun}}^{\rm{tr}}}$ generated by the teacher model.
% The shifted queue ${[{l_{{2}}^{j}}, \ldots, {l_{{n_{\rm{mem}}}}^{j}}, {l_{\rm{noun}}^{\rm{tr}}}]}$ are clustered with the K-means clustering method, leading to $\rm{K}$ cluster centers $\mathbf{L}_{\rm{c}}^j = \{{l_{c_1}^{j}}, \ldots, {l_{c_{\rm{K}}}^{j}}\}$.
\textcolor[RGB]{0,0,0}{
It consists of $n_{\rm{task}}$ queues of length $n_{\rm{mem}}$ for $n_{\rm{task}}$ tasks.
During training, for each sample from task $j$, we update the $j$-th queue $\mathbf{L}_{\rm{mem}}^j = {[{l_{\rm{1}}^{j}}, \ldots, {l_{{n_{\rm{mem}}}}^{j}} ]}$ by adding the noun feature ${l_{\rm{noun}}^{\rm{tr}}}$ generated by the teacher model and removing the existing one closest to ${l_{\rm{noun}}^{\rm{tr}}}$.
The updated queue is clustered with the K-means clustering method, leading to $\rm{K}$ cluster centers $\mathbf{L}_{\rm{c}}^j = \{{l_{c_1}^{j}}, \ldots, {l_{c_{\rm{K}}}^{j}}\}$.
}
Then the student model uses a cluster selector, which is implemented as the nearest neighbor classifier, to select a prototype ${l_{c_s}^{j}} \in \mathbf{L}_{\rm{c}}^j$ according to the pronoun feature ${l_{\rm{pron}}^{\rm{tr}}}$ and replace ${l_{\rm{pron}}^{\rm{tr}}}$ with ${l_{c_s}^{j}}$.
Concatenating other tokens and the selected prototype together, the output of student transformer encoder $[{v}_{s_1}^{\rm{tr}}, \ldots, {v}_{s_n}^{\rm{tr}}, {l}_{s_1}^{\rm{tr}}, \ldots, {l_{\rm{pron}}^{\rm{tr}}}, \ldots,{l}_{s_{n_l}}^{\rm{tr}}]$ is modified into $[{v}_{s_1}^{\rm{tr}}, \ldots, {v}_{s_n}^{\rm{tr}}, {l}_{s_1}^{\rm{tr}}, \ldots, {l_{c_s}^{j}}, \ldots, {l}_{s_{n_l}}^{\rm{tr}}]$ and fed into the transformer decoder.
%As such, the noun referring knowledge is transferred into the pronoun proxy and then the verb description of the task.
To distill knowledge to the student transformer encoder, we define cluster loss as:
\begin{equation}
\mathcal{L}_{\rm{cluster}} = \| {l_{\rm{pron}}^{\rm{tr}}} - {l_{c_s}^{j}} \|_2,
\end{equation}
with which the privileged object-centric knowledge is distilled from clustered noun features to pronoun feature and further to the student TOIST encoder.

\textbf{Preference Distillation.}
We use a soft binary target loss to distill preference knowledge from teacher to student.
For an object query, we first define binary query probabilities of being positive-query or negative-query, which denotes whether or not the query is matched to a ground truth object target, as $\mathbf{p} = [p^{\rm{pos}}, p^{\rm{neg}}] \in \mathbb{R}^{1\times2}$. The probabilities can be calculated by the softmax function:
\begin{equation}
\label{pos_neg}
p^{\rm{pos}}=\frac{\sum_{j=1}^{n_{\rm{max}}-1} \exp \left(\hat g_{j}\right)}{\sum_{j=1}^{n_{\rm{max}}} \exp \left(\hat g_{j}\right)}, 
p^{\rm{neg}}=\frac{\exp \left(\hat g_{n_{\rm{max}}}\right)}{\sum_{j=1}^{n_{\rm{max}}} \exp \left(\hat g_{j}\right)},
\end{equation}
where $\hat g_j$ and $\hat g_{n_{\rm{max}}}$ represent the logits corresponding to $i$-th text token and "no-object" token, respectively. 
For all the object queries in teacher and student, the probability sequences are denoted by $\mathbf{P}_t = [\mathbf{p}_{t_1}, \ldots, \mathbf{p}_{t_{n_{\rm{pred}}}}]$ and $\mathbf{P}_s = [\mathbf{p}_{s_1}, \ldots, \mathbf{p}_{s_{n_{\rm{pred}}}}]$. 
Then we use the Hungarian algorithm to find a bipartite matching between the two sequences of object queries with $\mathbf{P}_t$ and $\mathbf{P}_s$. Formally, we search for a permutation of ${n_{\rm{pred}}}$ elements $\sigma \in \mathfrak{S}_{{n_{\rm{pred}}}}$ which minimizes the matching loss:
\begin{equation}
\hat{\sigma}=\mathop{\arg \min }\limits_{\sigma \in \mathfrak{S}_{{n_{\rm{pred}}}}} \sum_{i}^{{n_{\rm{pred}}}} \mathcal{L}_{\rm{match}}({y}_{t_i}, {y}_{s_{\sigma(i)}}),
\end{equation}
where $y_{t_i} = (\hat b_{t_i} \mathbf{p}_{t_i})$ and $\hat b_{t_i}$ is the predicted bounding box.  $\mathcal{L}_{\rm{match}}$ is a linear combination of box prediction losses (L1 \& GIoU) and KL-Divergence. The KL-Divergence $\mathcal{L}_{\rm{KL}}$ can be written as:
\begin{equation}
\mathcal{L}_{\rm{KL}}(\mathbf{p}_{t_i}, \mathbf{p}_{s_{\sigma(i)}}) =\mathbf{KL}\left(\mathbf{p}_{t_i} \| \mathbf{p}_{s_{\sigma(i)}}\right)
= p_{t_i}^{\rm{pos}} \log \left(\frac {p_{t_i}^{\rm{pos}}} {p_{s_{\sigma(i)}}^{\rm{pos}}}\right)  +  p_{t_i}^{\rm{neg}} \log \left(\frac {p_{t_i}^{\rm{neg}}} {p_{s_{\sigma(i)}}^{\rm{neg}}}\right).
\end{equation}
With the optimal permutation $\hat \sigma$, we define the soft binary target loss as:
\begin{equation}
\mathcal{L}_{\rm{binary}} = \sum_{i}^{{n_{\rm{pred}}}} \mathcal{L}_{\rm{KL}}(\mathbf{p}_{t_i}, \mathbf{p}_{s_{\hat \sigma(i)}}).
\end{equation}
It makes the binary query probabilities of the student model similar to the matched ones of the teacher model.
And because the preference score ${\hat s}$ (Eq.\ref{preference_score}) is defined in the same way as the probability $p^{\rm{pos}}$ (Eq.\ref{pos_neg}), the preference knowledge is distilled from teacher to student as the loss decreases.

\textbf{Summary.} The final training loss function for TOIST with noun-pronoun distillation is:
\begin{equation}
\mathcal{L}_{\rm{TOIST-NP}} = \mathcal{L}_{\rm{TOIST}}^t + \mathcal{L}_{\rm{TOIST}}^s + \lambda_7 \mathcal{L}_{\rm{cluster}}^s + \lambda_8 \mathcal{L}_{\rm{binary}}^s,
\end{equation}
where $\lambda_7$ and $\lambda_8$ are the weights of losses. $\mathcal{L}_{\rm{TOIST}}^t$ and $\mathcal{L}_{\rm{TOIST}}^s$ are separate TOIST loss terms defined by Eq.\ref{loss_TOIST} for teacher and student, respectively. Note that the cluster loss $\mathcal{L}_{\rm{cluster}}^s$ and the soft binary target loss $\mathcal{L}_{\rm{binary}}^s$ are only used for supervising the student model.

As a reminder, during inference, we only use the student TOIST model and the fixed memory bank to find the most suitable objects, without violating the noun-agnostic constraint.

\section{Experiments}

\begin{table}[t]
\centering
\caption{Comparison of the proposed method to SOTA baselines on the extended COCO-Tasks dataset. The methods with tags $\dagger$ and $\ddagger$ are used to calculate mAP margins. TOIST with noun-pronoun distillation achieves best results: +10.9\% and +6.6\% mAP on detection and segmentation.}
\resizebox{0.8\textwidth}{!}{
\begin{tabular}{@{}lc|lll@{}}
\toprule
\multicolumn{2}{c|}{{Object Detection}}         & \multicolumn{3}{c}{{Instance Segmentation}}     \\ \midrule
\multicolumn{1}{c}{{Method}} & {$\rm{mAP}^{\rm{box}}$} & \multicolumn{1}{c}{{Method}} & \multicolumn{1}{c}{$\rm{mAP}^{\rm{box}}$} & \multicolumn{1}{c}{$\rm{mAP}^{\rm{mask}}$} \\ \midrule
Faster-RCNN                         & 20.6             & Mask-RCNN                           & 23.4 & 20.0              \\
Faster-RCNN + pick best             & 14.1             & Mask-RCNN + pick best         & 18.8 & 16.8              \\
Faster-RCNN + ranker                & 9.1              & Mask-RCNN + ranker                  & 10.6 & 9.3              \\
Faster-RCNN + classifier            & 28.8             & Mask-RCNN + classifier          & 33.7 & 29.2              \\
Faster-RCNN + GGNN                  & 32.6             & Mask-RCNN + GGNN$\hspace{0em}^{\ddagger}$                    & 37.4 & 32.4              \\
\cline{3-5}
Yolo + classifier                   & 29.1             & \textbf{TOIST}              & {41.3 (\textcolor[RGB]{52,168,83}{+8.1})} & {35.2 (\textcolor[RGB]{52,168,83}{+2.8})}        \\
Yolo + GGNN$\hspace{0em}^{\dagger}$                         & 33.2             & \textbf{TOIST w/ distillation}   & \textbf{44.1 (\textcolor[RGB]{52,168,83}{+10.9})} & \textbf{39.0 (\textcolor[RGB]{52,168,83}{+6.6})}      \\ \bottomrule
\end{tabular}}
\label{table:main_result}
% \vspace{-1em}
\end{table}

% \textbf{Dataset.}
% We perform experiments on the COCO-Tasks dataset \cite{sawatzky2019object} which re-annotates the COCO dataset \cite{lin2014microsoft} with preference-aware affordance labels.
% This is the only dataset that provides preference-aware affordance labels for an image and a specific task like \emph{step on something}.
% Even though there are other datasets that provide affordance of objects, they only focus on low-level affordance (like \emph{grasp}) and do not care about preference, such as ADE-Affordance \cite{chuang2018learning} and IIT-AFF \cite{nguyen2017object}.
% The COCO-Tasks dataset contains 14 tasks.
% For each task, there are 3600 train images and 900 test images. 
% In an image, the most suitable objects (one or more) for solving the task are selected and their bounding boxes are taken as ground truth labels for detection. 
% %The number of selected objects in an image varies from zero to a dozen. 
% %For each task, the total number of selected objects varies between 1,105 and 9,870 and the number of different object categories varies between 6 and 30.
% Using existing COCO masks, we further extend the dataset to an upgraded instance segmentation version.
% We use the AP@0.5 metric for both detection and segmentation, where predicted preference scores $\mathbf{S}_{\rm{pred}}$ are used to rank objects.
% Averaging the AP@0.5 values of all tasks leads to mAP@0.5.
\textbf{Dataset.}
We conduct experiments on the COCO-Tasks dataset \cite{sawatzky2019object} which re-annotates the COCO dataset \cite{lin2014microsoft} with preference-aware affordance labels.
This is the only dataset that involves instance-level preference in affordance.
Though there are other datasets for affordance detection, they neither distinguish between instances nor involve preference, such as ADE-Affordance \cite{chuang2018learning} and IIT-AFF \cite{nguyen2017object}.
The COCO-Tasks dataset contains 14 tasks.
For each task, there are 3600 train images and 900 test images. 
In each image, the boxes of preferred objects (one or more) are taken as ground truth labels for detection. 
%The number of selected objects in an image varies from zero to a dozen. 
%For each task, the total number of selected objects varies between 1,105 and 9,870 and the number of different object categories varies between 6 and 30.
Using existing COCO masks, we extend the dataset to an instance segmentation version.
\textcolor[RGB]{0,0,0}{
In Appendix~\ref{appendix:dataset} and ~\ref{appendix:analysis}, we present more dataset details and show its diversity.
}
\textbf{Metric.} We use the AP@0.5 metric for both detection and segmentation, where predicted preference scores $\mathbf{S}_{\rm{pred}}$ are used to rank objects.
Averaging the AP@0.5 values of all tasks leads to mAP@0.5.
% More details for implementation and dataset can be found in Appendix~\ref{appendix:implementation_details} and ~\ref{appendix:dataset}.
\textcolor[RGB]{0,0,0}{
The implementation details for TOIST and distillation can be found in Appendix~\ref{appendix:implementation_details}.
}

\begin{figure}[b]
\vspace{-1em}
\begin{center}
\includegraphics[width=0.8\textwidth]{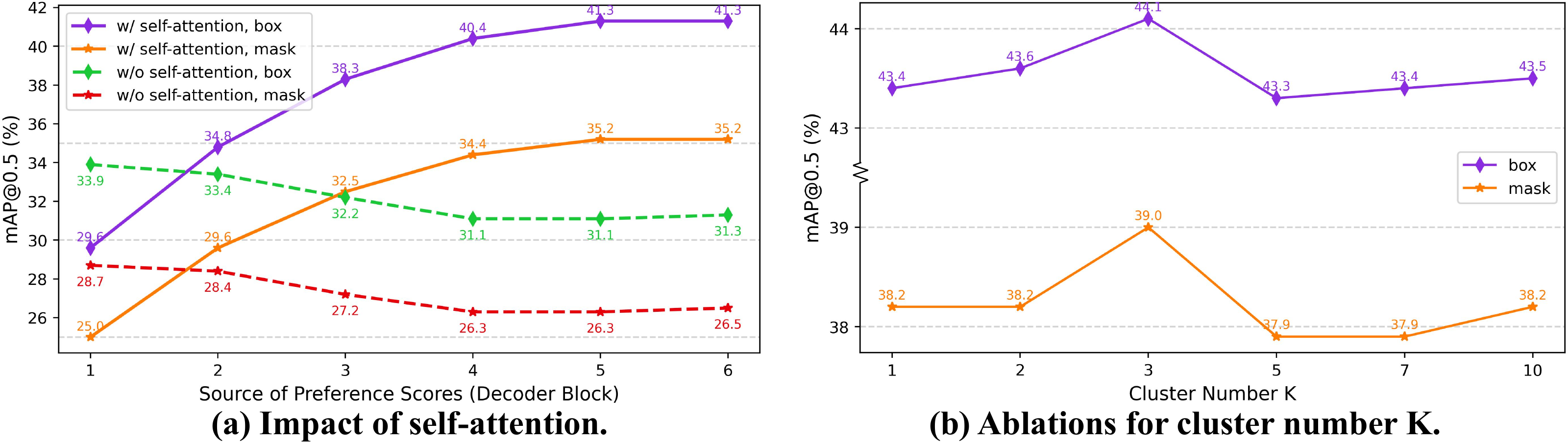}
\end{center}
\vspace{-1em}
\caption{Experiments to study the impact of (a) self-attention and (b) cluster number.}
\label{fig:curve}
\vspace{-1em}
\end{figure}

\subsection{Comparisons with State-of-the-art Methods}
Table \ref{table:main_result} shows that TOIST with noun-pronoun distillation achieves state-of-the-art results compared to existing methods on COCO-Tasks.
For object detection, we use the results reported by \cite{sawatzky2019object} as baselines.
For instance segmentation, following the same experiment settings of \cite{sawatzky2019object}, we build new baselines using Mask-RCNN \cite{he2017mask}.
The methods in the first row treat the problem as a standard detection or segmentation task.
All other baselines use two-stage pipelines, in which objects are firstly detected or segmented then ranked. The proposed one-stage method achieves 41.3\% $\rm{mAP^{box}}$ and 35.2\% $\rm{mAP^{mask}}$, which are +8.1\% and +2.8\% better than the previous best results (\emph{Yolo+GGNN} and \emph{Mask-RCNN+GGNN}).
Noun-pronoun distillation further boosts the performance of TOIST to 44.1\% (+10.9\%) $\rm{mAP^{box}}$ and 39.0\% (+6.6\%) $\rm{mAP^{mask}}$. 
\textcolor[RGB]{0,0,0}{
Our method also out-performs another baseline with the same backbone, as shown in Table \ref{table:mdetr_ggnn} of Appendix~\ref{appendix:stronger_baseline}.
}
These results demonstrate the effectiveness of the proposed method for the new problem of task oriented instance segmentation.
\textcolor[RGB]{0,0,0}{
Per-task quantitative results and precision-recall curves are provided in Appendix~\ref{appendix:per_task_results} and ~\ref{appendix:prcurves}.
}

% \begin{table}
% \centering
% \caption{.}
% \begin{tabular}{@{}lc|lc@{}}
% \toprule
% \multicolumn{2}{c|}{{Object Detection}}                 & \multicolumn{2}{c}{{Instance Segmentation}}     \\ \midrule
% \multicolumn{1}{c}{{Method}} & {$\rm{mAP}^{\rm{box}}$}  & \multicolumn{1}{c}{{Method}} & {$\rm{mAP}^{\rm{mask}}$} \\ \midrule
% Faster-RCNN                         & 20.6              & Mask-RCNN                      & 20.0              \\
% Faster-RCNN + pick best             & 14.1              & Mask-RCNN + pick best          & 16.8              \\
% Faster-RCNN + ranker                & 9.1               & Mask-RCNN + ranker             & 9.3              \\
% Faster-RCNN + classifier            & 28.8              & Mask-RCNN + classifier         & 29.2              \\
% Faster-RCNN + GGNN                  & 32.6              & Mask-RCNN + GGNN               & 32.4              \\
% Yolo + classifier                   & 29.1              &&\\
% Yolo + GGNN                         & 33.2              &&\\\midrule
% \textbf{TOIST}                      & {\ul 41.3 (\textcolor[RGB]{52,168,83}{+8.1})}   & \textbf{TOIST}   & {\ul 35.2 (\textcolor[RGB]{52,168,83}{+2.8})}        \\
% \textbf{TOIST + distillation}       & \textbf{43.9 (\textcolor[RGB]{52,168,83}{+10.7})}     & \textbf{TOIST + distillation}  & \textbf{38.8 (\textcolor[RGB]{52,168,83}{+6.4})} \\
% \bottomrule
% \end{tabular}
% \label{table:main_result}
% \end{table}

\subsection{Preference Modeling with Self-Attention}
%We study the problem of task oriented instance segmentation with a transformer-based model and leverage its self-attention to model preference.
\textbf{Protocol.} Our design principle is that the self-attention layers in transformers can naturally model preference.
But is this really the case? 
%Fig.\ref{fig:curve} (a) shows the effect of self-attention on the performance of the proposed method.
Fig.\ref{fig:curve} (a) shows its effect.
Two plain TOIST models are trained separately, with the only difference being that one model does not contain self-attention operators in the decoder.
Note that the removal of self-attention does not impact the number of parameters.
The mAP metric is impacted by two kinds of errors: inaccurate box/mask localization or improper preference. To analyze the preference scores alone, for all object queries, we use the boxes and masks predicted by the last block of TOIST decoder, which are arguably the most accurate.
We use the corresponding preference scores predicted by each block to calculate mAP values.

\textbf{Interpretation.} For the TOIST with self-attention, the performance is gradually boosted as the source of preference scores becomes deeper: from 29.6\% $\rm{mAP^{box}}$ and 25.0\% $\rm{mAP^{mask}}$ to 41.3\% and 35.2\%.
For the one without self-attention, the preference scores from the first block lead to the best performance: 33.9\% $\rm{mAP^{box}}$ and 28.7\% $\rm{mAP^{mask}}$, which is -7.5\% and -6.5\% lower than another TOIST.
The results demonstrate that the self-attention in TOIST decoder models pair-wise relative preference between object candidates.
As the decoder deepens, the preference relationship between object candidates is gradually extracted by self-attention.
%Without it, the decoder makes the relationship unclear and cannot improve the performance.
% Without self attention, the decoder with the same number of parameters makes the relationship unclear and cannot improve the performance.

% \begin{wrapfigure}{r}{0.5\textwidth}
% \vspace{-2em}
% \begin{center}
% \includegraphics[width=0.5\textwidth]{curve.pdf}
% \end{center}
% \vspace{-1.5em}
% \caption{.}
% \vspace{-1em}
% \label{fig:curve}
% \end{wrapfigure}

\subsection{Effect of Clustering Distillation}

% \usepackage{booktabs}
\begin{wraptable}{r}{0.45\textwidth}
\vspace{-3em}
\centering
\caption{Ablations for distillation settings. CCR, CL and SBTL are short for cluster center replacement, cluster loss and soft binary target loss, respectively.}
\resizebox{0.45\textwidth}{!}{
\begin{tabular}{@{}c|ccc|ll@{}}
\toprule
Index & CCR & CL & SBTL & \multicolumn{1}{c}{$\rm{mAP}^{\rm{box}}$} & \multicolumn{1}{c}{$\rm{mAP}^{\rm{mask}}$}  \\ \midrule
(a) & \texttimes     & \texttimes & \texttimes                                                & 41.3                                                                    & 35.2                                                                       \\
(b) & \texttimes     & \texttimes & \checkmark                                                  & 43.4 (\textcolor[RGB]{52,168,83}{+2.1})                                                                    & 38.0 (\textcolor[RGB]{52,168,83}{+2.8})                                                                       \\
(c) & \texttimes     & \checkmark   & \texttimes                                                & 42.0 (\textcolor[RGB]{52,168,83}{+0.7})                                                                    & 37.1 (\textcolor[RGB]{52,168,83}{+1.9})                                                                       \\
(d) & \texttimes     & \checkmark   & \checkmark                                                  & {43.8 (\textcolor[RGB]{52,168,83}{+2.5})}                                                                   & {38.6 (\textcolor[RGB]{52,168,83}{+3.4})}                                                                      \\
(e) & \checkmark       & \texttimes & \texttimes                                                & 42.0 (\textcolor[RGB]{52,168,83}{+0.7})                                                                    & 37.0 (\textcolor[RGB]{52,168,83}{+1.8})                                                                       \\
(f) & \checkmark       & \texttimes & \checkmark                                                  & 42.3 (\textcolor[RGB]{52,168,83}{+1.0})                                                                    & 37.3 (\textcolor[RGB]{52,168,83}{+2.1})                                                                       \\
(g) & \checkmark       & \checkmark   & \texttimes                                                & 42.3 (\textcolor[RGB]{52,168,83}{+1.0})                                                                    & 37.5 (\textcolor[RGB]{52,168,83}{+2.3})                                                                       \\
(h) & \checkmark       & \checkmark   & \checkmark                                                  & \textbf{44.1 (\textcolor[RGB]{52,168,83}{+2.8})}                                                                    & \textbf{39.0 (\textcolor[RGB]{52,168,83}{+3.8})}                                                                       \\ \bottomrule
\end{tabular}}
\label{table:abl_loss}
\end{wraptable}

In Table \ref{table:abl_loss}, we show the effects of using cluster loss and replacing pronoun features with cluster centers (noun prototypes). In (c) and (e), leveraging the two components alone brings an increase of +0.7\% $\rm{mAP^{box}}$, +1.9\% $\rm{mAP^{mask}}$ and +0.7\% $\rm{mAP^{box}}$, +1.8\% $\rm{mAP^{mask}}$ over baseline (a) respectively. In (g), the complete clustering distillation leads to a performance improvement of +1.0\% $\rm{mAP^{box}}$ and +2.3\% $\rm{mAP^{mask}}$.
These results show that the clustering distillation method can improve student TOIST and enhance verb referring expression understanding.

% These results show that the cluster center replacement can boost student TOIST by specifying object categories with the noun prototype features stored in the memory bank.
% Cluster loss can distill information of noun referring expression to task-related pronoun, enabling student TOIST encoder to model preference.
% And the complete noun-pronoun distillation setting simultaneously uses both of these aspects to effectively enhance verb referring expression understanding.

\begin{figure*}[b]
\vspace{-1em}
\begin{center}
\includegraphics[width=0.85\textwidth]{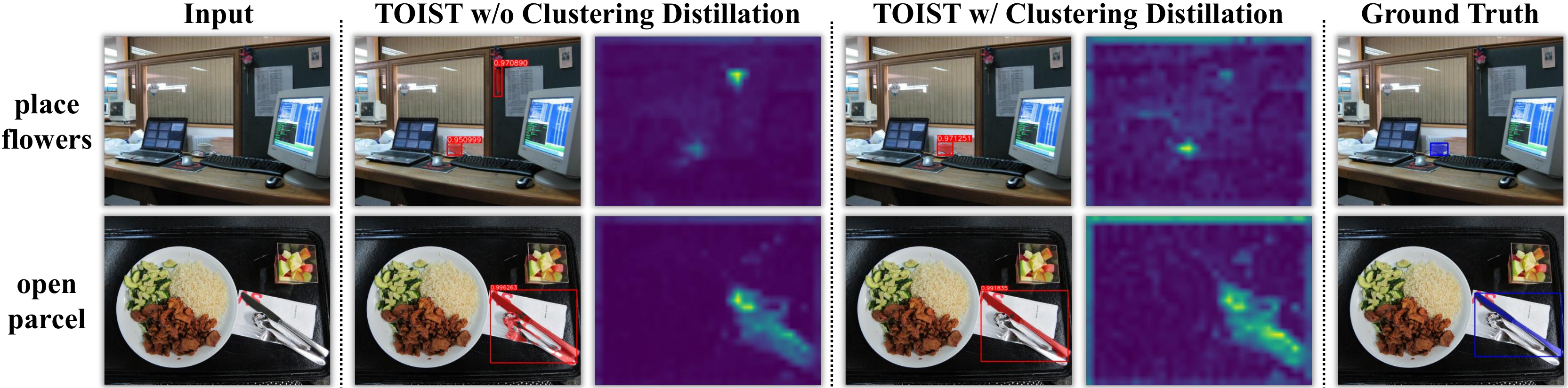}
\end{center}
\vspace{-1em}
\caption{Visualization of the predicted results and attention maps of pronoun tokens.}
\label{fig:cluster_vis}
\vspace{-1.5em}
\end{figure*}

In Fig.\ref{fig:cluster_vis}, we visualize the predicted results (filtered by a preference threshold of 0.9) and the attention maps of pronoun tokens.
In the first row, when there is no clustering distillation, TOIST wrongly prefers the flower to the cup, which is also confirmed by the attention map.
But the TOIST with clustering distillation correctly selects the cup, and the attention on the flower is weakened.
This shows that clustering distillation enables the student TOIST to reduce the ambiguity of verb-pronoun referring expression.
In the second row, the bounding box of the knife is correctly detected by both two models.
However, in the absence of the distillation, extra instance masks are predicted on the spoon and fork within the box.
Instead, with the distillation, the masks predicted by TOIST are concentrated on the knife and the attention is more focused on it.
This demonstrates that in the case of clustering distillation, TOIST can better ground the task into pixels within an object box.

Meanwhile, the fact that predicted masks may be inaccurate even if the box is correct makes it challenging for a robot to accurately grasp the preferred object when performing a specific task.
This proves the importance of extending task oriented object detection to instance segmentation.

\subsection{Effect of Preference Distillation}

\begin{wrapfigure}{r}{0.45\textwidth}
\vspace{-4em}
\begin{center}
\includegraphics[width=0.45\textwidth]{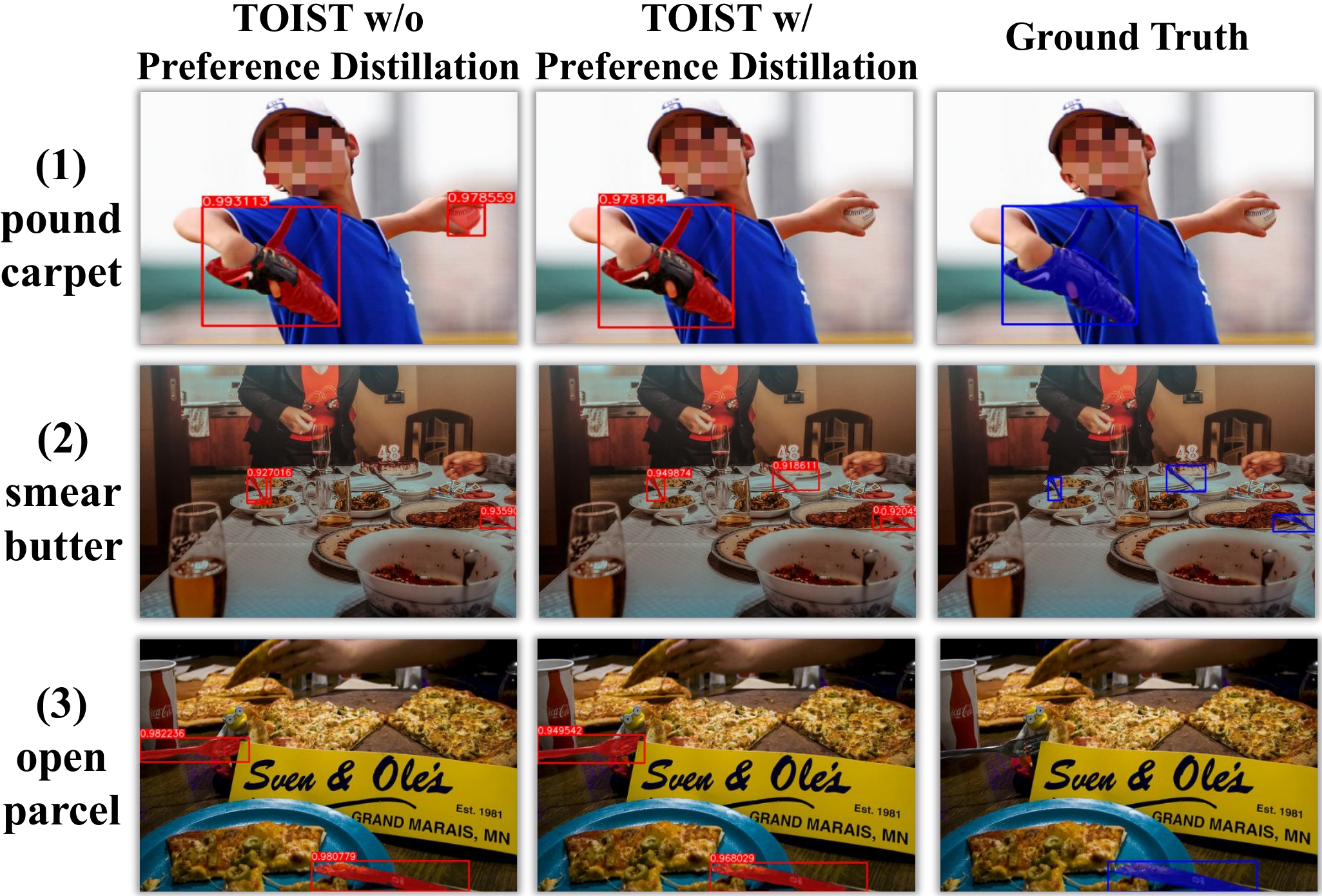}
\end{center}
% \vspace{-1em}
\caption{Examples of three scenarios where preference distillation clearly works.}
\vspace{-1em}
\label{fig:soft_binary}
\end{wrapfigure}

In Table \ref{table:abl_loss} (b), preference distillation with soft binary target loss achieves +2.1\% $\rm{mAP^{box}}$ and +2.8\% $\rm{mAP^{mask}}$ higher results than baseline (a). 
This loss acts on the preference probabilities of each object candidate in student TOIST. And the probabilities are used as scores to sort the object candidates for the calculation of mAP values. Therefore, the result of Table \ref{table:abl_loss} (b) strongly supports that the preference information is distilled to the student TOIST.

A simple taxonomy differs three scenarios where preference distillation works.
As shown in Fig.\ref{fig:soft_binary}, the predicted results (filtered by a preference threshold of 0.9) of the TOIST models w/ or w/o preference distillation are compared.
(1) Preference distillation makes the preference score of the false positive object (the baseball in the left picture) lower than the threshold.
(2) The preference score of the false negative object (the spoon in the middle) is raised above the threshold with the distillation.
(3) When there is no distillation, the false positive object (fork) scores higher than the true positive object (knife) ($0.9822>0.9808$).
Although the distillation fails to lower the preference score of the false positive object below the threshold, its score is updated to be lower than the true positive one ($0.9495<0.9680$).
These specific results demonstrate that the information of noun referring expression is distilled to the noun-agnostic student model in the form of preference scores.
%Thus, the student model's ability to understand verbs is improved.

\subsection{Ablation Study and Qualitative Results}
\textcolor[RGB]{0,0,0} {
\textbf{Distillation Methods.}
Instead of minimizing the distance between $l_{\rm{pron}}^{\rm{tr}}$ and $l_{c_s}^j$, a straightforward way is to directly minimize the distance between $l_{\rm{pron}}^{\rm{tr}}$ and $l_{\rm{noun}}^{\rm{tr}}$. As shown in Table \ref{table:direct_distill}, this simplified method does not work well, which prompts us to develop the distillation framework.
}

\begin{table}[htbp]
	\centering
	\begin{minipage}[t]{0.45\linewidth}
		\small\centering
        \caption{Different distillation methods.}
%  		\vspace{-0.3cm}
		\resizebox{1\textwidth}{!}{
            \begin{tabular}{l|ll}
            \toprule
            Method                                                              & \multicolumn{1}{c}{$\rm{mAP}^{box}$} & \multicolumn{1}{c}{$\rm{mAP}^{mask}$} \\ \midrule
            TOIST                                                               & 41.3                                 & 35.2                                  \\
            distill from $l_{c_s}^j$ to $l_{\rm{pron}}^{\rm{tr}}$              & \textbf{44.1 (\textcolor[RGB]{52,168,83}{+2.8})}                  & \textbf{39.0 (\textcolor[RGB]{52,168,83}{+3.8})}                   \\
            distill from $l_{\rm{noun}}^{\rm{tr}}$ to $l_{\rm{pron}}^{\rm{tr}}$ & {\ul 41.9 (\textcolor[RGB]{52,168,83}{+0.6})}                     & {\ul 36.0 (\textcolor[RGB]{52,168,83}{+0.8})}                      \\ \bottomrule
            \end{tabular}
        }
	    \label{table:direct_distill}
	\end{minipage}\hspace{1em}
	\begin{minipage}[t]{0.45\linewidth}
		\small\centering
        \caption{Results without pre-training.}
%  		\vspace{-0.3cm}
		\resizebox{1\textwidth}{!}{
            \begin{tabular}{l|ll}
            \toprule
            Method                    & \multicolumn{1}{c}{$\rm{mAP}^{box}$}                 & \multicolumn{1}{c}{$\rm{mAP}^{mask}$}                       \\ \midrule
            verb-pronoun input             & 3.65                                                 & 5.74                                                        \\
            verb-noun input                & 11.19 & 12.67                                                       \\
            noun-pronoun distillation & 7.43 (\textcolor[RGB]{52,168,83}{+3.78})                                          & 11.28 (\textcolor[RGB]{52,168,83}{+5.54}) \\ \bottomrule
            \end{tabular}
        }
		\label{table:without_pretraining}
 	\end{minipage}
\end{table}

\textbf{Interaction of the Two Distillation Components.}
In Table \ref{table:abl_loss} (d) and (f), we show the effects of cluster loss or cluster center replacement together with soft binary target loss. (d) achieves +2.5\% $\rm{mAP^{box}}$ and +3.4\% $\rm{mAP^{mask}}$ improvement, which demonstrates the two distillation losses collaborate well.
(f) only achieves +1.0\% $\rm{mAP^{box}}$ and +2.1\% $\rm{mAP^{mask}}$ improvement, slightly higher than (e) (using cluster loss only) but lower than (b) (using soft binary target loss only).
This shows that preference distillation effectively improves object preference modeling.
But solely replacing pronoun features to indicate target objects weakens the effect of preference distillation.

% \usepackage{booktabs}
% \usepackage{multirow}
%{\begin{tabular}[c]{@{}c@{}}TOIST\\ (verb + pronoun)\end{tabular}}
\begin{wraptable}{r}{0.45\textwidth}
% \vspace{-2em}
\centering
\caption{Ablations for pronoun input.}
\resizebox{0.45\textwidth}{!}{
\begin{tabular}{@{}cc|cc@{}}
\toprule
{Method}                                                                              & {Pronoun} & {$\rm{mAP}^{\rm{box}}$} & {$\rm{mAP}^{\rm{mask}}$} \\ \midrule
\multirow{4}{*}{TOIST}            & something        & 41.3                           & 35.2                              \\
                                                                                             & it               & 41.3                           & 35.2                              \\
                                                                                             & them             & 41.4                           & 35.0                              \\
                                                                                             & abcd             & 39.0                           & 33.2                              \\ \midrule
\multirow{4}{*}{\begin{tabular}[c]{@{}c@{}}TOIST\\ w/ distillation\end{tabular}} & something        & 44.1                           & 39.0                               \\
                                                                                             & it               & 43.8                           & 38.4                              \\
                                                                                             & them             & 43.8                           & 38.1                              \\
                                                                                             & abcd             & 42.8                           & 37.4                              \\ \bottomrule
\end{tabular}}
\label{table:abl_pronoun}
\vspace{-2em}
\end{wraptable}

\textbf{Ablations for Cluster Number $\rm{K}$.}
Fig.\ref{fig:curve} (b) shows the ablations for cluster number $\rm{K}$.
We perform distillation experiments on different $\rm{K}$ values between 1 to 10 because increasing $\rm{K}$ to an even higher value makes the clustering task more difficult.
All of the experiments yield better results than the plain TOIST (41.3\% $\rm{mAP^{box}}$, 35.2\% $\rm{mAP^{mask}}$) and $\rm{K}=3$ works the best.
This demonstrates that a modest $\rm{K}$ can better cluster the information of noun features and distill it to the student TOIST.

\textbf{Ablations for Pronoun Input.}
Table \ref{table:abl_pronoun} shows the results of TOIST with different pronoun input.
In the plain TOIST and TOIST with distillation, the usage of \emph{something}, \emph{it} or \emph{them} leads to similar results, while a meaningless string \emph{abcd} yields less improvement.
Nevertheless, the proposed distillation framework can still work well in the last case, which demonstrates the robustness of our method.

\textcolor[RGB]{0,0,0} {
\textbf{Results without Pre-training.}
In our architecture, the pre-trained noun referring expression comprehension models are leveraged. 
To investigate whether the noun-pronoun distillation framework is a standalone technical contribution, we conduct experiments without pre-training.
The models are trained from scratch on the COCO-Tasks dataset and the results are shown in Table \ref{table:without_pretraining}, which demonstrates that the proposed distillation can still improve performance even without pre-training.
}

\textcolor[RGB]{0,0,0} {
\textbf{Ablations for Task Number.}
Table \ref{table:task_number} shows the ablation study of different task numbers, in which the first row corresponds to the plain TOIST without distillation and the others show the results with distillation under different $n_{\rm{task}}$.
The results demonstrate our proposed distillation works for different $n_{\rm{task}}$, even if $n_{\rm{task}}$ = 1.
And overall, smaller $n_{\rm{task}}$ leads to better performance. 
We attribute this to the reduced problem complexity due to the less interaction between different tasks, which makes it easier to improve the ability of the model to understand verbs through noun-pronoun distillation.
}
% \usepackage[normalem]{ulem}
\begin{table}
\centering
\caption{Ablations for the task number $n_{\rm{task}}$ on task oriented object detection.}
\resizebox{0.8\textwidth}{!}{
\begin{tabular}{l|ccccc}
\hline
\diagbox{Method}{Task}                   & \begin{tabular}[c]{@{}c@{}}step on\\ something\end{tabular} & \begin{tabular}[c]{@{}c@{}}sit\\ comfortably\end{tabular} & \begin{tabular}[c]{@{}c@{}}place\\ flowers\end{tabular} & \begin{tabular}[c]{@{}c@{}}get potatoes\\ out of fire\end{tabular} & \begin{tabular}[c]{@{}c@{}}water\\ plant\end{tabular} \\ \hline
TOIST w/o dis            & 44.0                                                        & 39.5                                                      & 46.7                                                    & 43.1                                                               & 53.6                                                  \\
dis $n_{\rm{task}}$ = 14 & 46.2 (\textcolor[RGB]{52,168,83}{+2.2})                                                 & 39.6 (\textcolor[RGB]{52,168,83}{+0.1})                                               & 49.9 (\textcolor[RGB]{52,168,83}{+3.2})                                             & {\ul 47.1 (\textcolor[RGB]{52,168,83}{+4.0})}                                                  & 54.5 (\textcolor[RGB]{52,168,83}{+0.9})                                           \\
dis $n_{\rm{task}}$ = 5  & {\ul 46.4 (\textcolor[RGB]{52,168,83}{+2.4})}                                           & {\ul 40.7 (\textcolor[RGB]{52,168,83}{+1.2})}                                         & \textbf{51.3 (\textcolor[RGB]{52,168,83}{+4.6})}                                    & 46.8 (\textcolor[RGB]{52,168,83}{+3.7})                                                        & {\ul 54.6 (\textcolor[RGB]{52,168,83}{+1.0})}                                     \\
dis $n_{\rm{task}}$ = 1  & \textbf{47.0 (\textcolor[RGB]{52,168,83}{+3.0})}                                        & \textbf{42.1 (\textcolor[RGB]{52,168,83}{+2.6})}                                      & {\ul 50.8 (\textcolor[RGB]{52,168,83}{+4.1})}                                       & \textbf{47.4 (\textcolor[RGB]{52,168,83}{+4.3})}                                               & \textbf{55.2 (\textcolor[RGB]{52,168,83}{+1.6})}                                  \\ \hline
\end{tabular}
}
\label{table:task_number}
\end{table}

\begin{figure*}[h]
% \vspace{-2em}
\begin{center}
\includegraphics[width=0.85\textwidth]{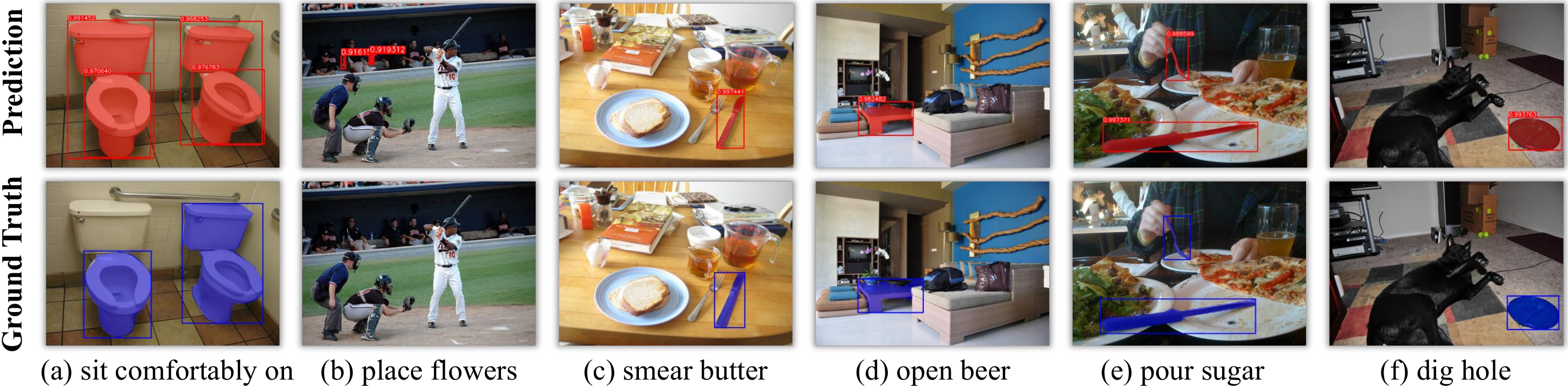}
\end{center}
% \vspace{-1em}
\caption{Qualitative results of proposed method for task oriented instance segmentation.}
\label{fig:quali}
\vspace{-1em}
\end{figure*}

\textbf{Qualitative Results.}
Fig.\ref{fig:quali} shows more qualitative results.
%In (a), two toilets are target objects. One of them is annotated partially while another totally.
In (a), two toilets are taken as target objects and annotated partially or totally.
But TOIST simultaneously predicts the two kinds of results for each toilet.
In (b), no object is annotated, while TOIST keenly detects two water bottles that afford the task.
In (c), TOIST predicts more accurate mask result than ground truth.
In (d), the table is selected and interestingly it does afford the task as the table edge can be used to open beer bottles.
\textcolor[RGB]{0,0,0}{
More qualitative results can be found in Appendix~\ref{appendix:Qualitative}.
}

% In the sample of \emph{sit comfortably on}, two toilets are selected as ground truth, but one of them are annotated partially while another one totally. However, TOIST simultaneously predicts the two kinds of results for each toilet.
% In the sample of \emph{place flowers}, no object is selected as suitable one, while TOIST is keen to detect two water bottles that afford the task.
% In the sample of \emph{smear butter}, TOIST predicts more accurate mask result than ground truth.
% In the sample of \emph{open beer}, the table is selected and interestingly it does afford the task as the table edge can be used to open beer bottles.

\section{Conclusion and Discussion}
\label{Conclusion}
We explore the problem of task oriented instance segmentation and propose a transformer-based method named as TOIST with a novel noun-pronoun distillation framework.
Experiments show our method successfully models affordance and preference, achieving SOTA results on the COCO-Tasks dataset.
\textbf{Limitations.} Due to the lack of large-scale datasets with more abundant tasks, TOIST is only evaluated on limited tasks. While this is sufficient for many robotics applications, it would be interesting to explore general verb reference understanding on more tasks.
\textbf{Potential Negative Social Impact.} Because TOIST is not perfect, when it  is used in robotics applications, robots may have difficulty in selecting the most suitable object to carry out a task or even cause damage.

% % Please add the following required packages to your document preamble:
% % \usepackage{booktabs}
% \begin{table}
% \begin{tabular}{@{}c|cccccc|cccccc@{}}
% \toprule
% model & \multicolumn{6}{c|}{TOIST}              & \multicolumn{6}{c}{TOIST w/o decoder self-attention}              \\ \midrule
% score & 1st  & 2nd  & 3rd  & 4th  & 5th  & 6th  & 1st  & 2nd  & 3rd  & 4th  & 5th  & 6th  \\ \midrule
% bbox  & 29.6 & 34.8 & 38.3 & 40.4 & 41.4 & 41.4 & 33.9 & 33.4 & 32.2 & 31.1 & 31.1 & 31.3 \\
% segm  & 25.0 & 29.6 & 32.5 & 34.4 & 35.2 & 35.2 & 28.7 & 28.4 & 27.2 & 26.3 & 26.3 & 26.5 \\ \bottomrule
% \end{tabular}
% \end{table}

%%%%%%%%%%%%%%%%%%%%%%%%%%%%%%%%%%%%%%%%%%%%%%%%%%%%%%%%%%%%

\bibliographystyle{plain}
\bibliography{neurips_2022}

\begin{thebibliography}{10}

\bibitem{al2019character}
Rami Al-Rfou, Dokook Choe, Noah Constant, Mandy Guo, and Llion Jones.
\newblock Character-level language modeling with deeper self-attention.
\newblock In {\em Proceedings of the AAAI conference on artificial
  intelligence}, volume~33, pages 3159--3166, 2019.

\bibitem{anderson2018vision}
Peter Anderson, Qi~Wu, Damien Teney, Jake Bruce, Mark Johnson, Niko
  S{\"u}nderhauf, Ian Reid, Stephen Gould, and Anton Van Den~Hengel.
\newblock Vision-and-language navigation: Interpreting visually-grounded
  navigation instructions in real environments.
\newblock In {\em Proceedings of the IEEE conference on computer vision and
  pattern recognition}, pages 3674--3683, 2018.

\bibitem{antol2015vqa}
Stanislaw Antol, Aishwarya Agrawal, Jiasen Lu, Margaret Mitchell, Dhruv Batra,
  C~Lawrence Zitnick, and Devi Parikh.
\newblock Vqa: Visual question answering.
\newblock In {\em Proceedings of the IEEE international conference on computer
  vision}, pages 2425--2433, 2015.

\bibitem{barnard2003matching}
Kobus Barnard, Pinar Duygulu, David Forsyth, Nando de~Freitas, David~M Blei,
  and Michael~I Jordan.
\newblock Matching words and pictures.
\newblock {\em Journal of Machine Learning Research}, 3:1107--1135, 2003.

\bibitem{brown2020language}
Tom Brown, Benjamin Mann, Nick Ryder, Melanie Subbiah, Jared~D Kaplan, Prafulla
  Dhariwal, Arvind Neelakantan, Pranav Shyam, Girish Sastry, Amanda Askell,
  et~al.
\newblock Language models are few-shot learners.
\newblock {\em Advances in neural information processing systems},
  33:1877--1901, 2020.

\bibitem{carion2020end}
Nicolas Carion, Francisco Massa, Gabriel Synnaeve, Nicolas Usunier, Alexander
  Kirillov, and Sergey Zagoruyko.
\newblock End-to-end object detection with transformers.
\newblock In {\em European conference on computer vision}, pages 213--229.
  Springer, 2020.

\bibitem{chen2020learning}
Dian Chen, Brady Zhou, Vladlen Koltun, and Philipp Kr{\"a}henb{\"u}hl.
\newblock Learning by cheating.
\newblock In {\em Conference on Robot Learning}, pages 66--75. PMLR, 2020.

\bibitem{chen2017learning}
Guobin Chen, Wongun Choi, Xiang Yu, Tony Han, and Manmohan Chandraker.
\newblock Learning efficient object detection models with knowledge
  distillation.
\newblock {\em Advances in neural information processing systems}, 30, 2017.

\bibitem{chen2022cerberus}
Xiaoxue Chen, Tianyu Liu, Hao Zhao, Guyue Zhou, and Ya-Qin Zhang.
\newblock Cerberus transformer: Joint semantic, affordance and attribute
  parsing.
\newblock In {\em Proceedings of the IEEE/CVF Conference on Computer Vision and
  Pattern Recognition}, pages 19649--19658, 2022.

\bibitem{chen2020uniter}
Yen-Chun Chen, Linjie Li, Licheng Yu, Ahmed El~Kholy, Faisal Ahmed, Zhe Gan,
  Yu~Cheng, and Jingjing Liu.
\newblock Uniter: Universal image-text representation learning.
\newblock In {\em European conference on computer vision}, pages 104--120.
  Springer, 2020.

\bibitem{cheung2003visual}
German~KM Cheung, Simon Baker, and Takeo Kanade.
\newblock Visual hull alignment and refinement across time: A 3d reconstruction
  algorithm combining shape-from-silhouette with stereo.
\newblock In {\em 2003 IEEE Computer Society Conference on Computer Vision and
  Pattern Recognition, 2003. Proceedings.}, volume~2, pages II--375. IEEE,
  2003.

\bibitem{chuang2018learning}
Ching-Yao Chuang, Jiaman Li, Antonio Torralba, and Sanja Fidler.
\newblock Learning to act properly: Predicting and explaining affordances from
  images.
\newblock In {\em Proceedings of the IEEE Conference on Computer Vision and
  Pattern Recognition}, pages 975--983, 2018.

\bibitem{do2018affordancenet}
Thanh-Toan Do, Anh Nguyen, and Ian Reid.
\newblock Affordancenet: An end-to-end deep learning approach for object
  affordance detection.
\newblock In {\em 2018 IEEE international conference on robotics and automation
  (ICRA)}, pages 5882--5889. IEEE, 2018.

\bibitem{dosovitskiy2020image}
Alexey Dosovitskiy, Lucas Beyer, Alexander Kolesnikov, Dirk Weissenborn,
  Xiaohua Zhai, Thomas Unterthiner, Mostafa Dehghani, Matthias Minderer, Georg
  Heigold, Sylvain Gelly, et~al.
\newblock An image is worth 16x16 words: Transformers for image recognition at
  scale.
\newblock In {\em International Conference on Learning Representations}, 2020.

\bibitem{duan2018weakly}
Xuguang Duan, Wenbing Huang, Chuang Gan, Jingdong Wang, Wenwu Zhu, and Junzhou
  Huang.
\newblock Weakly supervised dense event captioning in videos.
\newblock {\em Advances in Neural Information Processing Systems}, 31, 2018.

\bibitem{fried2018speaker}
Daniel Fried, Ronghang Hu, Volkan Cirik, Anna Rohrbach, Jacob Andreas,
  Louis-Philippe Morency, Taylor Berg-Kirkpatrick, Kate Saenko, Dan Klein, and
  Trevor Darrell.
\newblock Speaker-follower models for vision-and-language navigation.
\newblock {\em Advances in Neural Information Processing Systems}, 31, 2018.

\bibitem{girdhar2017attentional}
Rohit Girdhar and Deva Ramanan.
\newblock Attentional pooling for action recognition.
\newblock {\em Advances in Neural Information Processing Systems}, 30, 2017.

\bibitem{gou2021knowledge}
Jianping Gou, Baosheng Yu, Stephen~J Maybank, and Dacheng Tao.
\newblock Knowledge distillation: A survey.
\newblock {\em International Journal of Computer Vision}, 129(6):1789--1819,
  2021.

\bibitem{gurari2018vizwiz}
Danna Gurari, Qing Li, Abigale~J Stangl, Anhong Guo, Chi Lin, Kristen Grauman,
  Jiebo Luo, and Jeffrey~P Bigham.
\newblock Vizwiz grand challenge: Answering visual questions from blind people.
\newblock In {\em Proceedings of the IEEE Conference on Computer Vision and
  Pattern Recognition}, pages 3608--3617, 2018.

\bibitem{he2017mask}
Kaiming He, Georgia Gkioxari, Piotr Doll{\'a}r, and Ross Girshick.
\newblock Mask r-cnn.
\newblock In {\em Proceedings of the IEEE international conference on computer
  vision}, pages 2961--2969, 2017.

\bibitem{he2016deep}
Kaiming He, Xiangyu Zhang, Shaoqing Ren, and Jian Sun.
\newblock Deep residual learning for image recognition.
\newblock In {\em Proceedings of the IEEE conference on computer vision and
  pattern recognition}, pages 770--778, 2016.

\bibitem{hinton2015distilling}
Geoffrey Hinton, Oriol Vinyals, Jeff Dean, et~al.
\newblock Distilling the knowledge in a neural network.
\newblock {\em arXiv preprint arXiv:1503.02531}, 2(7), 2015.

\bibitem{hosseini2014learning}
Mohammad~Javad Hosseini, Hannaneh Hajishirzi, Oren Etzioni, and Nate Kushman.
\newblock Learning to solve arithmetic word problems with verb categorization.
\newblock In {\em EMNLP}, volume 523533. Citeseer, 2014.

\bibitem{hu2016segmentation}
Ronghang Hu, Marcus Rohrbach, and Trevor Darrell.
\newblock Segmentation from natural language expressions.
\newblock In {\em European Conference on Computer Vision}, pages 108--124.
  Springer, 2016.

\bibitem{johnson2017clevr}
Justin Johnson, Bharath Hariharan, Laurens Van Der~Maaten, Li~Fei-Fei,
  C~Lawrence~Zitnick, and Ross Girshick.
\newblock Clevr: A diagnostic dataset for compositional language and elementary
  visual reasoning.
\newblock In {\em Proceedings of the IEEE conference on computer vision and
  pattern recognition}, pages 2901--2910, 2017.

\bibitem{kamath2021mdetr}
Aishwarya Kamath, Mannat Singh, Yann LeCun, Gabriel Synnaeve, Ishan Misra, and
  Nicolas Carion.
\newblock Mdetr-modulated detection for end-to-end multi-modal understanding.
\newblock In {\em Proceedings of the IEEE/CVF International Conference on
  Computer Vision}, pages 1780--1790, 2021.

\bibitem{kazemzadeh2014referitgame}
Sahar Kazemzadeh, Vicente Ordonez, Mark Matten, and Tamara Berg.
\newblock Referitgame: Referring to objects in photographs of natural scenes.
\newblock In {\em Proceedings of the 2014 conference on empirical methods in
  natural language processing (EMNLP)}, pages 787--798, 2014.

\bibitem{krishna2017dense}
Ranjay Krishna, Kenji Hata, Frederic Ren, Li~Fei-Fei, and Juan Carlos~Niebles.
\newblock Dense-captioning events in videos.
\newblock In {\em Proceedings of the IEEE international conference on computer
  vision}, pages 706--715, 2017.

\bibitem{kuhn1955hungarian}
Harold~W Kuhn.
\newblock The hungarian method for the assignment problem.
\newblock {\em Naval research logistics quarterly}, 2(1-2):83--97, 1955.

\bibitem{kulkarni2013babytalk}
Girish Kulkarni, Visruth Premraj, Vicente Ordonez, Sagnik Dhar, Siming Li,
  Yejin Choi, Alexander~C Berg, and Tamara~L Berg.
\newblock Babytalk: Understanding and generating simple image descriptions.
\newblock {\em IEEE transactions on pattern analysis and machine intelligence},
  35(12):2891--2903, 2013.

\bibitem{lazebnik2001computing}
Svetlana Lazebnik, Edmond Boyer, and Jean Ponce.
\newblock On computing exact visual hulls of solids bounded by smooth surfaces.
\newblock In {\em Proceedings of the 2001 IEEE Computer Society Conference on
  Computer Vision and Pattern Recognition. CVPR 2001}, volume~1, pages I--I.
  IEEE, 2001.

\bibitem{lee2020learning}
Joonho Lee, Jemin Hwangbo, Lorenz Wellhausen, Vladlen Koltun, and Marco Hutter.
\newblock Learning quadrupedal locomotion over challenging terrain.
\newblock {\em Science robotics}, 5(47):eabc5986, 2020.

\bibitem{li2020oscar}
Xiujun Li, Xi~Yin, Chunyuan Li, Pengchuan Zhang, Xiaowei Hu, Lei Zhang, Lijuan
  Wang, Houdong Hu, Li~Dong, Furu Wei, et~al.
\newblock Oscar: Object-semantics aligned pre-training for vision-language
  tasks.
\newblock In {\em European Conference on Computer Vision}, pages 121--137.
  Springer, 2020.

\bibitem{li2022distance}
Yang Li, Yucheng Tu, Xiaoxue Chen, Hao Zhao, and Guyue Zhou.
\newblock Distance-aware occlusion detection with focused attention.
\newblock {\em IEEE Transactions on Image Processing}, 31:5661--5676, 2022.

\bibitem{li2020hoi}
Yong-Lu Li, Xinpeng Liu, Xiaoqian Wu, Yizhuo Li, and Cewu Lu.
\newblock Hoi analysis: Integrating and decomposing human-object interaction.
\newblock {\em Advances in Neural Information Processing Systems},
  33:5011--5022, 2020.

\bibitem{li2016gated}
Yujia Li, Richard Zemel, Marc Brockschmidt, and Daniel Tarlow.
\newblock Gated graph sequence neural networks.
\newblock In {\em Proceedings of ICLR'16}, 2016.

\bibitem{li2017learning}
Zhizhong Li and Derek Hoiem.
\newblock Learning without forgetting.
\newblock {\em IEEE transactions on pattern analysis and machine intelligence},
  40(12):2935--2947, 2017.

\bibitem{lin2017feature}
Tsung-Yi Lin, Piotr Doll{\'a}r, Ross Girshick, Kaiming He, Bharath Hariharan,
  and Serge Belongie.
\newblock Feature pyramid networks for object detection.
\newblock In {\em Proceedings of the IEEE conference on computer vision and
  pattern recognition}, pages 2117--2125, 2017.

\bibitem{lin2017focal}
Tsung-Yi Lin, Priya Goyal, Ross Girshick, Kaiming He, and Piotr Doll{\'a}r.
\newblock Focal loss for dense object detection.
\newblock In {\em Proceedings of the IEEE international conference on computer
  vision}, pages 2980--2988, 2017.

\bibitem{lin2014microsoft}
Tsung-Yi Lin, Michael Maire, Serge Belongie, James Hays, Pietro Perona, Deva
  Ramanan, Piotr Doll{\'a}r, and C~Lawrence Zitnick.
\newblock Microsoft coco: Common objects in context.
\newblock In {\em European conference on computer vision}, pages 740--755.
  Springer, 2014.

\bibitem{liu2021dab}
Shilong Liu, Feng Li, Hao Zhang, Xiao Yang, Xianbiao Qi, Hang Su, Jun Zhu, and
  Lei Zhang.
\newblock Dab-detr: Dynamic anchor boxes are better queries for detr.
\newblock In {\em International Conference on Learning Representations}, 2021.

\bibitem{liu2019roberta}
Yinhan Liu, Myle Ott, Naman Goyal, Jingfei Du, Mandar Joshi, Danqi Chen, Omer
  Levy, Mike Lewis, Luke Zettlemoyer, and Veselin Stoyanov.
\newblock Roberta: A robustly optimized bert pretraining approach.
\newblock {\em arXiv preprint arXiv:1907.11692}, 2019.

\bibitem{lu2019vilbert}
Jiasen Lu, Dhruv Batra, Devi Parikh, and Stefan Lee.
\newblock Vilbert: Pretraining task-agnostic visiolinguistic representations
  for vision-and-language tasks.
\newblock {\em Advances in neural information processing systems}, 32, 2019.

\bibitem{lu202012}
Jiasen Lu, Vedanuj Goswami, Marcus Rohrbach, Devi Parikh, and Stefan Lee.
\newblock 12-in-1: Multi-task vision and language representation learning.
\newblock In {\em Proceedings of the IEEE/CVF Conference on Computer Vision and
  Pattern Recognition}, pages 10437--10446, 2020.

\bibitem{mao2016generation}
Junhua Mao, Jonathan Huang, Alexander Toshev, Oana Camburu, Alan~L Yuille, and
  Kevin Murphy.
\newblock Generation and comprehension of unambiguous object descriptions.
\newblock In {\em Proceedings of the IEEE conference on computer vision and
  pattern recognition}, pages 11--20, 2016.

\bibitem{mccarthy2003detecting}
Diana McCarthy, Bill Keller, and John~A Carroll.
\newblock Detecting a continuum of compositionality in phrasal verbs.
\newblock In {\em Proceedings of the ACL 2003 workshop on Multiword
  expressions: analysis, acquisition and treatment}, pages 73--80, 2003.

\bibitem{milletari2016v}
Fausto Milletari, Nassir Navab, and Seyed-Ahmad Ahmadi.
\newblock V-net: Fully convolutional neural networks for volumetric medical
  image segmentation.
\newblock In {\em 2016 fourth international conference on 3D vision (3DV)},
  pages 565--571. IEEE, 2016.

\bibitem{nagarajan2020learning}
Tushar Nagarajan and Kristen Grauman.
\newblock Learning affordance landscapes for interaction exploration in 3d
  environments.
\newblock {\em Advances in Neural Information Processing Systems},
  33:2005--2015, 2020.

\bibitem{nguyen2017object}
Anh Nguyen, Dimitrios Kanoulas, Darwin~G Caldwell, and Nikos~G Tsagarakis.
\newblock Object-based affordances detection with convolutional neural networks
  and dense conditional random fields.
\newblock In {\em 2017 IEEE/RSJ International Conference on Intelligent Robots
  and Systems (IROS)}, pages 5908--5915. IEEE, 2017.

\bibitem{paszke2019pytorch}
Adam Paszke, Sam Gross, Francisco Massa, Adam Lerer, James Bradbury, Gregory
  Chanan, Trevor Killeen, Zeming Lin, Natalia Gimelshein, Luca Antiga, et~al.
\newblock Pytorch: An imperative style, high-performance deep learning library.
\newblock {\em Advances in neural information processing systems}, 32, 2019.

\bibitem{plummer2015flickr30k}
Bryan~A Plummer, Liwei Wang, Chris~M Cervantes, Juan~C Caicedo, Julia
  Hockenmaier, and Svetlana Lazebnik.
\newblock Flickr30k entities: Collecting region-to-phrase correspondences for
  richer image-to-sentence models.
\newblock In {\em Proceedings of the IEEE international conference on computer
  vision}, pages 2641--2649, 2015.

\bibitem{ramesh2021zero}
Aditya Ramesh, Mikhail Pavlov, Gabriel Goh, Scott Gray, Chelsea Voss, Alec
  Radford, Mark Chen, and Ilya Sutskever.
\newblock Zero-shot text-to-image generation.
\newblock In {\em International Conference on Machine Learning}, pages
  8821--8831. PMLR, 2021.

\bibitem{ren2015faster}
Shaoqing Ren, Kaiming He, Ross Girshick, and Jian Sun.
\newblock Faster r-cnn: Towards real-time object detection with region proposal
  networks.
\newblock {\em Advances in neural information processing systems}, 28, 2015.

\bibitem{rezatofighi2019generalized}
Hamid Rezatofighi, Nathan Tsoi, JunYoung Gwak, Amir Sadeghian, Ian Reid, and
  Silvio Savarese.
\newblock Generalized intersection over union: A metric and a loss for bounding
  box regression.
\newblock In {\em Proceedings of the IEEE/CVF conference on computer vision and
  pattern recognition}, pages 658--666, 2019.

\bibitem{sawatzky2019object}
Johann Sawatzky, Yaser Souri, Christian Grund, and Jurgen Gall.
\newblock What object should i use?-task driven object detection.
\newblock In {\em Proceedings of the IEEE/CVF Conference on Computer Vision and
  Pattern Recognition}, pages 7605--7614, 2019.

\bibitem{simonyan2014two}
Karen Simonyan and Andrew Zisserman.
\newblock Two-stream convolutional networks for action recognition in videos.
\newblock {\em Advances in neural information processing systems}, 27, 2014.

\bibitem{su2019vl}
Weijie Su, Xizhou Zhu, Yue Cao, Bin Li, Lewei Lu, Furu Wei, and Jifeng Dai.
\newblock Vl-bert: Pre-training of generic visual-linguistic representations.
\newblock In {\em International Conference on Learning Representations}, 2019.

\bibitem{wang2013dense}
Heng Wang, Alexander Kl{\"a}ser, Cordelia Schmid, and Cheng-Lin Liu.
\newblock Dense trajectories and motion boundary descriptors for action
  recognition.
\newblock {\em International journal of computer vision}, 103(1):60--79, 2013.

\bibitem{wang2017binge}
Xiaolong Wang, Rohit Girdhar, and Abhinav Gupta.
\newblock Binge watching: Scaling affordance learning from sitcoms.
\newblock In {\em Proceedings of the IEEE Conference on Computer Vision and
  Pattern Recognition}, pages 2596--2605, 2017.

\bibitem{yu2015visual}
Licheng Yu, Eunbyung Park, Alexander~C Berg, and Tamara~L Berg.
\newblock Visual madlibs: Fill in the blank description generation and question
  answering.
\newblock In {\em Proceedings of the ieee international conference on computer
  vision}, pages 2461--2469, 2015.

\bibitem{yu2016modeling}
Licheng Yu, Patrick Poirson, Shan Yang, Alexander~C Berg, and Tamara~L Berg.
\newblock Modeling context in referring expressions.
\newblock In {\em European Conference on Computer Vision}, pages 69--85.
  Springer, 2016.

\bibitem{zellers2019recognition}
Rowan Zellers, Yonatan Bisk, Ali Farhadi, and Yejin Choi.
\newblock From recognition to cognition: Visual commonsense reasoning.
\newblock In {\em Proceedings of the IEEE/CVF conference on computer vision and
  pattern recognition}, pages 6720--6731, 2019.

\bibitem{zhang2021mining}
Aixi Zhang, Yue Liao, Si~Liu, Miao Lu, Yongliang Wang, Chen Gao, and Xiaobo Li.
\newblock Mining the benefits of two-stage and one-stage hoi detection.
\newblock {\em Advances in Neural Information Processing Systems}, 34, 2021.

\bibitem{zhao2020learning}
Hao Zhao, Ming Lu, Anbang Yao, Yurong Chen, and Li~Zhang.
\newblock Learning to draw sight lines.
\newblock {\em International Journal of Computer Vision}, 128(5):1076--1100,
  2020.

\bibitem{zhu2020deformable}
Xizhou Zhu, Weijie Su, Lewei Lu, Bin Li, Xiaogang Wang, and Jifeng Dai.
\newblock Deformable detr: Deformable transformers for end-to-end object
  detection.
\newblock In {\em International Conference on Learning Representations}, 2020.

\bibitem{zhu2015understanding}
Yixin Zhu, Yibiao Zhao, and Song Chun~Zhu.
\newblock Understanding tools: Task-oriented object modeling, learning and
  recognition.
\newblock In {\em Proceedings of the IEEE Conference on Computer Vision and
  Pattern Recognition}, pages 2855--2864, 2015.

\bibitem{zuo2015interactive}
Xinxin Zuo, Chao Du, Sen Wang, Jiangbin Zheng, and Ruigang Yang.
\newblock Interactive visual hull refinement for specular and transparent
  object surface reconstruction.
\newblock In {\em Proceedings of the IEEE International Conference on Computer
  Vision}, pages 2237--2245, 2015.

\end{thebibliography}

%%%%%%%%%%%%%%%%%%%%%%%%%%%%%%%%%%%%%%%%%%%%%%%%%%%%%%%%%%%%
\section*{Checklist}

% %%% BEGIN INSTRUCTIONS %%%
% The checklist follows the references.  Please
% read the checklist guidelines carefully for information on how to answer these
% questions.  For each question, change the default \answerTODO{} to \answerYes{},
% \answerNo{}, or \answerNA{}.  You are strongly encouraged to include a {\bf
% justification to your answer}, either by referencing the appropriate section of
% your paper or providing a brief inline description.  For example:
% \begin{itemize}
%   \item Did you include the license to the code and datasets? \answerYes{See Section~\ref{gen_inst}.}
%   \item Did you include the license to the code and datasets? \answerNo{The code and the data are proprietary.}
%   \item Did you include the license to the code and datasets? \answerNA{}
% \end{itemize}
% Please do not modify the questions and only use the provided macros for your
% answers.  Note that the Checklist section does not count towards the page
% limit.  In your paper, please delete this instructions block and only keep the
% Checklist section heading above along with the questions/answers below.
% %%% END INSTRUCTIONS %%%

\begin{enumerate}

\item For all authors...
\begin{enumerate}
  \item Do the main claims made in the abstract and introduction accurately reflect the paper's contributions and scope?
    \answerYes{}
  \item Did you describe the limitations of your work?
    \answerYes{See Section~\ref{Conclusion}}
  \item Did you discuss any potential negative societal impacts of your work?
    \answerYes{See Section~\ref{Conclusion}}
  \item Have you read the ethics review guidelines and ensured that your paper conforms to them?
    \answerYes{}
\end{enumerate}

\item If you are including theoretical results...
\begin{enumerate}
  \item Did you state the full set of assumptions of all theoretical results?
    \answerNA{}
        \item Did you include complete proofs of all theoretical results?
    \answerNA{}
\end{enumerate}

\item If you ran experiments...
\begin{enumerate}
  \item Did you include the code, data, and instructions needed to reproduce the main experimental results (either in the supplemental material or as a URL)?
    \answerYes{See \url{https://github.com/AIR-DISCOVER/TOIST}.}
  \item Did you specify all the training details (e.g., data splits, hyperparameters, how they were chosen)?
    \answerYes{See Appendix~\ref{appendix:implementation_details} and ~\ref{appendix:dataset}.}
        \item Did you report error bars (e.g., with respect to the random seed after running experiments multiple times)?
    \answerYes{See Appendix~\ref{appendix:replicates}.}
        \item Did you include the total amount of compute and the type of resources used (e.g., type of GPUs, internal cluster, or cloud provider)?
    \answerYes{See Appendix~\ref{appendix:hyper}.}
\end{enumerate}

\item If you are using existing assets (e.g., code, data, models) or curating/releasing new assets...
\begin{enumerate}
  \item If your work uses existing assets, did you cite the creators?
    \answerYes{}
  \item Did you mention the license of the assets?
    \answerYes{(a) MIT License for the COCO-Tasks dataset. (b) Creative Commons Attribution 4.0 License for the Microsoft COCO dataset. (c) Apache License 2.0 for MDETR, DETR and Mask-RCNN implemented by Detectron2.}
  \item Did you include any new assets either in the supplemental material or as a URL?
    \answerYes{See Appendix~\ref{appendix:asserts}.}
  \item Did you discuss whether and how consent was obtained from people whose data you're using/curating?
    \answerYes{See Appendix~\ref{appendix:asserts}.}
  \item Did you discuss whether the data you are using/curating contains personally identifiable information or offensive content?
    \answerYes{See Appendix~\ref{appendix:asserts}.}
\end{enumerate}

\item If you used crowdsourcing or conducted research with human subjects...
\begin{enumerate}
  \item Did you include the full text of instructions given to participants and screenshots, if applicable?
    \answerNA{}
  \item Did you describe any potential participant risks, with links to Institutional Review Board (IRB) approvals, if applicable?
    \answerNA{}
  \item Did you include the estimated hourly wage paid to participants and the total amount spent on participant compensation?
    \answerNA{}
\end{enumerate}

\end{enumerate}

%%%%%%%%%%%%%%%%%%%%%%%%%%%%%%%%%%%%%%%%%%%%%%%%%%%%%%%%%%%%
%%%%%%%%%%%%%%%%%%%%%%%%%%%%%%%%%%%%%%%%%%%%%%%%%%%%%%%%%%%%
%%%%%%%%%%%%%%%%%%%%%%%%%%%%%%%%%%%%%%%%%%%%%%%%%%%%%%%%%%%%
%%%%%%%%%%%%%%%%%%%%%%%%%%%%%%%%%%%%%%%%%%%%%%%%%%%%%%%%%%%%
%%%%%%%%%%%%%%%%%%%%%%%%%%%%%%%%%%%%%%%%%%%%%%%%%%%%%%%%%%%%
%%%%%%%%%%%%%%%%%%%%%%%%%%%%%%%%%%%%%%%%%%%%%%%%%%%%%%%%%%%%
%%%%%%%%%%%%%%%%%%%%%%%%%%%%%%%%%%%%%%%%%%%%%%%%%%%%%%%%%%%%
%%%%%%%%%%%%%%%%%%%%%%%%%%%%%%%%%%%%%%%%%%%%%%%%%%%%%%%%%%%%
%%%%%%%%%%%%%%%%%%%%%%%%%%%%%%%%%%%%%%%%%%%%%%%%%%%%%%%%%%%%
%%%%%%%%%%%%%%%%%%%%%%%%%%%%%%%%%%%%%%%%%%%%%%%%%%%%%%%%%%%%
%%%%%%%%%%%%%%%%%%%%%%%%%%%%%%%%%%%%%%%%%%%%%%%%%%%%%%%%%%%%
%%%%%%%%%%%%%%%%%%%%%%%%%%%%%%%%%%%%%%%%%%%%%%%%%%%%%%%%%%%%
%%%%%%%%%%%%%%%%%%%%%%%%%%%%%%%%%%%%%%%%%%%%%%%%%%%%%%%%%%%%

\clearpage
\appendix

{\Large \textbf{Appendix}}

\section{Implementation Details}\label{appendix:implementation_details}
\subsection{Method}\label{appendix:im_method}
\paragraph{Noun Features.}
As mentioned in Section~\ref{section:formulation}, in an input image, it is possible that no objects or multiple objects afford a specific task. And in the latter case, the objects may belong to multiple classes. But for the language input $\mathbf{X}_l$ of verb-noun form, the noun corresponds to the ground truth object categories. Therefore, when the count of targets $n_{\rm{gt}}=0$, we use an empty string to construct $\mathbf{X}_l$. When $n_{\rm{gt}}>0$ and all target objects belong to the same category, we take a phrase like \emph{sit comfortably on sofa} as $\mathbf{X}_l$. When $n_{\rm{gt}}>0$ and the target objects belong to multiple classes, the language input $\mathbf{X}_l$ is set to the concatenation of multiple phrasal verbs, such as \emph{sit comfortably on chair sit comfortably on bed}.

We only update the proposed memory bank in the latter two cases. In these cases, if a noun is encoded into multiple tokenized features by the text encoder, we use the mean value of the features processed by the transformer encoder as ${l_{\rm{noun}}^{\rm{tr}}}$ for updating. In the last case, we take the average of multiple noun features ${l_{\rm{noun-1}}^{\rm{tr}}}, \ldots, {l_{\rm{noun-n_c}}^{\rm{tr}}}$ as the final  ${l_{\rm{noun}}^{\rm{tr}}}$, where $n_c$ is the count of classes. In this way, the privileged knowledge of multiple nouns is more easily distilled into the single pronoun feature of the student model when an image contains multiple classes of objects equally suitable for a task.

\paragraph{Components of TOIST.} 
RoBERTa-base \cite{liu2019roberta} and ResNet-101 \cite{he2016deep} are used as the text encoder and the CNN-based backbone (the image encoder).
For the logit head and the detect head, two feed forward networks of depth one and three are leveraged, respectively. For the segment head, following the network design of \cite{carion2020end}, a multi-head attention operator and a FPN \cite{lin2017feature}-like convolutional neural network are used.
After each block in the transformer decoder, TOIST generates auxiliary outputs \cite{al2019character} with the prediction heads.

\subsection{Loss Functions}\label{appendix:loss}
We present the loss terms used for the plain TOIST training in details.

For the ground truth objects $\mathbf{O_{\rm{gt}}}$ of each training sample, we define ${{\mathbf{p}}^{\rm{span}}_{i}}=[p^{\rm{span}}_{i,1}, \ldots, p^{\rm{span}}_{i,n_{\rm{max}}}] \in [0,1]^{n_{\rm{max}}}$ as a uniform distribution over the span positions of the text tokens corresponding to the $i$-th ground truth object. $p^{\rm{span}}_{i,n_{\rm{max}}}$ stands for the probability of "no-object", which is 0 for the ground truth objects.
As a reminder, we use the whole verb-pronoun (or verb-noun) description as token span. Assuming that $n_{\rm{gt}}$ (the count of $\mathbf{O_{\rm{gt}}}$) is smaller that $n_{\rm{pred}}$ (the count of the predicted objects $\mathbf{O_{\rm{pred}}}$), we pad $\mathbf{O_{\rm{gt}}}$ with $\varnothing$ ("no-object") to be of size $n_{\rm{pred}}$, denoted as $\mathbf{O_{\rm{gt}}^{\prime}}$. For $\varnothing$, ${{\mathbf{p}}^{\rm{span}}_{i}}$ is assigned to be $p^{\rm{span}}_{i,j} = \mathds{1}_{\{j = n_{\rm{max}}\}}$: if $j = n_{\rm{max}}$, $p^{\rm{span}}_{i,j}=1$; otherwise, $p^{\rm{span}}_{i,j}=0$.

We denote the bipartite matching between $\mathbf{O_{\rm{gt}}^{\prime}}$ and $\mathbf{O_{\rm{pred}}}$ as $\hat \sigma_0$, which is calculated by  minimizing the matching loss with the Hungarian algorithm \cite{carion2020end}:
\begin{equation}
\hat{\sigma_0}=\mathop{\arg \min }\limits_{\sigma_0 \in \mathfrak{S}_{{n_{\rm{pred}}}}} \sum_{i}^{{n_{\rm{pred}}}}
\mathds{1}_{\{p^{\rm{span}}_{i,n_{\rm{max}}} = 0\}}
\left[\mathcal{L}_{\rm{l1}}(b_{i}, \hat{b}_{\sigma_0(i)})+
\mathcal{L}_{\rm{giou}}(b_{i}, \hat{b}_{\sigma_0(i)})+
\mathcal{L}_{\rm{token-m}}({{\mathbf{p}}^{\rm{span}}_{i}}, {\hat{\mathbf{g}}_{\sigma_0(i)}})\right].
\end{equation}
\textcolor[RGB]{0,0,0} {
Here, $\mathfrak{S}_{{n_{\rm{pred}}}}$ is the set of all permutations of $n_{\rm{pred}}$ elements.
}
$b_{i}$ and $\hat{b}_{\sigma_0(i)}$ are the ground truth box and the predicted box, respectively. ${\hat{\mathbf{g}}_{\sigma_0(i)}}$ is the predicted logit, as detailed in the main paper. The loss terms are defined as follows:
\begin{equation}
    \mathcal{L}_{\rm{l1}}(b_{i}, \hat{b}_{\sigma_0(i)}) = \left\|b_{i}-\hat{b}_{\sigma_0(i)}\right\|_{1}.
\end{equation}
\begin{equation}
    \mathcal{L}_{\text {giou}}(b_{i}, \hat{b}_{\sigma_0(i)})=1-\left(\frac{|b_{i} \cap \hat{b}_{\sigma_0(i)}|}{|b_{i} \cup \hat{b}_{\sigma_0(i)}|}-\frac{|B(b_{i}, \hat{b}_{\sigma_0(i)}) \backslash b_{i} \cup \hat{b}_{\sigma_0(i)}|}{|B(b_{i}, \hat{b}_{\sigma_0(i)})|}\right).
\end{equation}
\begin{equation}
    \mathcal{L}_{\rm{token-m}}({{\mathbf{p}}^{\rm{span}}_{i}}, {\hat{\mathbf{g}}_{\sigma_0(i)}}) = -\sum_{j}^{n_{\rm{max}}} p^{\rm{span}}_{i,j} \frac{\exp \left(\hat g_{j}^{\sigma_0(i)}\right)}{\sum_{l=1}^{n_{\rm{max}}} \exp \left(\hat g_{l}^{\sigma_0(i)}\right)}.
\end{equation}
$\mathcal{L}_{\text {giou}}$ is the Generalized Intersection over Union (GIoU) loss \cite{rezatofighi2019generalized}. $|\cdot|$ calculates the size of an area. $B(b_{i}, \hat{b}_{\sigma_0(i)})$ stands for the smallest box containing $b_{i}$ and $\hat{b}_{\sigma_0(i)}$. The calculation of $\mathcal{L}_{\text {giou}}$ is implemented by linear functions, so it is differentiable and can be used for back propagation.

For segmentation, Dice/F-1 loss \cite{milletari2016v} $\mathcal{L}_{\rm{dice}}$ and Focal cross-entropy loss \cite{lin2017focal} $\mathcal{L}_{\rm{cross}}$ are leveraged:
\begin{equation}
\mathcal{L}_{\mathrm{dice}}(m_i, {\hat m_{\sigma_0(i)}})=1-\frac{2 m_i \delta({\hat m_{\sigma_0(i)}})+1}{\delta({\hat m_{\sigma_0(i)}})+m_i+1}.
\end{equation}
Here, $m_i$ is the ground truth instance mask of the $i$-th object. $\hat m_{\sigma_0(i)}$ is the corresponding predicted mask logits.  $\delta$ is the sigmoid function.
\begin{equation}
\mathcal{L}_{\mathrm{cross}}(m_i, {\hat m_{\sigma_0(i)}})=-\alpha_t (1 - p_t)^\gamma \left[m_i\log\delta({\hat m_{\sigma_0(i)}}) + (1-m_i)\log(1-\delta({\hat m_{\sigma_0(i)}}))\right],
\end{equation}
where
\begin{equation}
    \alpha_t = \alpha m_i + (1-\alpha) (1-m_i),
\end{equation}
\begin{equation}
    p_t = m_i \delta({\hat m_{\sigma_0(i)}}) + (1-m_i) (1-\delta({\hat m_{\sigma_0(i)}})),
\end{equation}
$\alpha$ and $\gamma$ are hyper-parameters.

The soft-token prediction loss $\mathcal{L}_{\rm{token}}$ is defined as:
\begin{equation}
    \mathcal{L}_{\rm{token}}({{\mathbf{p}}^{\rm{span}}_{i}}, {\hat{\mathbf{g}}_{\sigma_0(i)}}) = -\sum_{j}^{n_{\rm{max}}} p^{\rm{span}}_{i,j} \log\frac{\exp \left(\hat g_{j}^{\sigma_0(i)}\right)}{\sum_{l=1}^{n_{\rm{max}}} \exp \left(\hat g_{l}^{\sigma_0(i)}\right)}.
\end{equation}

The contrastive alignment loss is used to align the embedded features of the predicted objects and the corresponding text tokens. The embedded features are obtained by projecting the processed text features of the transformer encoder and the output features of transformer decoder to the same smaller dimension. We follow the definition of \cite{kamath2021mdetr}:
\begin{equation}
\begin{split}
    \mathcal{L}_{\rm{align}} &= \frac{1}{2} \sum_{i}^{n_{\rm{pred}}} \frac{1}{\left|T_{i}^{+}\right|} \sum_{j \in T_{i}^{+}}-\log \frac{\exp \left(o_{i}^{\top} t_{j} / \tau\right)}{\sum_{k=1}^{n_{\rm{max}}} \exp \left(o_{i}^{\top} t_{k} / \tau\right)} \\
    &+\frac{1}{2} \sum_{i}^{n_{\rm{max}}} \frac{1}{\left|O_{i}^{+}\right|} \sum_{j \in O_{i}^{+}}-\log \frac{\exp \left(t_{i}^{\top} o_{j} / \tau\right)}{\sum_{k=1}^{n_{\rm{pred}}} \exp \left(t_{i}^{\top} o_{k} / \tau\right)}.
\end{split}
\end{equation}
$T_{i}^{+}$ is the set of token features that a predicted object feature $o_i$ should be aligned to. $O_{i}^{+}$ is the set of object features to be aligned with a token feature $t_i$. Here the predicted objects matched to $\varnothing$ are not included. $\tau$ is a hyper-parameter.

Finally, the total loss for the plain TOIST can be written as:
\begin{equation}
\begin{split}
\mathcal{L}_{\rm{TOIST}} &= \mathds{1}_{\{p^{\rm{span}}_{i,n_{\rm{max}}} = 0\}}[ \lambda_1 \mathcal{L}_{\rm{l1}}(b_{i}, \hat{b}_{\hat{\sigma_0}(i)}) + \lambda_2 \mathcal{L}_{\rm{giou}}(b_{i}, \hat{b}_{\hat{\sigma_0}(i)}) ]\\
&+ \mathds{1}_{\{p^{\rm{span}}_{i,n_{\rm{max}}} = 0\}} [\lambda_3 \mathcal{L}_{\rm{dice}}(m_i, {\hat m_{\hat{\sigma_0}(i)}}) + \lambda_4 \mathcal{L}_{\rm{cross}}(m_i, {\hat m_{\hat{\sigma_0}(i)}})] \\
&+\lambda_5 \mathcal{L}_{\rm{token}}({{\mathbf{p}}^{\rm{span}}_{i}}, {\hat{\mathbf{g}}_{\hat{\sigma_0}(i)}}) \\
&+ \lambda_6 \mathcal{L}_{\rm{align}},
\end{split}
\end{equation}

\subsection{Hyper-parameters and Training Details}\label{appendix:hyper}

For the proposed architecture, we set $d=256$, $n_{\rm{max}}=256$, $n_{tr}=6$, ${n_{\rm{pred}}}=100$, $n_{\rm{mem}}=1024$ and $\rm{K}=3$. For the loss functions, we set $\lambda_1=5$, $\lambda_2=2$, $\lambda_3=1$, $\lambda_4=1$, $\lambda_5=1$, $\lambda_6=1$, $\lambda_7=10^4$, $\lambda_8=50$, $\alpha=0.25$, $\gamma=2$ and $\tau=0.07$.

During training, we augment input images with random resize and random crop. Specifically, each image is resized such that the shortest side is between 480 and 800 pixels and the longest side is less than 1333 pixels.
With probability 0.5, an image is cropped to a random size, where each side is between 384 and 1333 pixels.

We implement TOIST with PyTorch \cite{paszke2019pytorch}. Both of the student and teacher TOIST models are initialized with the model pre-trained by \cite{kamath2021mdetr}.
We fine-tune the two models on the COCO-Tasks dataset separately for 30 epochs. 
Then we use the fine-tuned teacher model to distill knowledge to the student model for 15 epochs.
We train the models with an AdamW optimizer. The initial learning rates are set to $10^{-5}$, $10^{-5}$, $5\times10^{-5}$ for text encoder, backbone and transformer, respectively.
The weight decay is $10^{-4}$.
Our experiments were preformed on 8 NVIDIA A100 GPUs.

\section{Dataset Details}\label{appendix:dataset}
We perform experiments on the COCO-Tasks dataset \cite{sawatzky2019object} which re-annotates the COCO dataset \cite{lin2014microsoft} with preference-aware affordance labels.
The COCO-Tasks dataset contains 14 tasks.
For each task, there are 3600 train images and 900 test images.
We train the proposed architecture on all the train images and evaluate it on all the test images. 

In an image, the most suitable objects (one or more) for solving the task are selected and their bounding boxes are taken as ground truth labels for detection. 
The number of selected objects in an image varies from zero to a dozen. 
For each task, the total number of selected objects varies between 1,105 and 9,870 and the number of different object categories varies between 6 and 30.
\textcolor[RGB]{0,0,0} {
Totally, the COCO-Tasks dataset contains 65797 selected objects spanning 49 of the 80 COCO object class categories.
}
This shows the diversity of the dataset. For each task, the total count of all instances belonging to the selected categories largely varies between 7,172 and 34,160. This shows the task oriented object detection problem on this dataset is a non-trivial problem and solving it with traditional methods is very challenging. The preference between multiple classes and multiple instances of the same class must be taken into account.

The COCO-Tasks dataset is annotated with the available COCO detection boxes. Leveraging the corresponding COCO segmentation masks directly gives the upgraded instance segmentation version. We evaluate our proposed method on the upgraded dataset.

\section{Quantitative results}\label{appendix:quantitative}

% % Please add the following required packages to your document preamble:
% % \usepackage[normalem]{ulem}
% % \useunder{\uline}{\ul}{}
% \begin{table}
% \centering
% \caption{Comparison of different distillation methods.}
% \begin{tabular}{l|ll}
% \toprule
% Method                                                              & \multicolumn{1}{c}{$\rm{mAP}^{box}$} & \multicolumn{1}{c}{$\rm{mAP}^{mask}$} \\ \midrule
% TOIST                                                               & 41.3                                 & 35.2                                  \\
% distill from $l_{c_s}^j$ to $l_{\rm{pron}}^{\rm{tr}}$              & \textbf{44.1 (+2.8)}                  & \textbf{39.0 (+3.8)}                   \\
% distill from $l_{\rm{noun}}^{\rm{tr}}$ to $l_{\rm{pron}}^{\rm{tr}}$ & {\ul 41.9 (+0.6)}                     & {\ul 36.0 (+0.8)}                      \\ \bottomrule
% \end{tabular}
% \label{table:direct_distill}
% \end{table}

\textcolor[RGB]{0,0,0} {
\textbf{Strategies for Updating Memory Bank.}
To update the text feature memory bank in noun-pronoun distillation, two strategies are compared: (1) First-in-first-out. (2) Replacing the feature closest to the new-coming ${l_{\rm{noun}}^{\rm{tr}}}$. As shown in Table \ref{table:update_method}, the second strategy leads to better performance.
}
\begin{table}
\centering
\caption{Comparison of different updating strategies for the memory bank in the distillation.}
\begin{tabular}{l|ll}
\toprule
Method                    & \multicolumn{1}{c}{$\rm{mAP}^{box}$} & \multicolumn{1}{c}{$\rm{mAP}^{mask}$} \\ \midrule
TOIST w/o distillation    & 41.3                                 & 35.2                                  \\
first-in-first-out        & 43.9 (+2.6)                           & 38.8 (+3.6)                            \\
replacing the closest one & \textbf{44.1 (+2.8)}                  & \textbf{39.0 (+3.8)}                   \\ \bottomrule
\end{tabular}
\label{table:update_method}
\end{table}

% % Please add the following required packages to your document preamble:
% % \usepackage[table,xcdraw]{xcolor}
% % If you use beamer only pass "xcolor=table" option, i.e. \documentclass[xcolor=table]{beamer}
% \begin{table}
% \centering
% \caption{The results of TOIST without pre-training.}
% \begin{tabular}{l|ll}
% \toprule
% Method                    & \multicolumn{1}{c}{$\rm{mAP}^{box}$}                 & \multicolumn{1}{c}{$\rm{mAP}^{mask}$}                       \\ \midrule
% pronoun input             & 3.65                                                 & 5.74                                                        \\
% noun input                & 11.19 & 12.67                                                       \\
% noun-pronoun distillation & 7.43 (+3.78)                                          & 11.28 (+5.54) \\ \bottomrule
% \end{tabular}
% \label{table:without_pretraining}
% \end{table}

\textcolor[RGB]{0,0,0} {
\textbf{Comparison to the Baseline with the Same Backbone.} \label{appendix:stronger_baseline}
To investigate whether our TOIST architecture is a standalone technical contribution by marginalizing the benefits brought by pre-trained models, we present another baseline 'MDETR+GGNN'.
Specially, to leverage the knowledge in noun referring expression comprehension, we use the official pre-trained model of MDETR \cite{kamath2021mdetr} and then fine-tune it on the COCO-Task dataset.
We use the class names of the ground truth objects in each image as the text input to detect these objects.
Then we use the GGNN model \cite{sawatzky2019object} to infer which objects are preferred for a task. 
The results are shown in Table \ref{table:mdetr_ggnn}.
Note that this baseline is also tested with privileged noun ground truth, but our distillation method only use the privileged knowledge during training.
Nevertheless, our proposed method still has a significant performance improvement over this strong baseline (+7.3\% $\rm{mAP^{box}}$ and +8.7\% $\rm{mAP^{mask}}$).
This demonstrates that our TOIST architecture is a standalone technical contribution towards task oriented instance segmentation and pretraining is necessary but insufficient to get the performance level of TOIST.
}
% \usepackage[normalem]{ulem}
\begin{table}
\centering
\caption{Comparison of the proposed method to 'MDETR+GGNN' baseline on the COCO-Tasks dataset.}
\begin{tabular}{l|ll}
\toprule
Method                & \multicolumn{1}{c}{$\rm{mAP}^{box}$} & \multicolumn{1}{c}{$\rm{mAP}^{mask}$} \\ \midrule
MDETR + GGNN w/o pretraining            & 9.6                                 & 8.6                                  \\
MDETR + GGNN            & 36.8                                 & 30.3                                  \\
TOIST                 & {\ul 41.3 (+4.5)}                    & {\ul 35.2 (+4.9)}                     \\
TOIST w/ distillation & \textbf{44.1 (+7.3)}                 & \textbf{39.0 (+8.7)}                  \\ \bottomrule
\end{tabular}
\label{table:mdetr_ggnn}
\end{table}

\textcolor[RGB]{0,0,0} {
\textbf{Ablations for Loss Terms.}
To demonstrate the effectiveness of the used loss terms, we provide ablation studies in which we remove $\mathcal{L}_{\rm{align}}$ or $\mathcal{L}_{\rm{token}}$ or both. The quantitative results are demonstrated in Table \ref{table:loss_term}. 
It shows that removing $\mathcal{L}_{\rm{token}}$ brings a performance drop of -1.2\% $\rm{mAP}^{box}$ and -0.4\% $\rm{mAP}^{mask}$, because the association between the matched object predictions and the task descriptions is weakened.
Removing $\mathcal{L}_{\rm{align}}$ brings a performance drop of -0.2\% $\rm{mAP}^{box}$ and -0.1\% $\rm{mAP}^{mask}$, because the features of an object and its corresponding text features cannot be explicitly constrained to be closer.
Interestingly, removing both of them brings a significant performance drop of -17.9\% $\rm{mAP}^{box}$ and -14.5\% $\rm{mAP}^{mask}$, implying the two loss terms enhance the effect of each other to make TOIST understand verb reference better.
}
\begin{table}
\centering
\caption{Ablations for the soft-token prediction loss and the contrastive alignment loss.}
\begin{tabular}{l|ll}
\toprule
Method                                                                & \multicolumn{1}{c}{$\rm{mAP}^{box}$} & \multicolumn{1}{c}{$\rm{mAP}^{mask}$} \\ \midrule
TOIST                                                                 & \textbf{41.3}                         & \textbf{35.2}                          \\
TOIST w/o $\mathcal{L}_{\rm{token}}$                                 & 40.1 (-1.2)                            & 34.8 (-0.4)                             \\
TOIST w/o $\mathcal{L}_{\rm{align}}$                                 & {\ul 41.1 (-0.2)}                   & {\ul 35.1 (-0.1)}                    \\
TOIST w/o $\mathcal{L}_{\rm{token}}$ and $\mathcal{L}_{\rm{align}}$ & 23.4 (-17.9)                           & 20.7 (-14.5)                            \\ \bottomrule
\end{tabular}
\label{table:loss_term}
\end{table}

\begin{table}[htbp]
	\small\centering
	\begin{minipage}[t]{0.45\linewidth}
		\centering
		\setlength{\tabcolsep}{4pt}
		\caption{Ablations for distillation settings. CCR, CL and SBTL are short for cluster center replacement, cluster loss and soft binary target loss, respectively.}
		\resizebox{1\textwidth}{!}{
            \begin{tabular}{@{}c|ccc|ll@{}}
            \toprule
            Index & CCR & CL & SBTL & \multicolumn{1}{c}{$\rm{mAP}^{\rm{box}}$} & \multicolumn{1}{c}{$\rm{mAP}^{\rm{mask}}$}  \\ \midrule
            (a) & \texttimes     & \texttimes & \texttimes                                                & 41.3$\pm$0.44                                                                    & 35.0$\pm$0.28                                                                       \\
            (b) & \texttimes     & \texttimes & \checkmark                                                  & 43.2$\pm$0.20                                                                    & 38.0$\pm$0.02                                                                \\
            (c) & \texttimes     & \checkmark   & \texttimes                                                & 42.0$\pm$0.03                                     & 37.1$\pm$0.04                                                              \\
            (d) & \texttimes     & \checkmark   & \checkmark                                                  & {43.5$\pm$0.30}                                                                   & {38.6$\pm$0.05}                                                                      \\
            (e) & \checkmark       & \texttimes & \texttimes                                                & 42.0$\pm$0.03                                                              & 36.9$\pm$0.06                                       \\
            (f) & \checkmark       & \texttimes & \checkmark                                                  & 42.3$\pm$0.02                                                                   & 37.3$\pm$0.02                                                                  \\
            (g) & \checkmark       & \checkmark   & \texttimes                                                & 42.3$\pm$0.02                                                        & 37.5$\pm$0.03                                                           \\
            (h) & \checkmark       & \checkmark   & \checkmark                                                  & \textbf{44.1$\pm$0.12}                                                                    & \textbf{39.0$\pm$0.07}                                                                       \\ \bottomrule
            \end{tabular}
        }%
		\label{table:distillation}
	\end{minipage}\hspace{1em}
	\begin{minipage}[t]{0.45\linewidth}
	\vspace{1em}
		\centering
		\setlength{\tabcolsep}{4pt}
		\caption{Ablations for pronoun input.}
		\resizebox{1\textwidth}{!}{
        \begin{tabular}{@{}cc|cc@{}}
        \toprule
        {Method}                                                                              & {Pronoun} & {$\rm{mAP}^{\rm{box}}$} & {$\rm{mAP}^{\rm{mask}}$} \\ \midrule
        \multirow{4}{*}{TOIST}            & something        & 41.3$\pm$0.44                           & 35.0$\pm$0.28                              \\
                                                                                                     & it               & 41.4$\pm$0.26                           & 35.1$\pm$0.20                              \\
                                                                                                     & them             & 41.5$\pm$0.28                           & 34.7$\pm$0.28                              \\
                                                                                                     & abcd             & 38.9$\pm$0.10                           & 33.2$\pm$0.15                              \\ \midrule
        \multirow{4}{*}{\begin{tabular}[c]{@{}c@{}}TOIST\\ w/ distillation\end{tabular}} & something        & 44.1$\pm$0.12                          & 39.0$\pm$0.07                              \\
                                                                                                     & it               & 43.8$\pm$0.12                          & 38.4$\pm$0.02                              \\
                                                                                                     & them             & 43.7$\pm$0.13                           & 38.1$\pm$0.03                              \\
                                                                                                     & abcd             & 42.8$\pm$0.06                           & 37.4$\pm$0.02                              \\ \bottomrule
        \end{tabular}
        }%
		\label{table:pronoun}
 	\end{minipage}
\end{table}

\textbf{Replicates.}\label{appendix:replicates}
In Table \ref{table:distillation} and \ref{table:pronoun}, we show the mean and standard deviation of the results obtained by running experiments three times under different random seeds.

\textbf{Per-task Results.} \label{appendix:per_task_results}
In Table \ref{table:dete} and \ref{table:seg}, we provide per-task results of the proposed method and existing state-of-the-art baselines on the COCO-Tasks dataset. The results show that our TOIST model with noun-pronoun distillation achieves the best performance in most tasks. %except for the 5th task \emph{water plant} and the 10th task \emph{serve wine}.

% \usepackage{booktabs}
% \usepackage[normalem]{ulem}
\begin{table}
\caption{Per-task object detection results on COCO-Tasks.}
\resizebox{1\textwidth}{!}{
\begin{tabular}{@{}lccccccccccccccc@{}}
\toprule
\multicolumn{16}{c}{Object Detection, AP@0.5}                                                                                                                                                                                                                                                                                       \\ \midrule
\multicolumn{1}{c|}{Method}                   & 1                        & 2             & 3             & 4             & 5             & 6             & 7             & 8             & 9             & 10            & 11            & 12                  & 13            & \multicolumn{1}{c|}{14}            & mean          \\ \midrule
\multicolumn{1}{l|}{Faster-RCNN}              & 28.1                     & 25.8          & 30.1          & 22.0          & 30.5          & 11.7          & 30.8          & 0.0           & 5.1           & 33.4          & 9.7           & 6.1                 & 24.6          & \multicolumn{1}{c|}{30.9}          & 20.6          \\
\multicolumn{1}{l|}{Faster-RCNN + pick best}  & 22.9                     & 18.1          & 19.8          & 15.0          & 21.3          & 5.8           & 20.4          & 3.9           & 3.3           & 22.0          & 11.1          & 5.0                 & 12.5          & \multicolumn{1}{c|}{15.6}          & 14.1          \\
\multicolumn{1}{l|}{Faster-RCNN + ranker}     & 10.7                     & 10.4          & 11.5          & 11.6          & 11.8          & 3.3           & 15.0          & 2.4           & 4.6           & 10.5          & 5.2           & 5.0                 & 8.3           & \multicolumn{1}{c|}{17.2}          & 9.1           \\
\multicolumn{1}{l|}{Faster-RCNN + classifier} & 33.1                     & 26.7          & 36.8          & 32.9          & 35.4          & 14.6          & 40.3          & 14.4          & 17.6          & 38.4          & 17.1          & 24.5                & 33.2          & \multicolumn{1}{c|}{38.1}          & 28.8          \\
\multicolumn{1}{l|}{Faster-RCNN + GGNN}       & 36.6                     & 29.8          & 40.5          & 37.6          & 41.0          & 17.2          & 43.6          & 17.9          & 21.0          & 40.6          & 22.3          & 28.4                & 39.1          & \multicolumn{1}{c|}{40.7}          & 32.6          \\
\multicolumn{1}{l|}{Yolo + GGNN}              & 36.8                     & 31.9          & 39.1          & 38.0          & 41.6          & 16.5          & 44.4          & 18.7          & 23.0          & 39.0          & 22.3          & 26.9                & 44.0          & \multicolumn{1}{c|}{42.0}          & 33.2          \\
\multicolumn{1}{l|}{Mask-RCNN}                & \multicolumn{1}{l}{30.8} & 25.0          & 32.7          & 20.4          & 33.1          & 8.0           & 28.2          & 7.9           & 11.2          & 43.1          & 8.1           & 14.7                & 32.2          & \multicolumn{1}{c|}{32.0}          & 23.4          \\
\multicolumn{1}{l|}{Mask-RCNN + pick best}    & 21.9                     & 19.6          & 22.1          & 21.6          & 28.3          & 12.6          & 26.2          & 3.4           & 3.6           & 29.1          & 20.4          & 3.7                 & 22.4          & \multicolumn{1}{c|}{28.6}          & 18.8          \\
\multicolumn{1}{l|}{Mask-RCNN + ranker}       & 11.6                     & 11.6          & 12.2          & 14.0          & 15.0          & 5.0           & 18.9          & 2.6           & 3.3           & 12.2          & 7.1           & 4.7                 & 9.2           & \multicolumn{1}{c|}{21.3}          & 10.6          \\
\multicolumn{1}{l|}{Mask-RCNN + classifier}   & 36.8                     & 31.1          & 42.4          & 39.5          & 40.8          & 18.6          & 48.3          & 13.9          & 17.2          & 48.4          & 21.4          & 23.1                & 43.5          & \multicolumn{1}{c|}{46.4}          & 33.7          \\
\multicolumn{1}{l|}{Mask-RCNN + GGNN}         & 39.0                     & 33.2          & 46.4          & {\ul 43.8}    & 47.7          & 21.4          & 51.2          & 16.7          & 20.3          & \textbf{51.3} & 27.3          & 26.8                & {\ul 50.2}    & \multicolumn{1}{c|}{48.9}          & 37.4          \\
\multicolumn{1}{l|}{MDETR + GGNN}         & {\ul 44.3}                     & 36.5          & 45.2          & 28.6    & 44.0          & \textbf{27.6}          & 35.9          & 20.7          & \textbf{34.7}          & 46.3 & 27.8          & 41.5                & 46.5    & \multicolumn{1}{c|}{36.2}          & 36.8          \\ \midrule
\multicolumn{1}{l|}{TOIST}                    & 44.0               & {\ul 39.5}    & {\ul 46.7}    & 43.1          & \textbf{53.6}    & 23.5    & {\ul 52.8}    & {\ul 21.3}    & 32.0    & 46.3          & {\ul 33.1}    & {\ul {41.7}} & 48.1          & \multicolumn{1}{c|}{{\ul 52.9}}    & {\ul 41.3}    \\
\multicolumn{1}{l|}{TOIST w/ distillation}    & \textbf{45.8}            & \textbf{40.0} & \textbf{49.4} & \textbf{49.6} & {\ul 53.4} & {\ul 26.9} & \textbf{58.3} & \textbf{22.6} & {\ul 32.5} & {\ul 50.0}    & \textbf{35.5} & \textbf{43.7}       & \textbf{52.8} & \multicolumn{1}{c|}{\textbf{56.2}} & \textbf{44.1} \\ \bottomrule
\end{tabular}}
\label{table:dete}
\end{table}

% \usepackage[normalem]{ulem}
\begin{table}
\caption{Per-task instance segmentation results on COCO-Tasks.}
\resizebox{1\textwidth}{!}{
\begin{tabular}{lccccccccccccccc}
\toprule
\multicolumn{16}{c}{Instance Segmentation, AP@0.5}                                                                                                                                                                                                                                                                          \\ \midrule
\multicolumn{1}{c|}{Method}                 & 1                        & 2             & 3             & 4             & 5             & 6             & 7             & 8             & 9             & 10            & 11            & 12            & 13            & \multicolumn{1}{c|}{14}            & mean          \\ \midrule
\multicolumn{1}{l|}{Mask-RCNN}              & \multicolumn{1}{l}{23.8} & 21.0          & 33.0          & 12.1          & 33.0          & 6.8           & 19.3          & 6.4           & 7.7           & 43.0          & 6.2           & 10.9          & 31.8          & \multicolumn{1}{c|}{24.8}          & 20.0          \\
\multicolumn{1}{l|}{Mask-RCNN + pick best}  & 17.9                     & 16.4          & 21.7          & 16.8          & 27.9          & 10.6          & 23.3          & 2.9           & 2.8           & 28.7          & 18.6          & 2.3           & 21.9          & \multicolumn{1}{c|}{23.4}          & 16.8          \\
\multicolumn{1}{l|}{Mask-RCNN + ranker}     & 9.0                      & 10.0          & 11.9          & 11.2          & 15.3          & 4.4           & 14.4          & 1.8           & 2.3           & 12.0          & 6.2           & 3.1           & 9.1           & \multicolumn{1}{c|}{19.1}          & 9.3           \\
\multicolumn{1}{l|}{Mask-RCNN + classifier} & 30.0                     & 27.3          & 41.4          & 30.3          & 39.9          & 14.8          & 35.7          & 12.5          & 13.5          & 46.8          & 19.4          & 16.0          & 43.0          & \multicolumn{1}{c|}{38.1}          & 29.2          \\
\multicolumn{1}{l|}{Mask-RCNN + GGNN}       & 31.8                     & 28.6          & {\ul 45.4}    & 33.7          & 46.8          & 16.6          & 37.8          & 15.1          & 15.0          & \textbf{49.9} & 24.9          & 18.9          & {\ul 49.8}    & \multicolumn{1}{c|}{39.7}          & 32.4          \\
\multicolumn{1}{l|}{MDETR + GGNN}       & 36.9                     & 31.3          & 43.6    & 17.1          & 42.9          & {\ul 20.1}          & 19.9          & {\ul 18.7}          & {\ul 24.5}          & 45.5 & 23.1          & {\ul 30.9}          & 46.2    & \multicolumn{1}{c|}{24.0}          & 30.3          \\ \midrule
\multicolumn{1}{l|}{TOIST}                  & {\ul 37.0}               & {\ul 34.4}    & 44.7          & {\ul 34.2}    & {\ul 51.3}    & 18.6    & {\ul 40.5}    & 17.1    & 23.4    & 43.8          & {\ul 29.3}    & 29.9    & 46.6          & \multicolumn{1}{c|}{{\ul 42.4}}    & {\ul 35.2}    \\
\multicolumn{1}{l|}{TOIST w/ distillation}  & \textbf{40.8}            & \textbf{36.5} & \textbf{48.9} & \textbf{37.8} & \textbf{53.4} & \textbf{22.1} & \textbf{44.4} & \textbf{20.3} & \textbf{26.9} & {\ul 48.1}    & \textbf{31.8} & \textbf{34.8} & \textbf{51.5} & \multicolumn{1}{c|}{\textbf{46.3}} & \textbf{38.8} \\ \bottomrule
\end{tabular}}
\label{table:seg}
\end{table}

%%%%%%%%%%%%%%%%%%%%%%%%%%%%%%%%%%%%%%%%%%%%%%%%%5

\section{More Analysis}\label{appendix:analysis}
\begin{figure*}
\begin{center}
\includegraphics[width=0.9\textwidth]{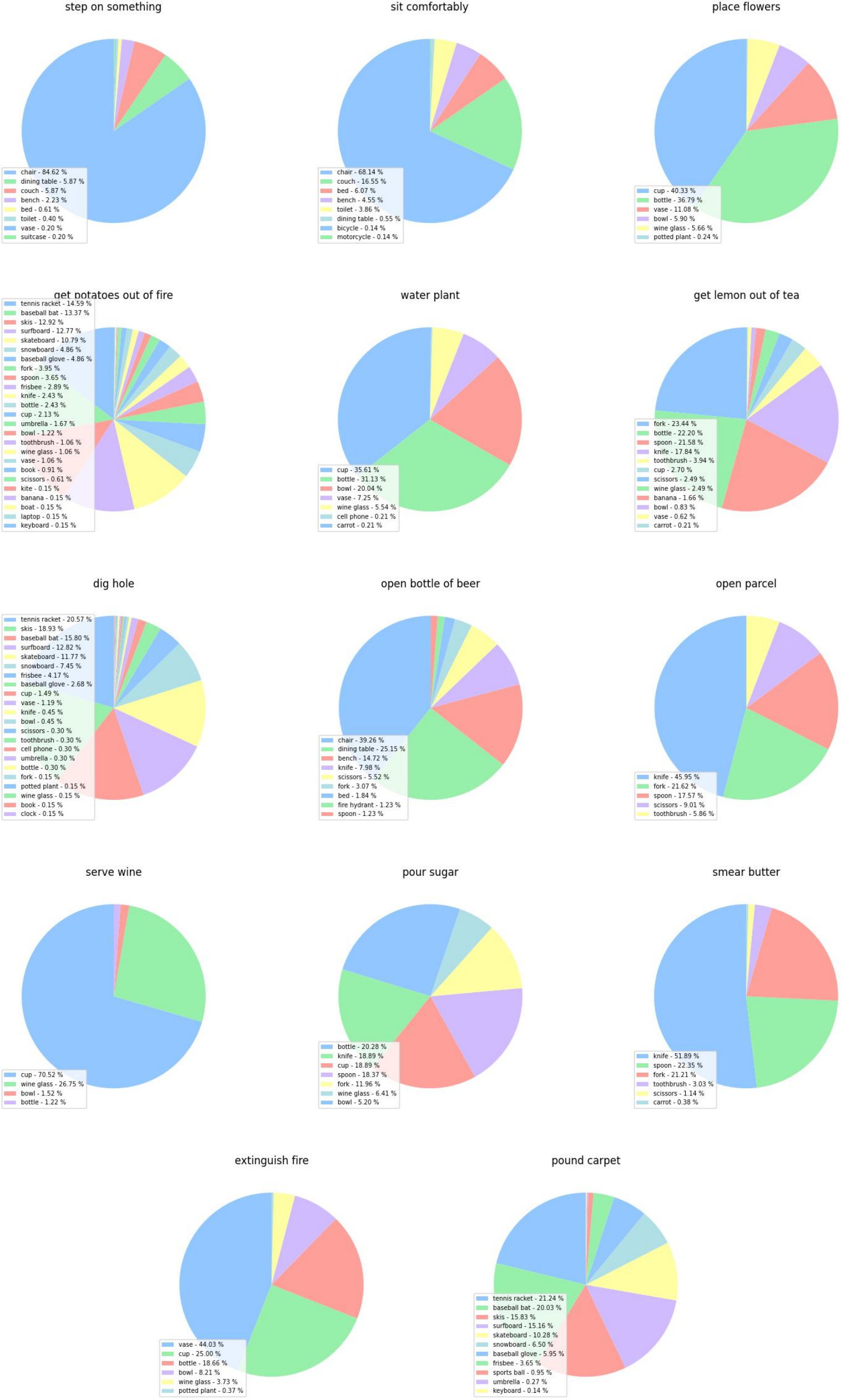}
\end{center}
% \vspace{-0.5em}
\caption{\textcolor[RGB]{0,0,0} {Statistics on the number of images about each ground truth category selected in each task.}}
\label{fig:class_count}
\end{figure*}

\begin{figure*}
\begin{center}
\includegraphics[width=0.9\textwidth]{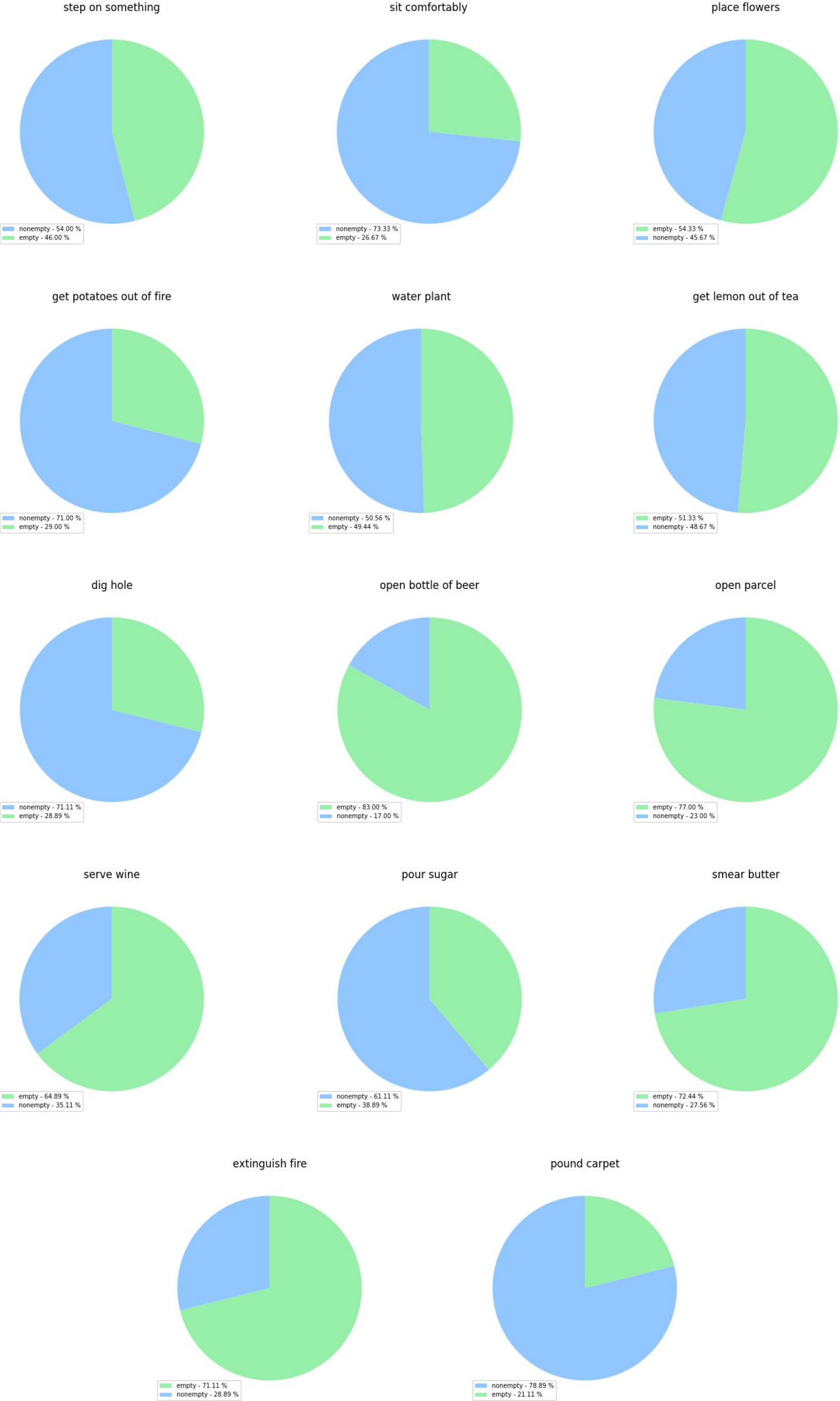}
\end{center}
% \vspace{-0.5em}
\caption{\textcolor[RGB]{0,0,0} {Statistics on the number of images that contains selected objects (denoted as 'nonempty') or not (denoted as 'empty') in each task.}}
\label{fig:empty}
\end{figure*}

In Fig.\ref{fig:class_count}, we present the percentage of ground truth object categories, in each task. The figures show that the distribution of object categories in each task in the COCO-Tasks dataset is very diverse. In Fig.\ref{fig:empty}, we present the percentage of the images that contain at least one ground truth object or contain no object in each task. These figures show that many of the images do not contain objects that are suitable for the tasks. All these statistical results demonstrate that it is non-trivial to find the objects with affordance and preference for a specific task in a given image.

\begin{figure*}
\begin{center}
\includegraphics[width=0.95\textwidth]{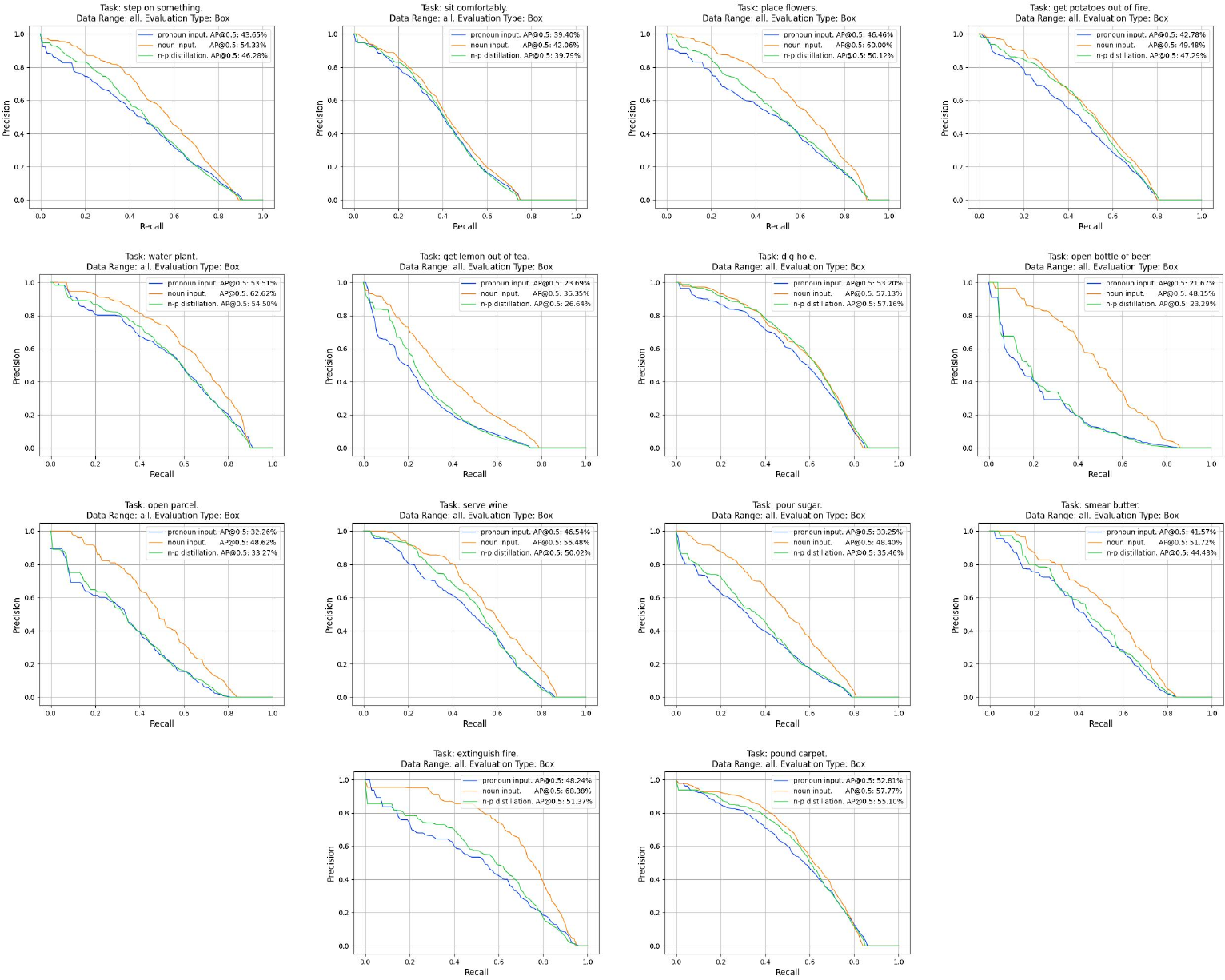}
\end{center}
\caption{\textcolor[RGB]{0,0,0} {The precision-recall curves for object detection on all the test data in each task. The evaluation type 'Box' means object box instead of object mask.}}
\label{fig:all_box}
\end{figure*}

\begin{figure*}
\begin{center}
\includegraphics[width=0.95\textwidth]{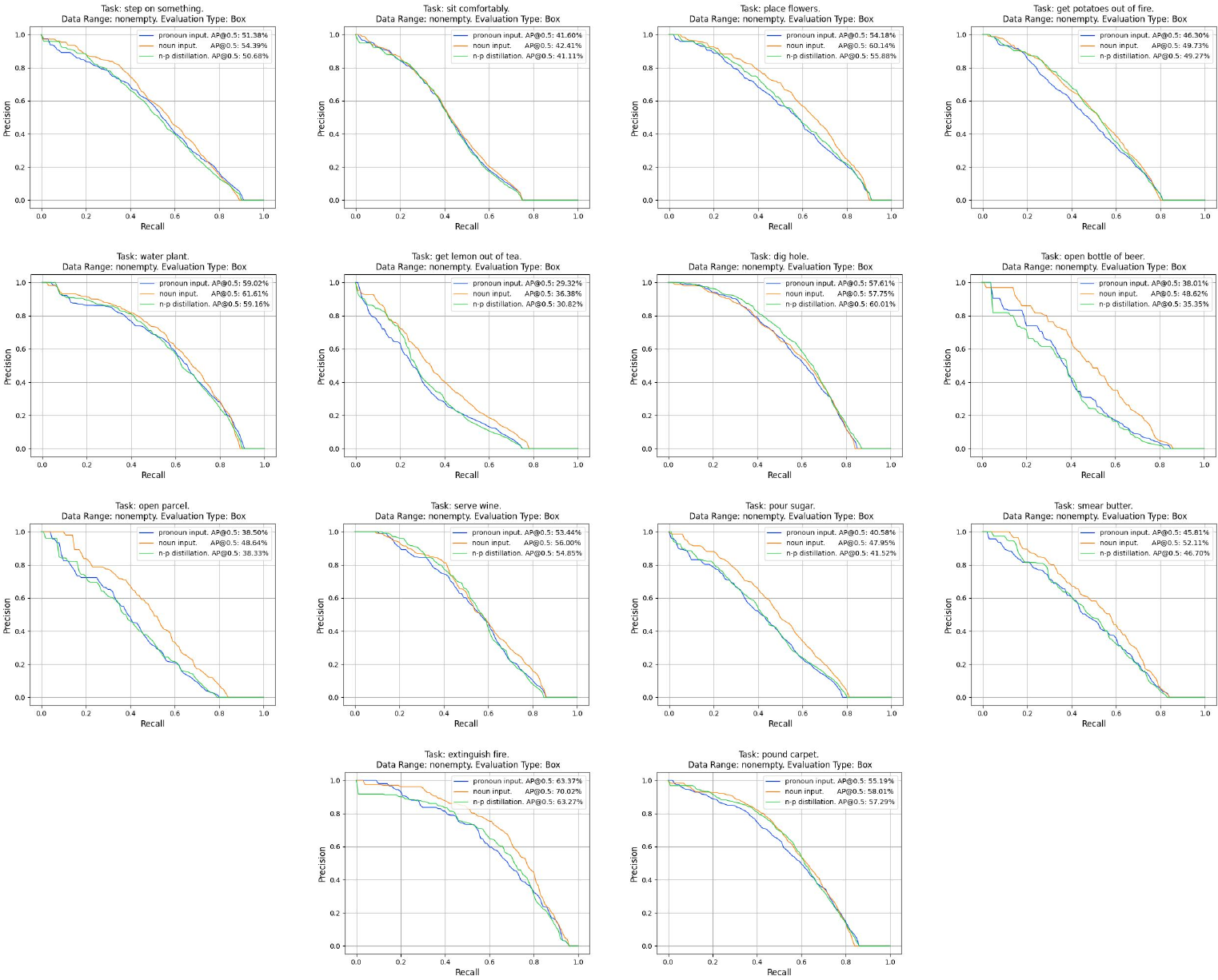}
\end{center}
\caption{\textcolor[RGB]{0,0,0} {The precision-recall curves for object detection on the test data that contain selected objects in each task. The evaluation type 'Box' means object box instead of object mask.}}
\label{fig:nonempty_box}
\end{figure*}

\begin{figure*}
\begin{center}
\includegraphics[width=0.95\textwidth]{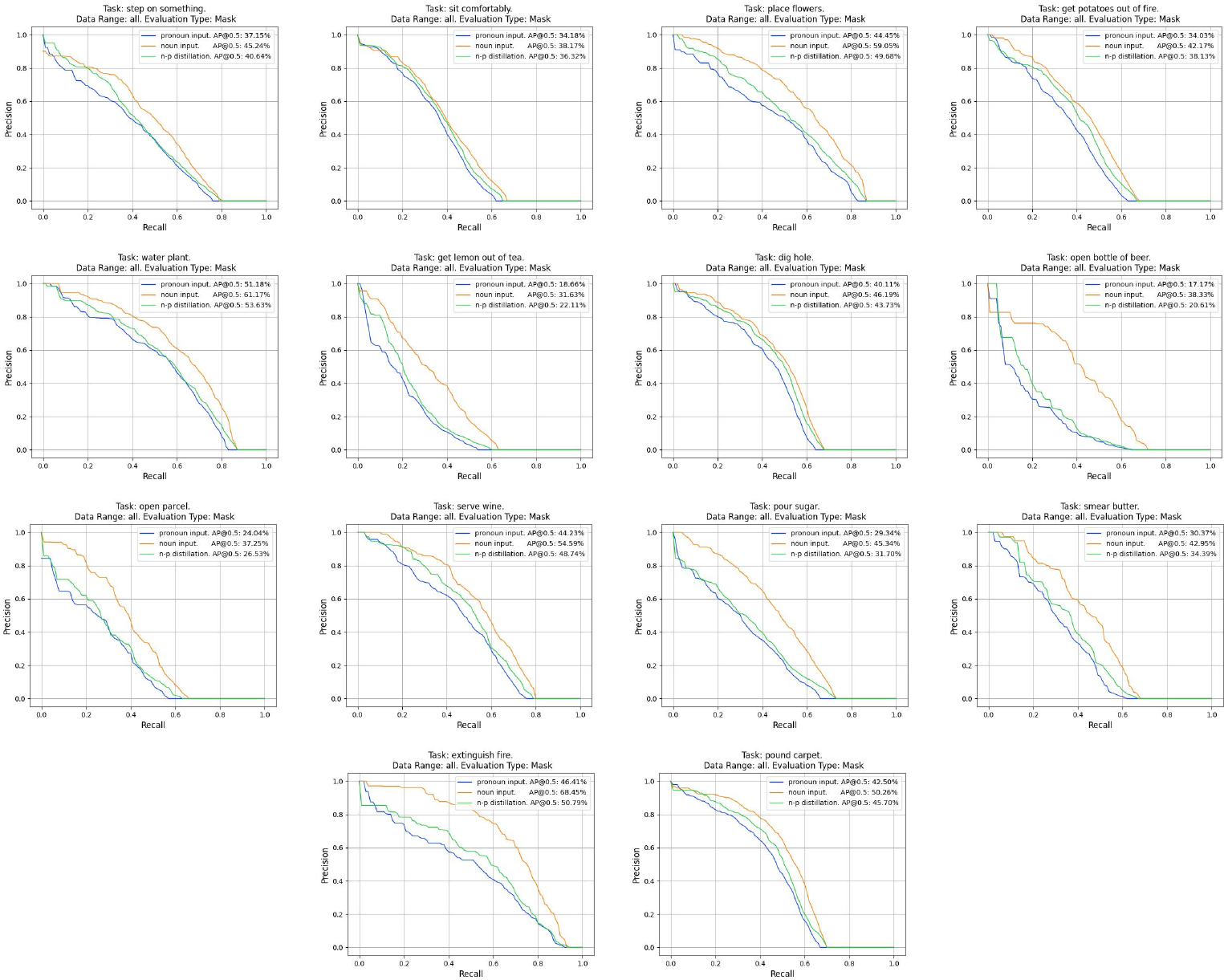}
\end{center}
\caption{\textcolor[RGB]{0,0,0} {The precision-recall curves for instance segmentation on all the test data in each task. The evaluation type 'Mask' means object mask instead of object box.}}
\label{fig:all_mask}
\end{figure*}

\begin{figure*}
\begin{center}
\includegraphics[width=0.95\textwidth]{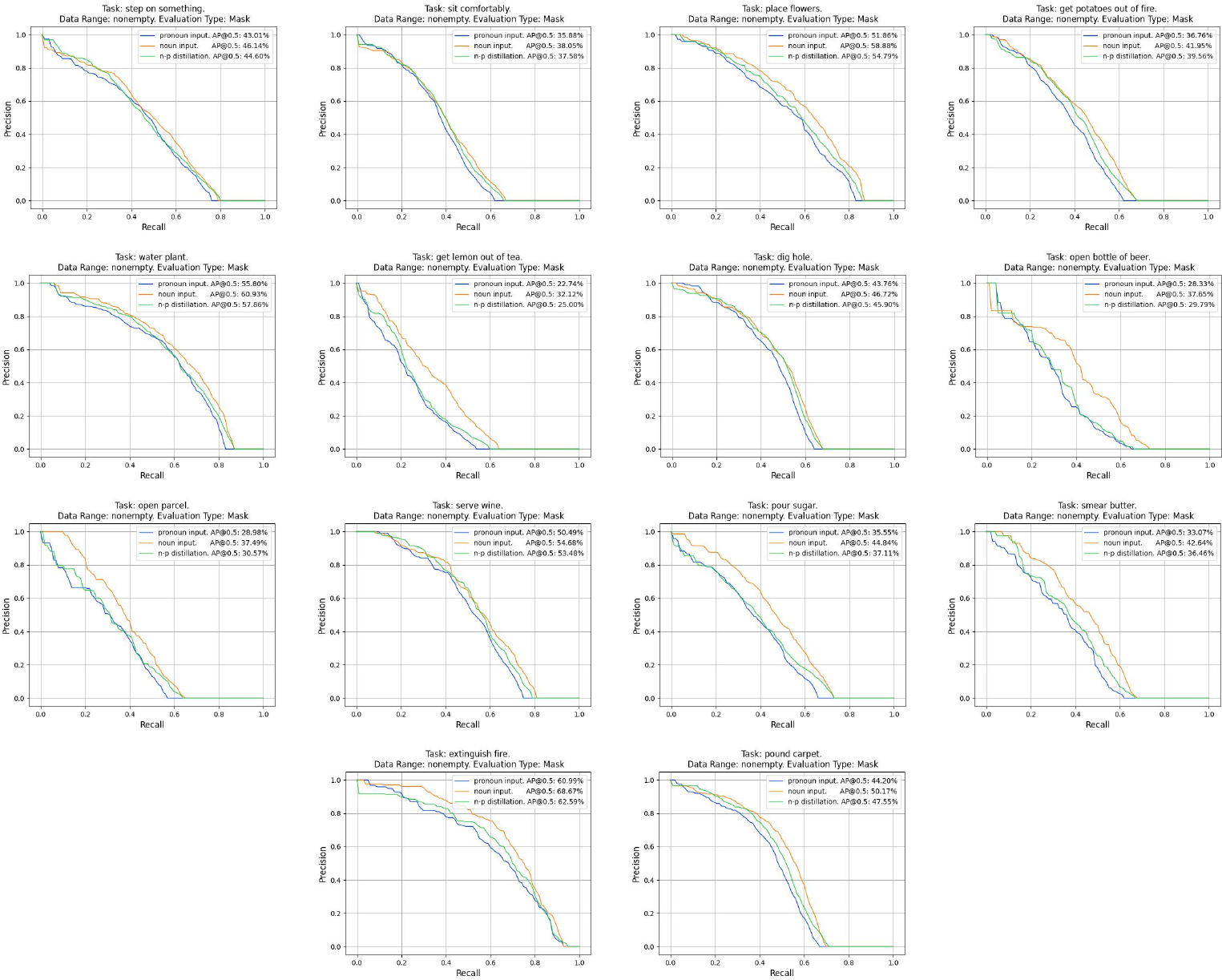}
\end{center}
\caption{\textcolor[RGB]{0,0,0} {The precision-recall curves for instance segmentation on the test data that contain selected objects in each task. The evaluation type 'Mask' means object mask instead of object box.}}
\label{fig:nonempty_mask}
\end{figure*}

\label{appendix:prcurves}
We present the precision-recall curves of our detection and segmentation results on all the test data or the test data that contain at least one ground truth object (termed nonempty data) in Fig.\ref{fig:all_box}-\ref{fig:nonempty_mask}. 
It can be seen that the curves of the results with noun input are generally higher than those with pronoun input.
The proposed noun-pronoun distillation makes the curves of the results with pronoun input closer to those with noun input, which demonstrates the effectiveness of the distillation method.
Meanwhile, on the nonempty data, the noun-pronoun distillation only marginally improves TOIST, indicating that the method makes TOIST more capable of filtering out objects that do not afford the tasks in the empty data.

We further present the precision-recall curves of our results on the test data that contain objects of specified classes in each task.
The results show that the distillation method has different performance on different categories.
For example, in the task \emph{step on something}, it works for the data containing \emph{chair} but does not work for those containing \emph{dining table} or \emph{couch} (see the first row of Fig.\ref{fig:per_class_box}). 
However, in the task \emph{place flowers} and \emph{get potatoes out of fire}, the proposed distillation method works for almost every category (see the fifth row to the twelfth row of Fig.\ref{fig:per_class_box}). 
Combined with the statistics of category proportion in Fig.\ref{fig:class_count}, we think one possible reason is that the effect of the distillation on different categories is influenced by the proportion of categories in the tasks. 
When a few classes take a large portion of selected objects in a certain task, the effect of the distillation on these classes is good, while that on others is poor. If the number of categories in the whole task is distributed more evenly, the distillation can boost performance for most categories.

\begin{figure*}
\centerline{\includegraphics[width=0.95\textwidth]{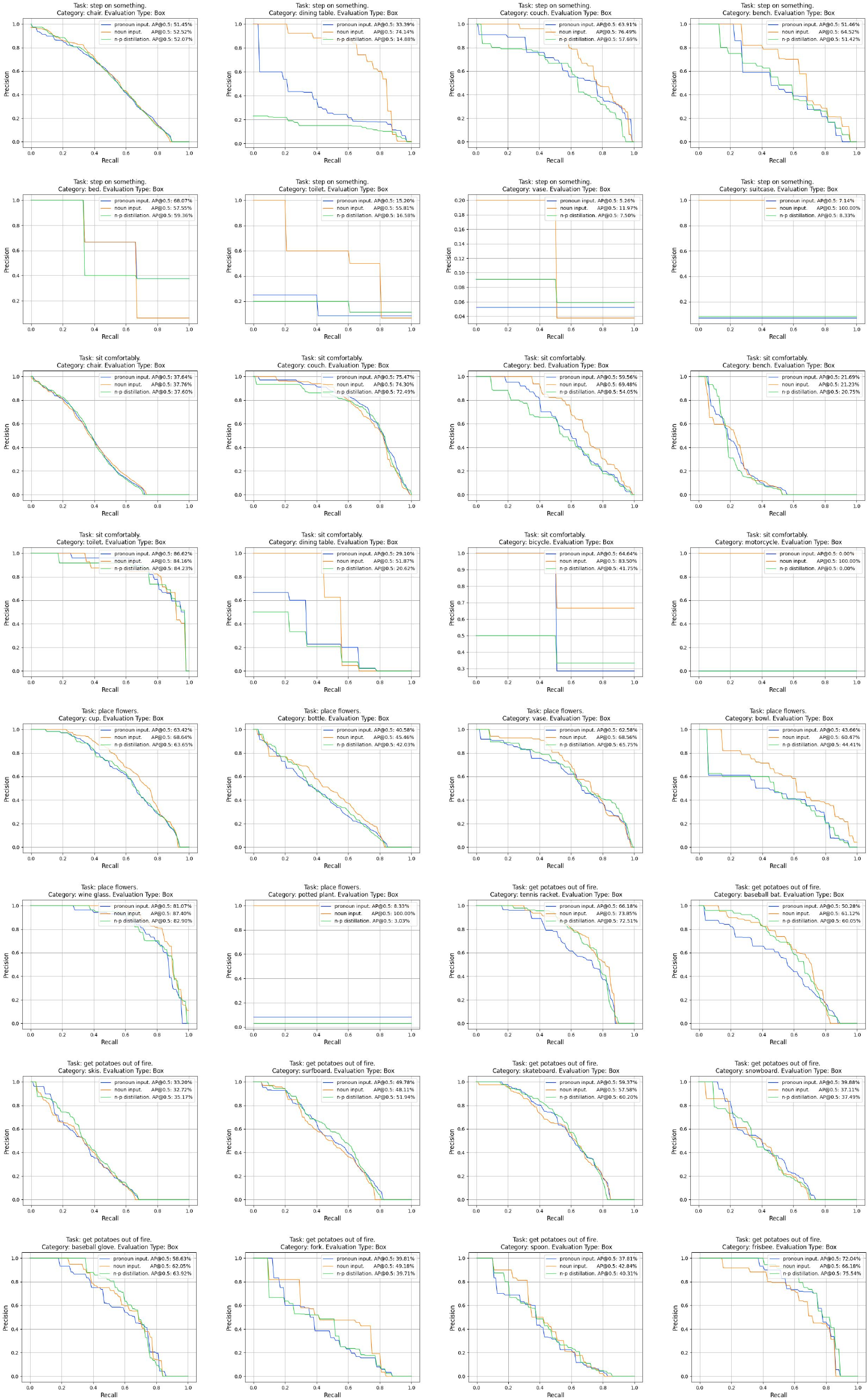}}
\caption{\textcolor[RGB]{0,0,0} {The precision-recall curves for object detection on the test data that contain objects of specified classes in each task.}}
\label{fig:per_class_box}
\end{figure*}

\begin{figure*}
\ContinuedFloat
\centerline{\includegraphics[width=0.95\textwidth]{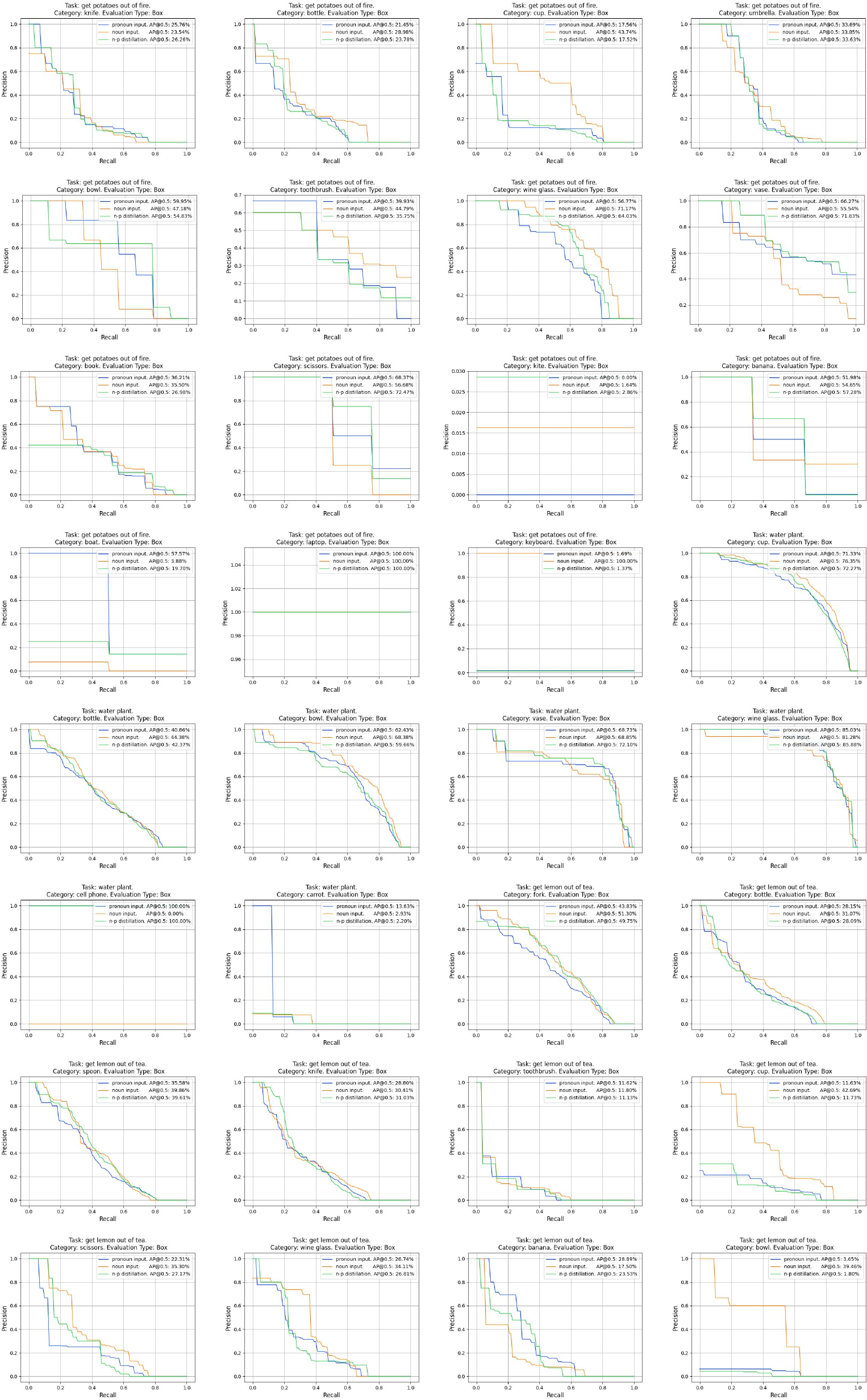}}
\caption{\textcolor[RGB]{0,0,0} {The precision-recall curves for object detection on the test data that contain objects of specified classes in each task (cont.).}}
% \label{fig:per_class_box}
\end{figure*}

\begin{figure*}
\ContinuedFloat
\centerline{\includegraphics[width=0.95\textwidth]{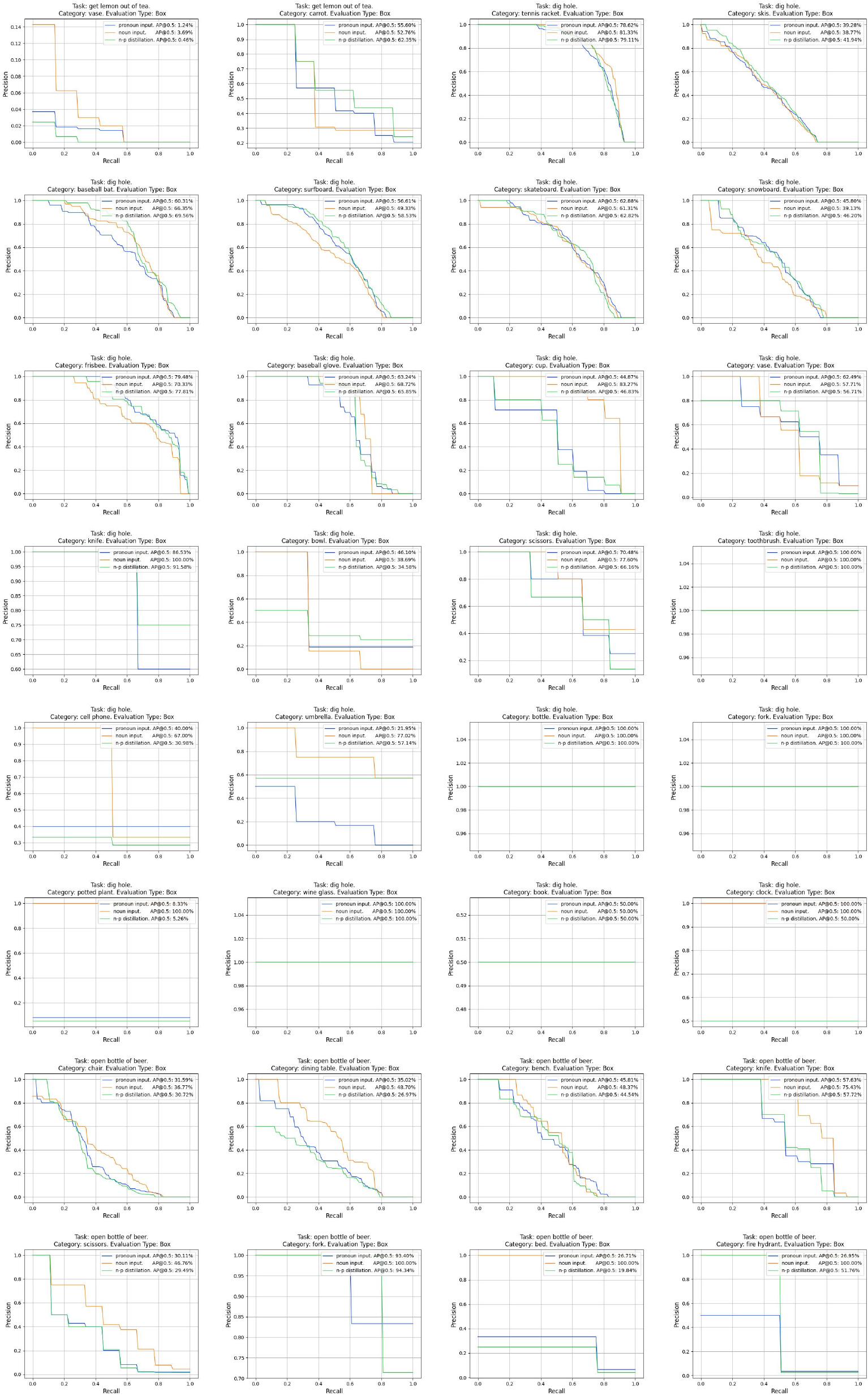}}
\caption{\textcolor[RGB]{0,0,0} {The precision-recall curves for object detection on the test data that contain objects of specified classes in each task (cont.).}}
% \label{fig:per_class_box}
\end{figure*}

\begin{figure*}
\ContinuedFloat
\centerline{\includegraphics[width=0.95\textwidth]{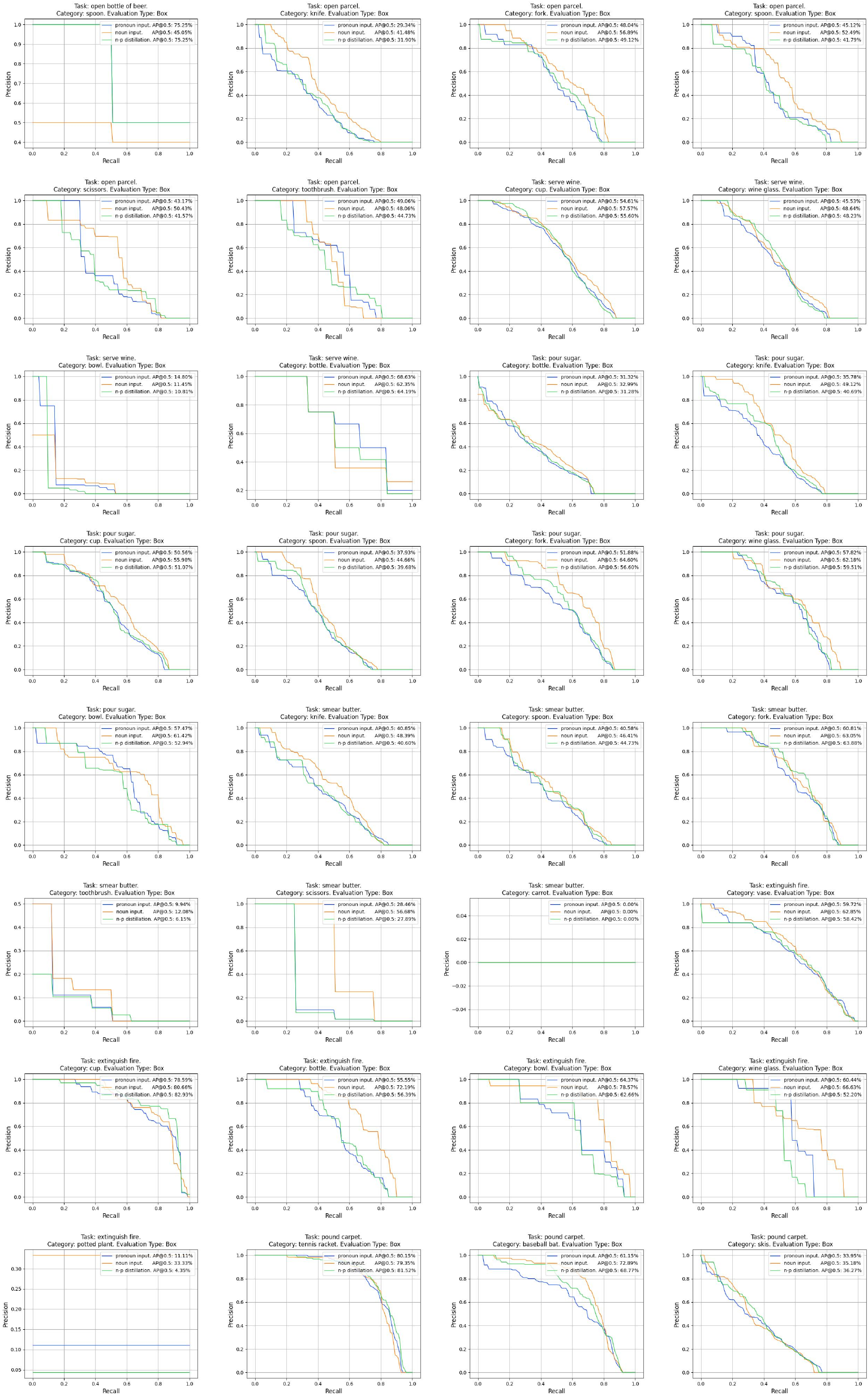}}
\caption{\textcolor[RGB]{0,0,0} {The precision-recall curves for object detection on the test data that contain objects of specified classes in each task (cont.).}}
% \label{fig:per_class_box}
\end{figure*}

\begin{figure*}
\ContinuedFloat
\centerline{\includegraphics[width=0.95\textwidth]{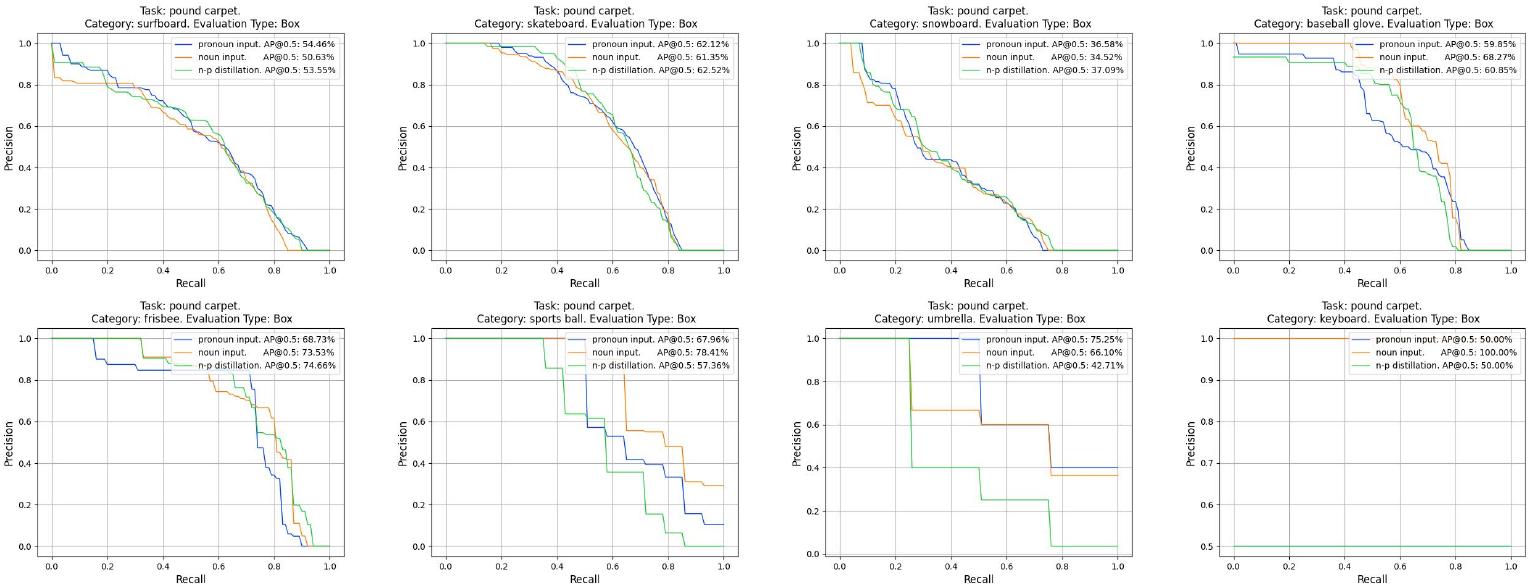}}
\caption{\textcolor[RGB]{0,0,0} {The precision-recall curves for object detection on the test data that contain objects of specified classes in each task (cont.).}}
% \label{fig:per_class_box}
\end{figure*}

\begin{figure*}
\centerline{\includegraphics[width=0.95\textwidth]{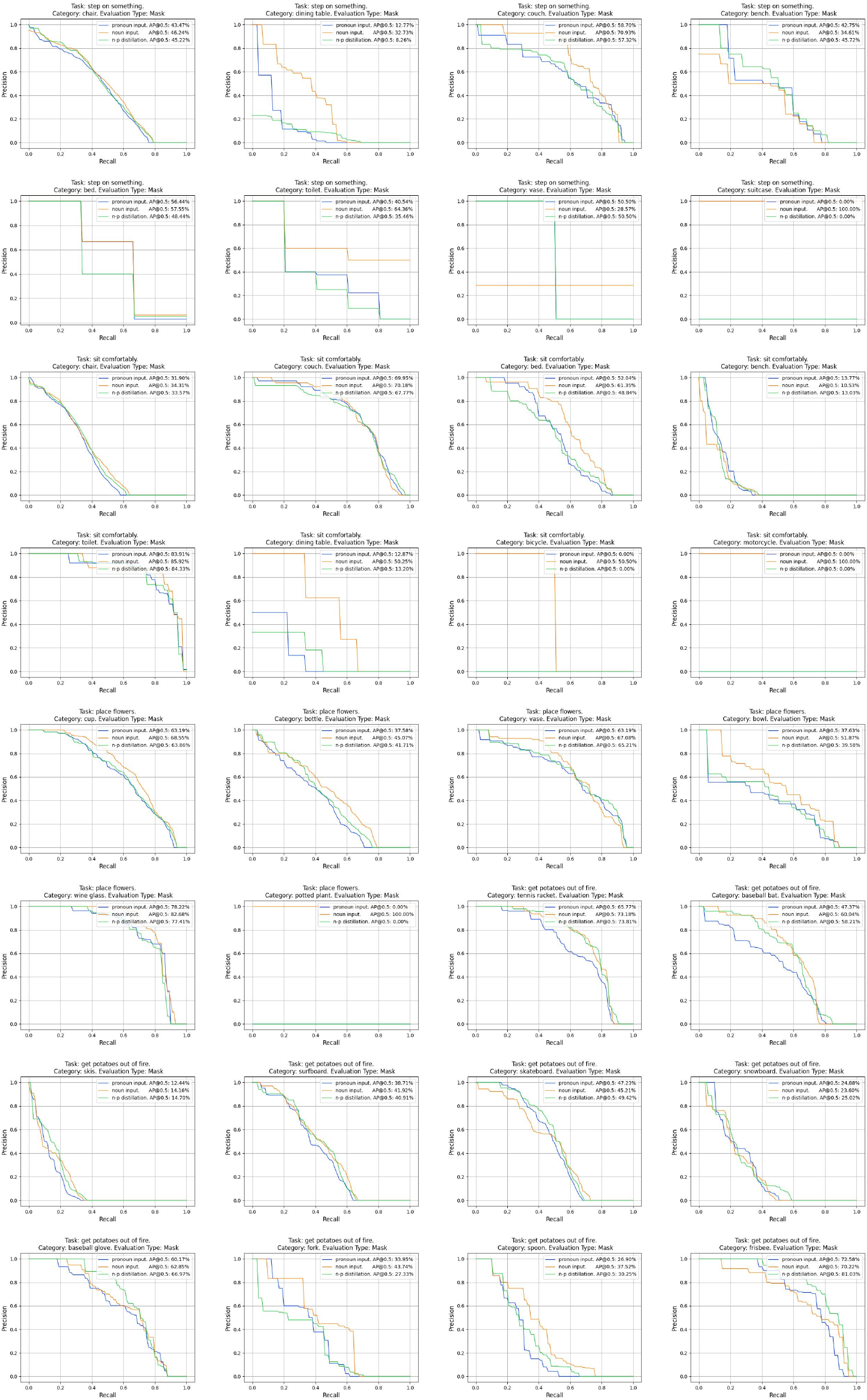}}
\caption{\textcolor[RGB]{0,0,0} {The precision-recall curves for instance segmentation on the test data that contain objects of specified classes in each task.}}
\label{fig:per_class_mask}
\end{figure*}

\begin{figure*}
\ContinuedFloat
\centerline{\includegraphics[width=0.95\textwidth]{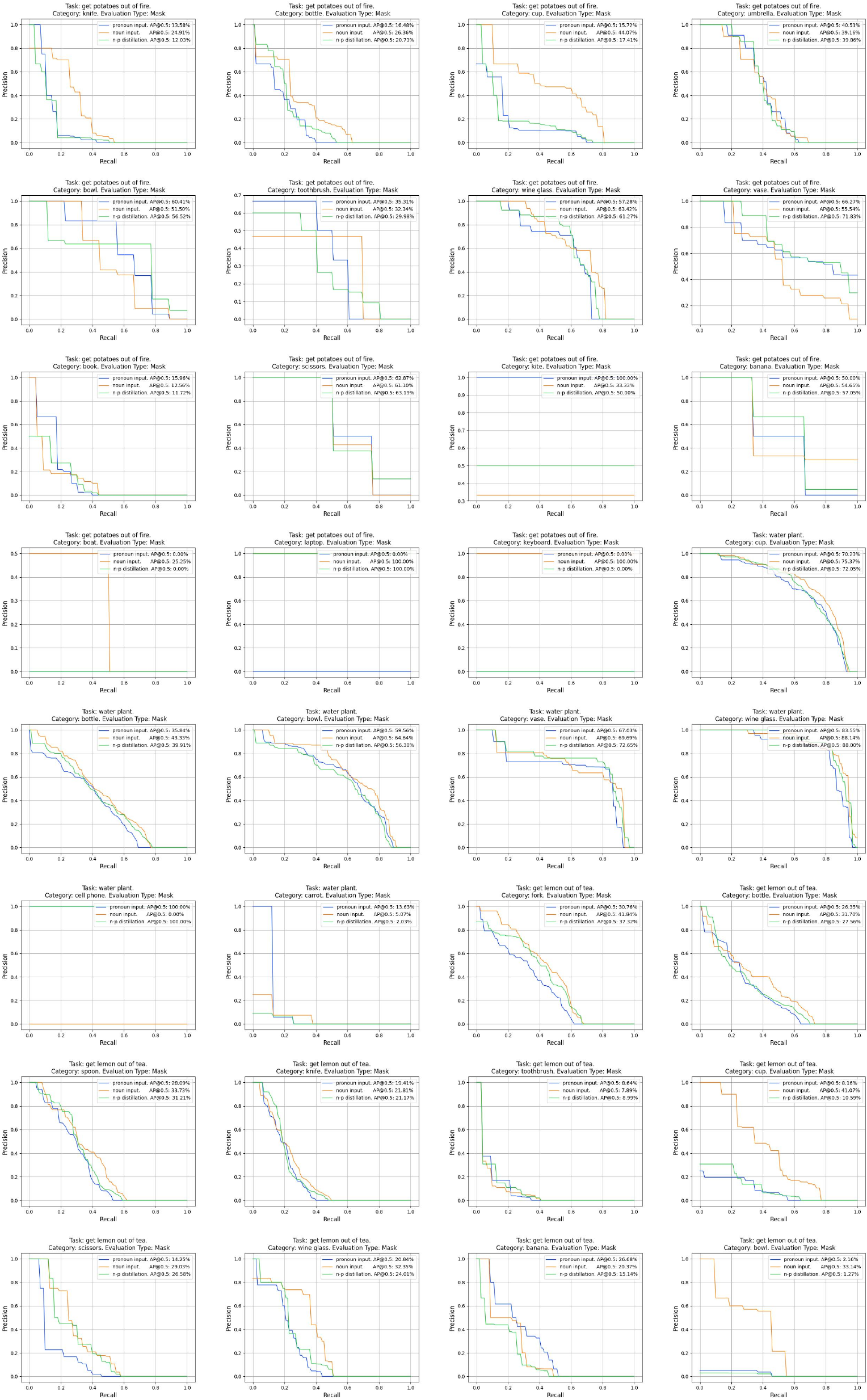}}
\caption{\textcolor[RGB]{0,0,0} {The precision-recall curves for instance segmentation on the test data that contain objects of specified classes in each task (cont.).}}
% \label{fig:per_class_mask}
\end{figure*}

\begin{figure*}
\ContinuedFloat
\centerline{\includegraphics[width=0.95\textwidth]{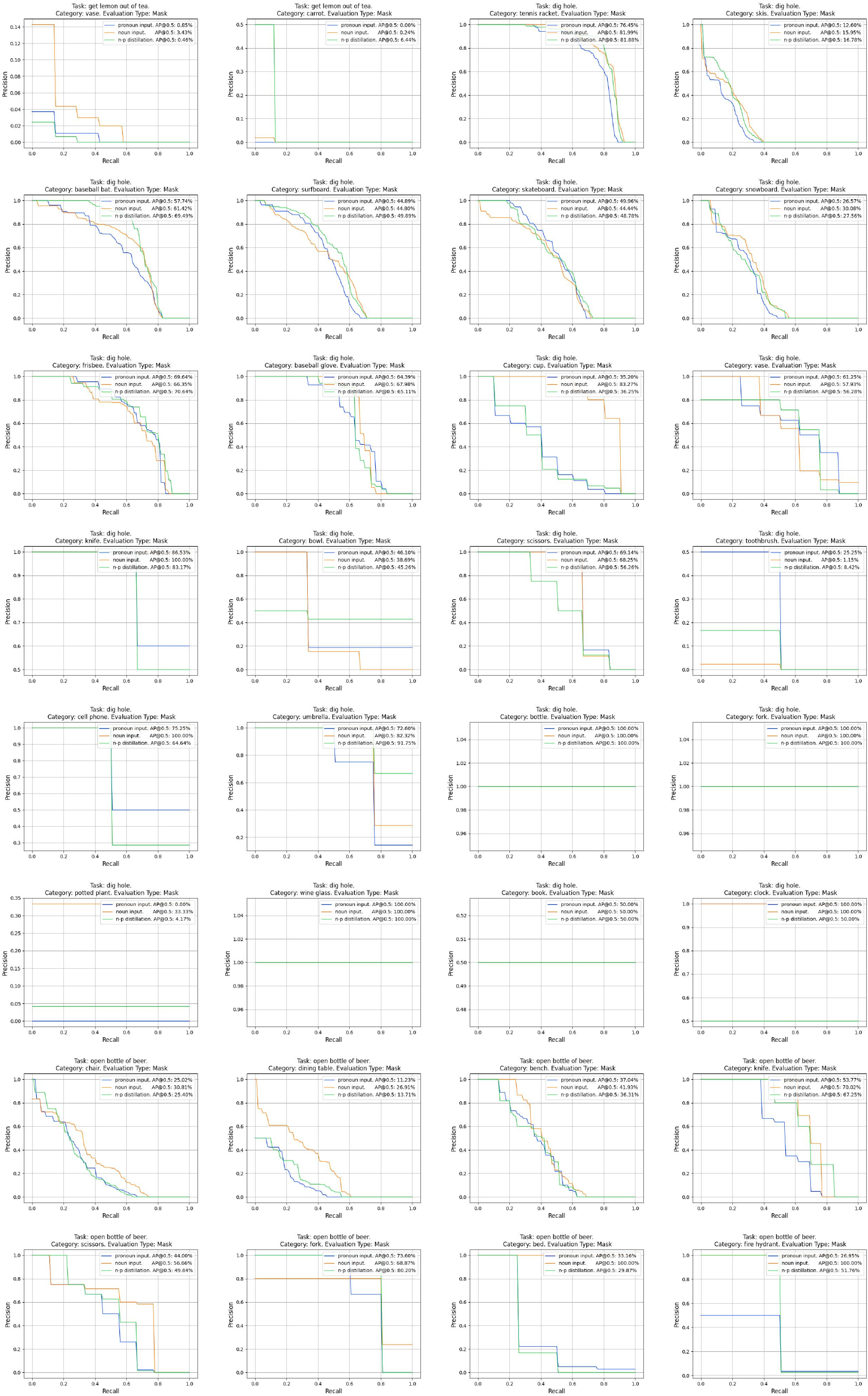}}
\caption{\textcolor[RGB]{0,0,0} {The precision-recall curves for instance segmentation on the test data that contain objects of specified classes in each task (cont.).}}
% \label{fig:per_class_mask}
\end{figure*}

\begin{figure*}
\ContinuedFloat
\centerline{\includegraphics[width=0.95\textwidth]{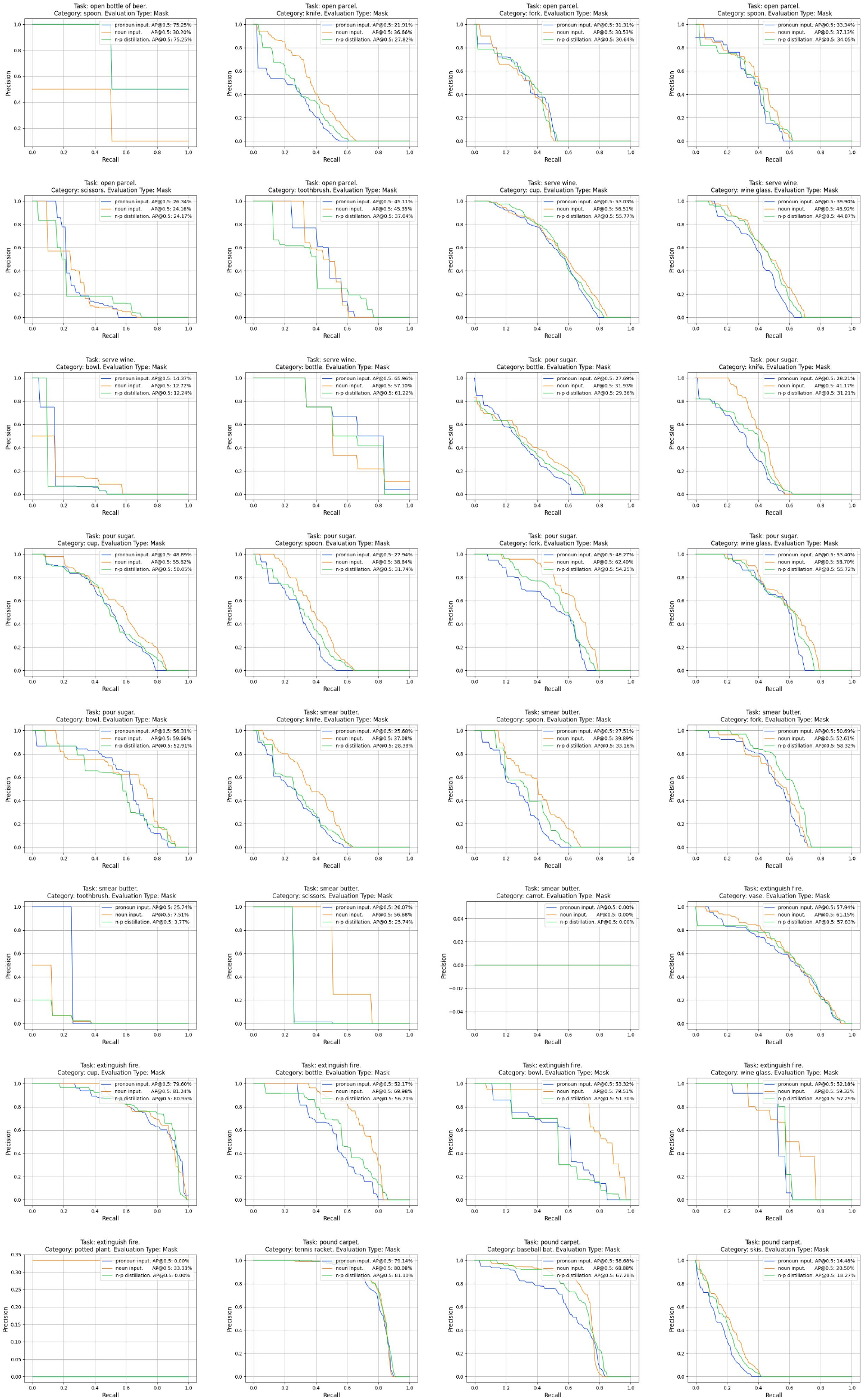}}
\caption{\textcolor[RGB]{0,0,0} {The precision-recall curves for instance segmentation on the test data that contain objects of specified classes in each task (cont.).}}
% \label{fig:per_class_mask}
\end{figure*}

\begin{figure*}
\ContinuedFloat
\centerline{\includegraphics[width=0.95\textwidth]{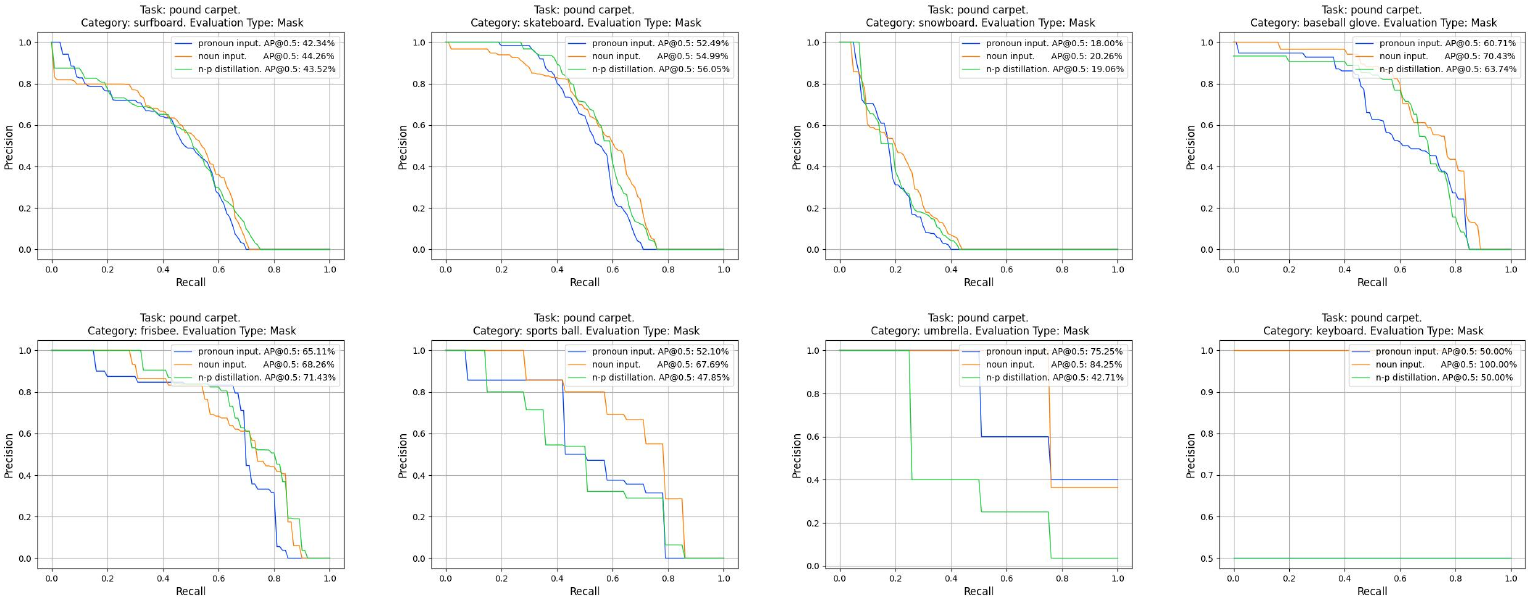}}
\caption{\textcolor[RGB]{0,0,0} {The precision-recall curves for instance segmentation on the test data that contain objects of specified classes in each task (cont.).}}
% \label{fig:per_class_mask}
\end{figure*}

%%%%%%%%%%%%%%%%%%%%%%%%%%%%%%%%%%%%%%%%%%%%%%%%%5

\section{More Qualitative results}\label{appendix:Qualitative}
We present more qualitative results in Fig.\ref{fig:more_qualitative}.
For each task, we select four diverse scenes for visualization.
These results show that our method is robust to complex scenes of different tasks.

\section{Used and Released Asserts}\label{appendix:asserts}
The license of the assets we used is as follows: (a) MIT License for the COCO-Tasks dataset. (b) Creative Commons Attribution 4.0 License for the Microsoft COCO dataset. (c) Apache License 2.0 for MDETR, DETR and Mask-RCNN implemented by Detectron2.
All existing codes and dataset we used are open-source and allowed for research.
To avoid the disclosure of personally identifiable information and the presentation of the content that might be considered offensive, we have blurred out some of the figures in this paper.
Our code is released under the MIT license.

\begin{figure*}
\centerline{\includegraphics[width=1\textwidth]{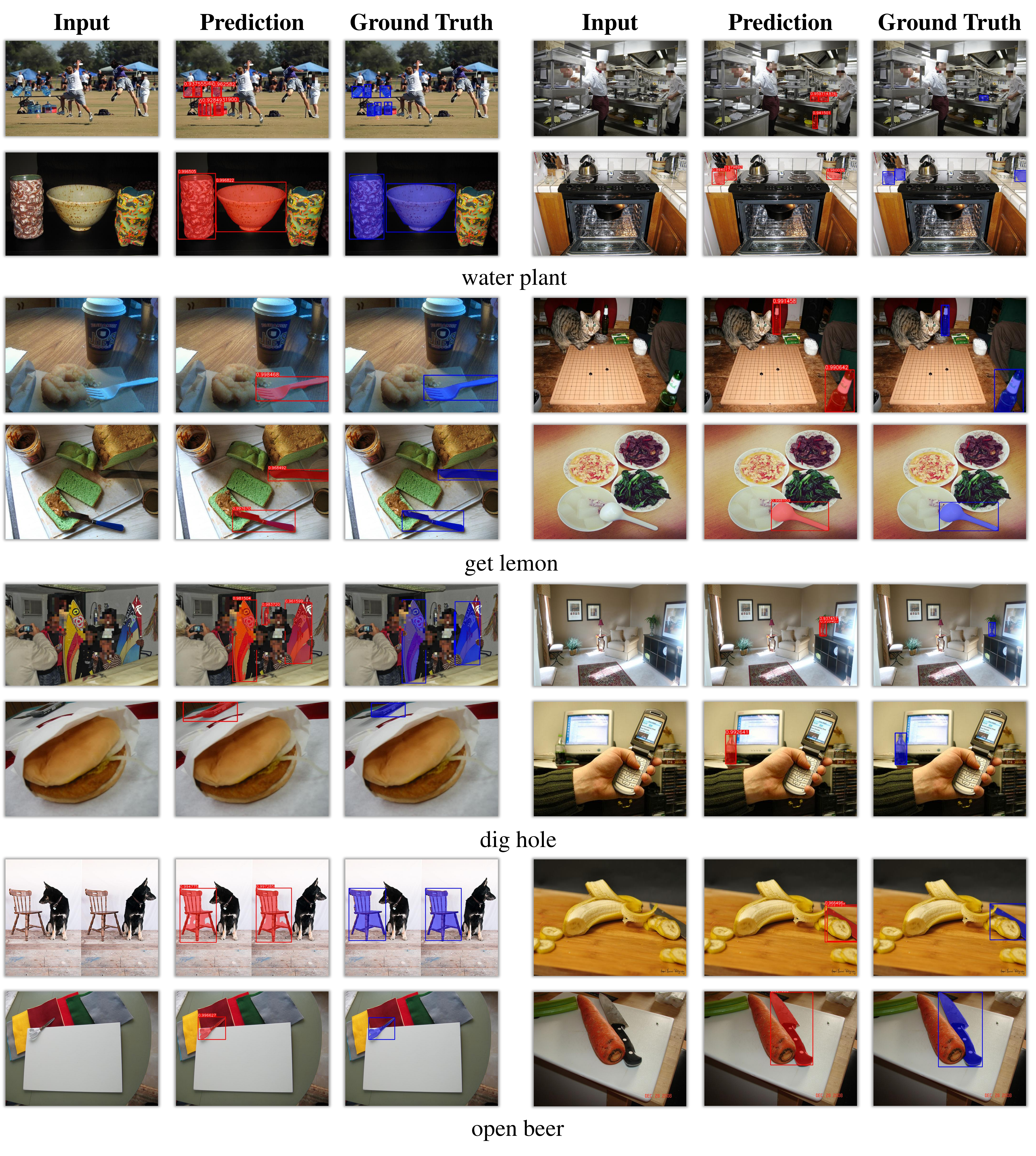}}
\caption{More qualitative results.}
\label{fig:more_qualitative}
\end{figure*}

\begin{figure*}
\ContinuedFloat
\centerline{\includegraphics[width=1\textwidth]{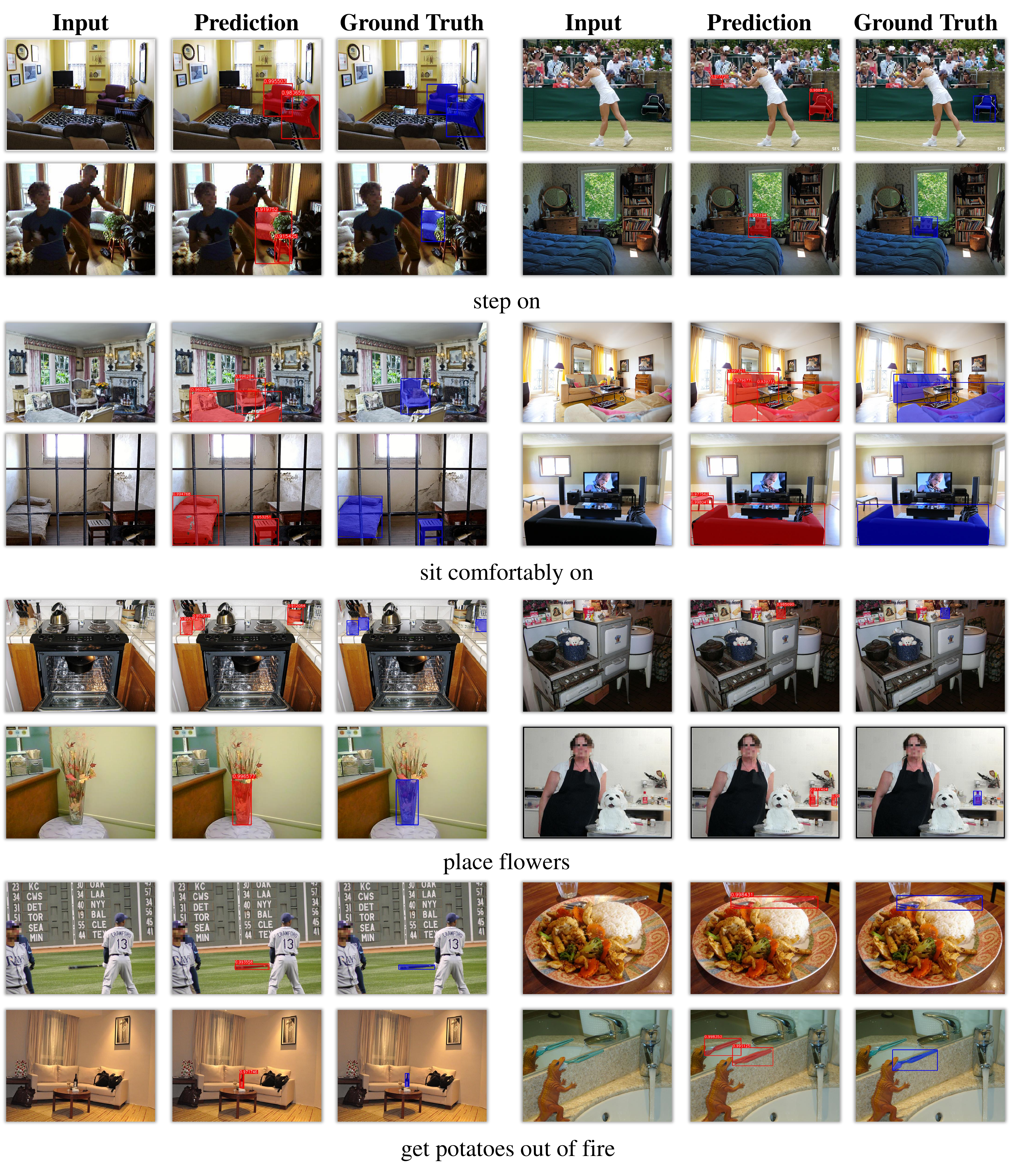}}
\caption{More qualitative results (cont.).}
% \label{fig:more_qualitative}
\end{figure*}

\begin{figure*}
\ContinuedFloat
\centerline{\includegraphics[width=1\textwidth]{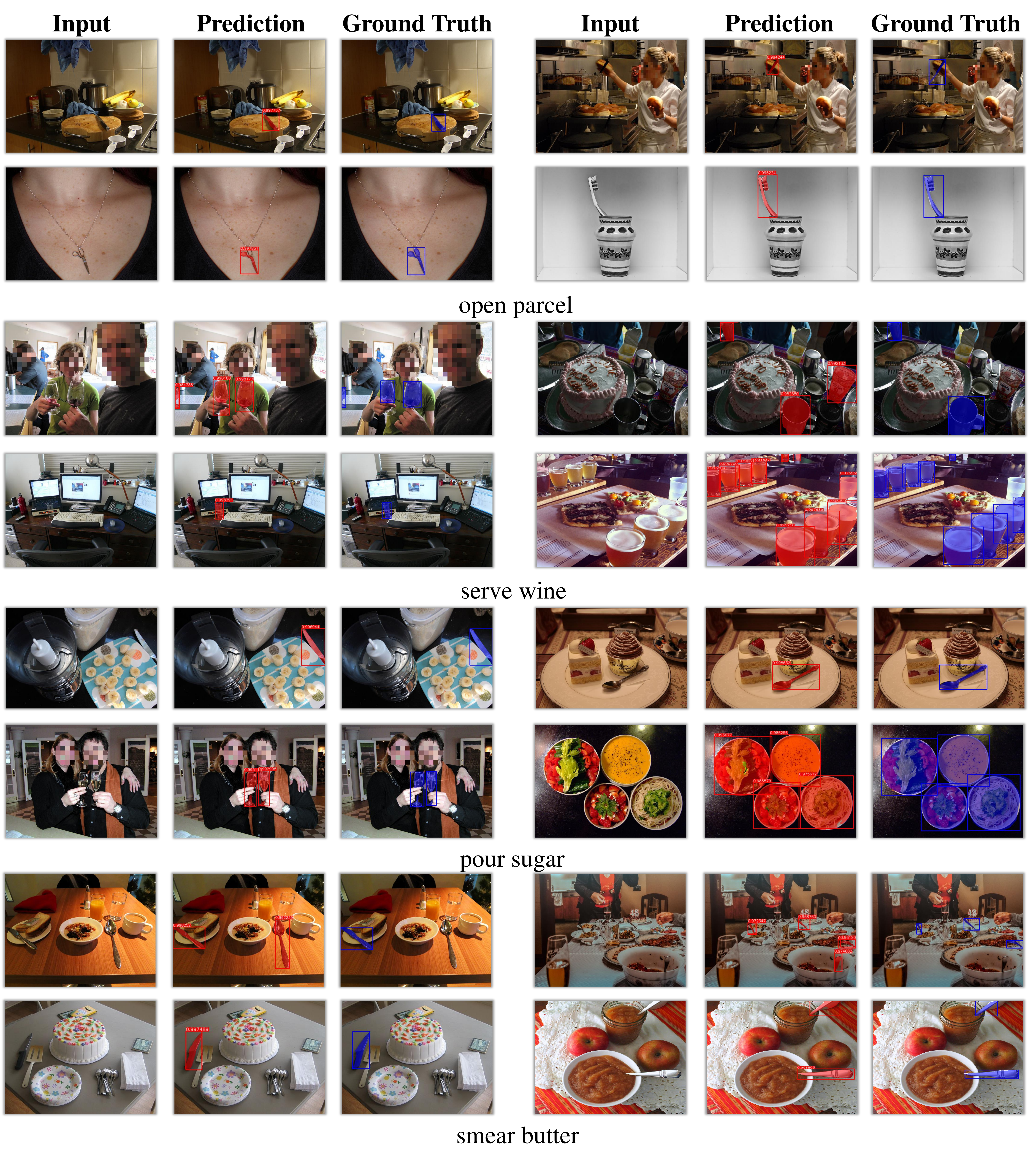}}
\caption{More qualitative results (cont.).}
% \label{fig:more_qualitative}
\end{figure*}

\begin{figure*}
\ContinuedFloat
\centerline{\includegraphics[width=1\textwidth]{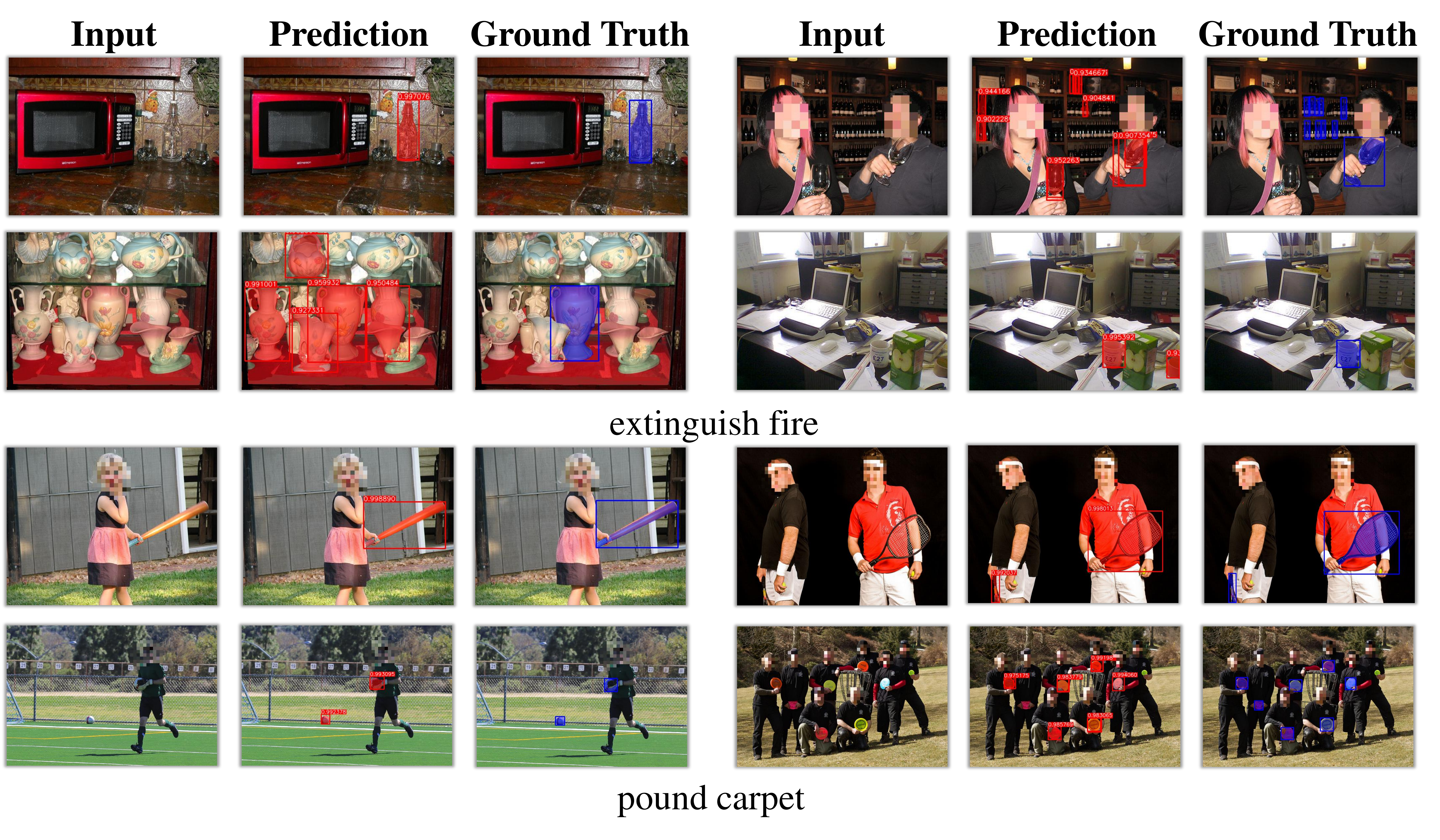}}
\caption{More qualitative results (cont.).}
% \label{fig:more_qualitative}
\end{figure*}

% \bibliographystyle{plain}
% \bibliography{egbib}

\end{document}